%% file: imperfect-imitation-merged.tex
\PassOptionsToPackage{hypertexnames=false}{hyperref}
\documentclass[pdflatex,sn-mathphys-num]{sn-jnl}

\usepackage{graphicx}%
\usepackage{multirow}%
\usepackage{amsmath,amssymb,amsfonts}%
\usepackage{amsthm}%
\usepackage{mathrsfs}%
\usepackage[title]{appendix}%
\usepackage{xcolor}%
\usepackage{textcomp}%
\usepackage{manyfoot}%
\usepackage{booktabs}%
\usepackage{algorithm}%
\usepackage{algorithmicx}%
\usepackage{algpseudocode}%
\usepackage{listings}%
\usepackage{subcaption}

\usepackage{bibunits}
\newif\ifsupplementbibliography

\makeatletter
\AddToHook{env/bibunit/begin}{%
  \def\@extra@binfo{@\@bibunitname}%
  \let\@extra@b@citeb\@extra@binfo
}
\makeatother

\usepackage{placeins}

\usepackage{pifont}
\usepackage{makecell} 
\newcommand{\cmark}{\ding{51}} 
\newcommand{\xmark}{\ding{55}} 

\newcommand{\bE}{\mathbb{E}}
\newcommand{\bR}{\mathbb{R}}

\theoremstyle{thmstyleone}%
\newtheorem{theorem}{Theorem}
\theoremstyle{thmstyletwo}%

\theoremstyle{thmstylethree}%
\newtheorem{lemma}[theorem]{Lemma}

\begin{document}
\begin{bibunit}[sn-mathphys-num]
\title[Article Title]{Imitating the Imperfect: Offline-to-Online Robust Imitation Learning from Heterogeneous Demonstrators}


\author*[1]{\fnm{Cheng} \sur{Pan}}\email{cheng.pan@epfl.ch}
\equalcont{These authors contributed equally to this work.}

\author*[2]{\fnm{Nan} \sur{Lu}}\email{nan.lu@bristol.ac.uk}
\equalcont{These authors contributed equally to this work.}

\author[3]{\fnm{Kai} \sur{Arulkumaran}}\email{kailash@sakana.ai}

\author[1]{\fnm{Josie} \sur{Hughes}}\email{josie.hughes@epfl.ch}

\affil[1]{\orgname{Swiss Federal Institute of Technology in Lausanne (EPFL)}, \orgaddress{\city{Lausanne}, \country{Switzerland}}}

\affil[2]{\orgname{University of Bristol}, \orgaddress{\city{Bristol}, \country{UK}}}

\affil[3]{\orgname{Araya Inc.}, \orgaddress{\city{Tokyo}, \country{Japan}}}


\abstract{Imitation learning~(IL) typically relies on large-scale \emph{offline} demonstrations collected from multiple human or algorithmic demonstrators. 
However, most existing approaches assume these demonstrators are homogeneous or near-optimal experts, a convenient but unrealistic assumption in real-world applications. 
In practice, demonstrations are often collected from heterogeneous demonstrators with unknown and varying levels of expertise, resulting in highly mixed-quality data.
In this work, we study a challenging yet practical setting of imitation learning from such heterogeneous imperfect demonstrators. 
We propose \emph{Latent Expertise and Optimality Scoring based Imitation Learning} (LEOS-IL), 
a principled framework that jointly learns (i) a state-action optimality scoring model and (ii) latent expertise levels for each demonstrator from unlabeled suboptimal \emph{offline} demonstrations. The learned scoring model is then integrated into an \emph{online} policy optimization procedure, where the agent is trained to maximize estimated optimality scores while its online rollouts are iteratively leveraged to refine the scoring model, thereby mitigating distributional shift during online adaptation.
{
We further provide theoretical guarantees for optimality score identification and conditional optimal occupancy and policy recovery under explicit sufficient conditions.} Experiments on continuous control benchmarks demonstrate that our approach consistently outperforms existing baselines across highly heterogeneous and low-quality demonstration regimes.}




\maketitle

\section{Introduction}\label{sec1}

Imitation learning {(IL)} offers a convenient framework that enables agents to acquire skills directly from expert demonstrations \cite{zare2024survey}. Prior research has shown that effective policies can be learned efficiently when provided with large-scale and high-quality demonstrations \cite{belkhale2023data,saxena2025matters}.
However, in real-world scenarios, large-scale datasets are often collected from  \emph{multiple} demonstrators, and human or algorithmic demonstrators typically possess strengths in specific contexts and may perform suboptimally outside their areas of expertise. 
For example, in autonomous driving data collected from electric vehicles \cite{lee2022semi,zhang2024100}, drivers demonstrate a wide range of skill levels, resulting in \emph{heterogene}ous and \emph{imperfect} trajectories with unknown degrees of expertise. 
Treating all such demonstrators as perfect experts can lead to unreliable or even unsafe agent behavior, posing serious risks in safety-critical applications.

\begin{figure*}[t]
     \centering
     \includegraphics[width=1\columnwidth]{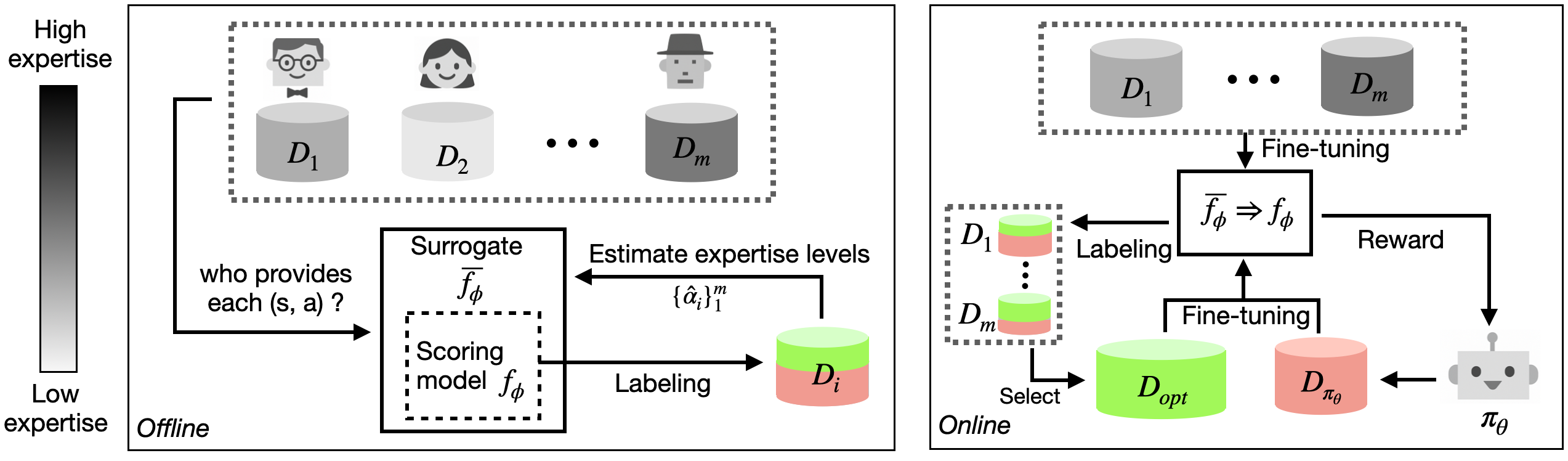}
     \caption{\textbf{Overview of our heterogeneous IL framework}, illustrated with $m$ demonstrators possessing (unknown) heterogeneous expertise levels (represented in grayscale; darker shades indicate higher expertise). Green/red indicate inferred optimal/suboptimal pairs.
     \textbf{Stage 1} (left panel): Joint learning of \emph{demonstrator expertise levels} $\{{\alpha}_i\}_{i=1}^m$ and the \emph{optimality scoring model} $f_\phi$ through a a closed-loop process: (1) a surrogate classifier $\overline{f}_\phi$ is trained to predict the demonstrator identity for each $(s,a)$ pair, which induces $f_\phi$; (2) $\{\hat{\alpha}_i\}_{i=1}^m$ are updated based on predictions from $f_\phi$.
     \textbf{Stage 2} (right panel): Iterative refinement of both $f_\phi$ and the agent policy $\pi_\theta$ using the demonstrator dataset $D = \{(i, D_i)\}_{i=1}^m$, a selected set of inferred optimal demonstrations $D_{\text{opt}}$, and agent-generated samples $D_{\pi_\theta}$.}
     \label{fig:filter_sort}
\end{figure*}

In this work, we study IL from heterogeneous imperfect demonstrators whose expertise levels are unknown and variable, naturally leading to datasets containing suboptimal demonstrations, and learn a policy online based on the estimated demonstrator expertise and data optimality.
Existing methods (see details in Section~\ref{sec:related}) addressing this setting often rely on one or more of the following assumptions: (i) demonstrators are homogeneous, i.e., with the same expertise levels, and the optimal data dominates the dataset, (ii) access to explicit labels indicating the optimality of demonstrations, or (iii) the expertise level of each demonstrator is known a priori.
These requirements are demanding and often impractical when demonstrations come from uncontrolled or large-scale sources.

This brings us to a central question:
\emph{Can we design effective IL algorithms without relying on these assumptions?}
This problem setting is inherently challenging because neither demonstrator expertise levels nor demonstration optimality are observed. In the worst case, all demonstrators may exhibit very low expertise, making it extremely difficult to extract reliable signals for policy learning.

To address the challenges, we propose a novel \emph{Latent Expertise and Optimality Scoring based Imitation Learning} framework (LEOS-IL) for learning from heterogeneous imperfect demonstrators. Our main contributions are summarized below.

\paragraph{Contribution 1: A formulation for modeling heterogeneous imperfect demonstrators.}
We introduce a principled formulation in which each demonstrator is modeled as a flexible mixture of optimal and suboptimal behaviors. We characterize a demonstrator’s \emph{expertise level} as the probability of selecting actions aligned with an underlying latent optimal policy. These expertise levels vary across demonstrators, naturally capturing heterogeneous demonstration quality, while remaining fully latent and unobserved. This formulation provides a principled foundation for learning from realistic mixed-quality datasets without requiring explicit supervision\footnote{{We only require 1 expert query to resolve ambiguity from the purely unsupervised setting (Sec. \ref{sec:em_style_optimization}).}} or curated expert data.

\paragraph{Contribution 2: A principled offline-to-online heterogeneous IL framework.}
Building on this formulation, we propose a novel two-stage IL framework (Fig.~\ref{fig:filter_sort}).
In the offline stage (Stage 1), we jointly estimate demonstrator expertise and learn a \emph{scoring model} that assigns \emph{optimality scores} to each state-action pair, representing the likelihood that the pair corresponds to the (latent) optimal policy.
{This is achieved through an EM-style alternating self-training method that iteratively refines the scoring model and expertise estimates based on its current hard pseudo-labels.}
In the online stage (Stage 2), the learned scoring model is used as a surrogate reward function for policy optimization. To address the distributional shift arising from offline training, we introduce an online iterative refinement procedure that alternates between policy learning and scoring model fine-tuning, yielding a progressively less biased and more informative reward signal.

\paragraph{Contribution 3: Theoretical and empirical validation.}
{
We provide theoretical analysis establishing optimality-score identification and conditional optimal occupancy and policy recovery under explicit sufficient conditions.} Empirically, we evaluate our method in two challenging heterogeneous scenarios: (i) a \emph{wide-range setting}, with demonstrator expertise levels spanning $[0.1, 0.9]$, and (ii) a \emph{low-quality setting}, where all demonstrators are highly suboptimal with expertise in $[0.05, 0.15]$. Across both regimes, our method outperforms baselines and, remarkably, achieves performance comparable to methods trained on the subset of purely optimal demonstrations. These results demonstrate the effectiveness of our framework for learning from uncurated demonstrations with unknown and highly variable quality.

\section{Related work}
\label{sec:related}
We review imitation learning methods under different assumptions on demonstration quality, demonstrator heterogeneity, and supervision. The comparison between our method and representative IL baselines is summarized in Table~\ref{tab:compare}.

\paragraph{IL with Homogeneous Demonstrators}
A large body of IL research assumes that demonstrations originate from a single, near-optimal expert policy. This assumption underpins classical approaches such as \emph{behavior cloning}~\cite[BC;][]{pomerleau1988alvinn}, as well as more recent methods such as \emph{generative adversarial imitation learning}~\cite[GAIL;][]{ho2016generative}, which matches the state-action occupancy measures of the learner and the demonstrator via adversarial training.
Methods designed to handle noisy demonstrations, such as \emph{robust imitation learning}~\cite[RIL;][]{tangkaratt2020robust}, also implicitly retain this homogeneity assumption by modeling demonstrator behavior as arising from a mixture of optimal and suboptimal policies, with a single mixing coefficient representing the expertise level, and by assuming that expert data dominates the dataset.
While effective under relatively homogeneous conditions, these methods may be less well suited to settings where demonstrator expertise varies widely.

\begin{table*}[t]
    \caption{Comparison of IL methods. 
         \cmark\ = satisfies criterion, \xmark\ = does not.
         ``Non-dominant optimal demo'' means the method does not require optimal demonstrations to dominate the dataset.  
         ``Multi-demo support'' refers to the ability to handle demonstrations from multiple demonstrators with varying expertise levels.
         ``Handles structured suboptimality'' indicates whether the method can manage structured (i.e., non-random) suboptimal demonstrations.
         {``No optimality labels for training'' means that no sample-level optimality or demonstrator-expertise labels are used for training. Ours uses a few expert queries only to resolve label permutation; see the query-number ablation in Supplementary Tables~\ref{tab:acc_prec_stages_exp_ask_num}--\ref{tab:query_ablation}.}
         ``Probabilistic demo scoring'' denotes whether the method infers a probabilistic measure of demonstration quality. ``Interactive IL'' denotes whether the method can improve with online environment interaction (without the ground truth reward function).
    }
    \label{tab:compare}
   \centering
    \scriptsize
    \setlength{\tabcolsep}{3pt}
    \renewcommand{\arraystretch}{1.05}
    \resizebox{\textwidth}{!}{%
    \begin{tabular}{l|ccccccccc}
        \toprule
        \textbf{Method}                         & BC & GAIL & RIL & IC-GAIL & UID-GAIL & RCE & WGAIL & ILEED & Ours \\
        \midrule
        {Non-dominant optimal demo}      & \xmark & \xmark & \xmark & \cmark & \cmark & \xmark & \xmark & \cmark & \cmark \\
        Multi-demo support                      & \xmark & \xmark & \xmark & \xmark & \xmark & \xmark & \cmark & \cmark & \cmark \\
        Handles structured suboptimality              & \xmark & \xmark & \cmark & \cmark & \cmark & \xmark & \cmark & \xmark & \cmark \\
        {No optimality labels for training}               & \xmark & \xmark & \cmark & \xmark & {\cmark} & \xmark & \cmark & \cmark & \cmark \\
        Probabilistic demo scoring                      & \xmark & {\cmark} & \cmark & \xmark & {\cmark} & \cmark & \xmark & \xmark & \cmark \\
        Interactive IL                      & \xmark & {\cmark} & \cmark & \cmark & {\cmark} & \cmark & \cmark & \xmark & \cmark \\
        \bottomrule
    \end{tabular}%
    }
\end{table*}

\paragraph{IL with Optimality Labels}
To better handle suboptimal demonstrations, many IL methods use supervision to distinguish optimal from non-optimal behavior.
Extensions of GAIL, such as {\emph{imperfect demonstration and confidence GAIL}}~\cite[IC-GAIL;][]{wu2019imitation} uses confidence labels to reweight imperfect demonstrations, whereas {\emph{positive-unlabeled-style GAIL for unlabeled imperfect demonstrations}}~\cite[UID-GAIL;][]{wang2023unlabeled} does not assume labeled successful trajectories. Instead, it treats the demonstration set as an unlabeled mixture and uses a positive-unlabeled-style risk decomposition based on the current policy occupancy to guide learning toward the residual demonstration distribution.
Similarly, example-based methods like \emph{recursive classification of examples}~\cite[RCE;][]{eysenbach2021replacing} train success classifiers from labeled examples to create proxy rewards. These methods are effective when labeled data are available, but their applicability can be limited when
such supervision is difficult to obtain.


{
\paragraph{IL from Heterogeneous and Imperfect Demonstrations}
More recent efforts relax the assumption of homogeneous, near-optimal demonstrations.
\emph{Weighted GAIL}~\cite[WGAIL;][]{wang2021learning} estimates importance weights from discriminator outputs to emphasize higher-quality samples without requiring known expertise levels.
\emph{IL by estimating expertise of demonstrators}~\cite[ILEED;][]{beliaev2022imitation} jointly estimates a policy and state-dependent demonstrator expertise through offline maximum likelihood, but models suboptimal behavior through uniform action selection and does not use environment interaction for subsequent refinement.
\emph{Variational IL with diverse-quality demonstrations}~\cite[VILD;][]{tangkaratt2020variational} explicitly considers demonstrators of unknown and diverse quality and jointly infers expertise and reward, but explains heterogeneity through demonstrator-specific corruption of latent optimal actions.
Other closely related methods obtain additional quality information through limited supervision. \emph{Learning to discern}~\cite[L2D;][]{kuhar2023learning} uses sparse quality
annotations together with temporal representation learning to identify high-quality demonstrations across different behavioral styles, while \emph{preference aided IL from imperfect demonstrations}~\cite[PAIL;][]{cao2024limited} learns a reward from pairwise human preferences and uses it to reweight demonstrations and incorporate high-quality learner-generated trajectories. 
In contrast, our method instead primarily exploits heterogeneity across demonstrators as surrogate supervision and models expertise at two complementary levels: a demonstrator-level expertise parameter capturing overall skill and a state--action optimality posterior capturing context-specific quality. By combining this two-level expertise model with online scorer refinement, our framework turns heterogeneous, low-quality demonstrations into a usable learning signal for policy improvement with only minimal supervision.%
\footnote{We provide further discussion and comparisons with related methods
in Supplementary Sec.~\ref{appendix:related_works}.}
}

\section{IL from heterogeneous demonstrators}
In this section, we formally formulate the problem of IL from heterogeneous demonstrators and propose our method.

\subsection{Problem Setup}
In reinforcement learning \cite[RL;][]{sutton1998reinforcement}, the optimal policy maximizes the expected cumulative reward: $\pi^* = \arg\!\max_\pi \mathbb{E}_{p_\pi}\left[\sum_{t=1}^T\gamma^t r(s_t, a_t)\right],$ where $r(s_t, a_t)$ is the reward associated with the state-action pair $(s, a)$ at time $t$, $p_\pi$ is the trajectory distribution induced by policy $\pi$, and $\gamma$ is the discount factor.
Unlike RL, in IL the reward function is unobserved and learning relies on a dataset of demonstrator trajectories $\tau = \{(s_0,a_0), ..., (s_{T_{\tau}},a_{T_{\tau}})\}$ of varying lengths $T_{\tau}$.
The goal is to learn a policy $\pi$ that matches the optimal policy's state-action distribution.

To make this feasible, most IL algorithms assume that \emph{all} demonstrations are generated by the optimal policy.
However, this assumption is restrictive as real datasets are often collected from multiple demonstrators with varying skills.
Therefore, in this work, we study IL from heterogeneous demonstrators with different and unknown expertise. 

\textbf{Dataset:}
We consider \(m\) demonstrators.
Each demonstrator $i\in[m]$ provides a set of trajectories $D_i$ and is associated with an \emph{expertise level} $\alpha_i\in [0, 1]$.
We define $\alpha_i$ as the probability that any given \((s, a)\) pair from $D_i$ was generated by the optimal policy, i.e., $\alpha_i=\mathbb{P}[(s,a)\sim\pi^*|(s,a)\in D_i]$.
This represents a demonstrator-level attribute: a higher \( \alpha_i \) means demonstrator $i$ tends to act more consistently w.r.t. $\pi^*$.
We consider \emph{demonstrator heterogeneity}, i.e., $\exists~i,j\in[m], i\neq j$, such that $\alpha_i\neq \alpha_j$.
The full dataset is $D = \{(i,D_i)\}_{i=1}^m$, and the expertise levels $\{\alpha_i\}_{i=1}^m$ are \emph{unknown}.

\textbf{Optimality Score:}
For each \((s, a)\) pair, we introduce a binary latent variable \(z \in \{0,1\}\) indicating its \emph{optimality}: \(z = 1 \Rightarrow (s,a)\sim \pi^*\), while \(z = 0\Rightarrow(s,a)\sim \pi^{\mathrm{sub}}\), where $\pi^{\mathrm{sub}}$ denotes an unknown suboptimal policy. 
The \emph{optimality score} is defined as the probability that $(s,a)$ is optimal given its values: $\mathbb{P}(z=1|s,a)$, which can be seen as a sample-specific, conditional expertise level.
\footnote{Our expertise modeling is twofold: $\alpha_i$ represents a demonstrator's overall skill (general expertise), while the optimality score captures state-action specific expertise (in-context expertise).}

Let $\rho^{\pi^*}(s)$ and $\rho^{\pi^\mathrm{sub}}(s)$ denote the state visitation distribution under $\pi^*$ and $\pi^{\mathrm{sub}}$, respectively.
To capture demonstrator heterogeneity, we model demonstrations from each demonstrator-\(i\) as a statistical mixture%
\footnote{We note that $\rho_i(s,a)$ is a \emph{statistical} mixture over demonstrator-generated samples, analogous to the mixing proportions in \citet{lu2018minimal, lu2020mitigating}, rather than the occupancy induced by executing $\pi_i$ in the MDP.
In particular, its marginal $\rho_i(s)=\sum_a \rho_i(s,a)=\alpha_i \rho^{\pi^*}(s)+(1-\alpha_i)\rho^{\pi^{\mathrm{sub}}}(s)$ generally differs across demonstrators and is \emph{not} equal to the MDP visitation distribution $\rho^{\pi_i}(s)$. This mixture model is used only to describe demonstration generation process, not to define the learner’s policy.}
$$\rho_i(s,a) = \alpha_i \rho^{\pi^*}(s)\,\pi^*(a\mid s)+(1-\alpha_i)\rho^{\pi^\mathrm{sub}}(s)\pi^{\mathrm{sub}}(a\mid s)$$
where $\rho^{\pi}(s,a)=\rho^{\pi}(s)\,\pi(a\mid s)$; $\rho^{\pi^*}(s,a)$ and $\rho^{\pi^\mathrm{sub}}(s,a)$ are the occupancy measures of the optimal and suboptimal policies.
This formulation corresponds to each demonstrator generating samples by first drawing
\(z \sim \mathrm{Bernoulli}(\alpha_i)\) and then sampling \((s,a) \sim P(s,a \mid z)\).
It ensures that all demonstrators share the same optimal and suboptimal occupancy components,
differing only in expertise level $\alpha_i$.


\textbf{Policy Learning:}
Given demonstrations $D$ from these heterogeneous demonstrators, our goal is to match the optimal policy $\pi^*$.
{Without access to rewards, we learn a scoring model parameterized by $\phi$, with logit $g_\phi:\mathcal{S}\times\mathcal{A}\to\mathbb{R}$ and optimality probability $f_\phi(s,a)=\sigma(g_\phi(s,a))=\mathbb{P}(z=1\mid s,a)$. 
We use the negative log-complement transformation
$r_\phi(s,a)=-\log(1-f_\phi(s,a)) =\operatorname{softplus}(g_\phi(s,a))$
as the saturating surrogate reward in RL.
$\hat{\pi}^*=\arg\!\max_\pi \mathop{\mathbb{E}_{p_\pi}}\left[\sum_{t=1}^{T}\gamma^t r_\phi(s_t,a_t)\right]$.}

\subsection{Joint estimation of expertise and scoring model}
We now describe how to learn the scoring model $f_\phi$ directly from the heterogeneous dataset $D$.

\subsubsection{Surrogate demonstrator classification for optimality scoring}
\label{sec:uu_method}
The optimality scoring task can be framed as a binary \emph{class probability estimation} \cite{buja2005loss} problem, where we aim to estimate $\mathbb{P}(z = 1|s, a)$ for each $(s, a)$ pair being optimal (with label $z=1$) or {sub}-optimal (with label $z=0$).
In the supervised setting, if explicit optimality labels $z$ were available, one could directly train $f_{\phi}$ with a proper scoring rule, such as cross-entropy.
However, in our setting, such labels are not available.
To address this challenge, we adapt the surrogate learning framework from the surrogate set classification (SSC) method of \citet{lu2021binary}.

\paragraph{Surrogate Demonstrator Classification}
We use the demonstrator identity $\bar{z} \in [m]$ as a surrogate label. 
Then the surrogate task involves training a multi-class classifier $\overline{f}_{\phi}:\mathcal{S}\times\mathcal{A}\to\mathbb{R}^m$ to predict the demonstrator from which each $(s,a)$ pair originated, by minimizing the risk:
\begin{align}
\label{eq:ssc-risk}
    \mathcal{L}(\overline{f}_\phi)=\mathbb{E}_{(s,a)\sim \sum_i\mu_i\rho_i(s,a)}[\ell(\overline{f}_{\phi}(s,a),\bar{z})],
\end{align}
where $\mu_i$ is the empirical prior probability of sampling from demonstrator $i$, estimated as $\mu_i:=\frac{{|D_i|}}{\sum_{i=1}^m {|D_i|}}$, {where $|D_i|$} denotes the cardinality of the dataset $D_i$, and $\ell$ is the cross-entropy loss.
For simplicity, we denote $\rho:=\sum_i\mu_i\rho_i(s,a)$.

\paragraph{Recovering Optimality Scores}
Under demonstrator heterogeneity, \citet{lu2021binary} show that the optimality scoring model $f_{\phi}$ can be recovered from the surrogate classifier $\overline{f}_{\phi}$ via an injective transformation $T(\cdot): \bR \rightarrow{\bR^m}$, such that: $\overline{f}_{\phi} = T_i(f_{\phi})$,%
\footnote{With a slight abuse of notation, we use $f_{\phi}$ and $\overline{f}_{\phi}$ to refer to both classifiers and class probability estimators, i.e., scoring functions in the sense of \citet{buja2005loss}.}
where the $i$-th component of $T$ is
\begin{align}
\label{eq:ssc-transformation}
    T_i (x) & = \frac{\mu_i (\alpha_i - \alpha') x + \mu_i \alpha' (1 - \alpha_i)}{ \sum_{i=1}^m \mu_i (\alpha_i - \alpha') x + \sum_{i=1}^m \mu_i \alpha' (1-\alpha_i)},
\end{align}
Here, $\alpha_i=\mathbb{P}(z=1|\bar{z}=i)$ is the expertise level of demonstrator $i$, and $\alpha'$ is the expected expertise in the target environment where the scoring model is deployed.
If these expertise levels are known, $f_\phi$ can be learned indirectly by training $\overline{f}_{\phi}$ via Eq.~\ref{eq:ssc-risk}, which induces updates on $f_\phi$ through the following equivalent objective
\begin{align}
     \mathcal{L}(\phi) =  \mathbb{E}_{(s,a)\sim \rho}\left[\ell(T(f_\phi(s,a)), \bar{z})\right].
     \label{eq:ssc-risk-with-T}
\end{align}

\begin{lemma}[Optimal Scoring Model Recovery, \citet{lu2021binary}]
\label{lemma:ssc-recovery}
Assume the model class is well-specified, $\ell$ is a proper loss (e.g., cross-entropy), and $T$ is correctly computed as in Eq. \ref{eq:ssc-transformation}.  
Let $\overline{f}_{\phi}$ be the surrogate classifier trained optimally according to Eq. \ref{eq:ssc-risk-with-T}.  
Then the induced binary scoring model $f_\phi$ accurately estimates the optimality scores:
$f_{\phi}(s,a) = \mathbb{P}(z=1 \mid s,a)$.
\end{lemma}

{
An exact optimality scoring model identifies the optimal component of the
demonstration distribution through a density-reweighting relation. We use $r_\phi(s,a) = -\log(1-f_\phi(s,a))$
as the surrogate reward for policy learning. Although this transformation is
monotonically increasing in the optimality posterior and therefore preserves
the ordering of state--action pairs, maximizing it pointwise does not in
general guarantee recovery of the full optimal occupancy. In particular, it
may favor policies that concentrate occupancy on a subset of highly rewarded
state--action pairs. We therefore distinguish optimality-score identification
from policy recovery. The following theorem provides a sufficient condition
under which maximizing the transformed surrogate reward recovers the optimal
occupancy.

\begin{theorem}[Conditional Optimal Occupancy Recovery]
\label{prop:optimal-policy}
Let $\pi^\ast$ be the unknown optimal policy and assume that $\pi^\ast$
belongs to the feasible learner policy class. Let $f_\phi(s,a) = \mathbb{P}(z=1\mid s,a)$ be the exact optimality posterior, and define $r_\phi(s,a) = -\log(1-f_\phi(s,a))$.
Let $\rho$ denote the demonstration distribution and assume that the
occupancy measure of every feasible learner policy is supported on
$\operatorname{supp}(\rho)$. Assume further that $f_\phi(s,a)<1$ on the
relevant support so that $r_\phi(s,a)$ is finite.
Define $J_r(\pi) = \sum_{s,a} \rho^\pi(s,a)r_\phi(s,a)$.
Let $\mathcal{X}_\ast = \operatorname{supp}(\rho^{\pi^\ast})$.
Suppose that, for every feasible learner policy $\pi$,
\begin{equation}
\sum_{(s,a)\in\mathcal{X}_\ast}
\frac{\rho^\pi(s,a)^2\,r_\phi(s,a)}
     {\rho^{\pi^\ast}(s,a)}
\leq
\sum_{(s,a)\in\mathcal{X}_\ast}
\rho^{\pi^\ast}(s,a)r_\phi(s,a).
\label{eq:reward-weighted-occupancy-condition}
\end{equation}
Then $\pi^\ast$ is a global maximizer of $J_r$.
Moreover, any global maximizer $\hat{\pi}\in\arg\max_\pi J_r(\pi)$
satisfies $\rho^{\hat{\pi}}(s,a)=\rho^{\pi^\ast}(s,a)$ for all $(s,a)$. Consequently, for every state $s$ such that $\rho^{\pi^\ast}(s)>0$, $\hat{\pi}(a\mid s) = \pi^\ast(a\mid s)$.
\end{theorem}

Theorem~\ref{prop:optimal-policy} is deliberately conditional.
Eq. \ref{eq:reward-weighted-occupancy-condition} bounds the
reward-weighted concentration of any feasible learner occupancy relative to
the optimal-policy occupancy. It rules out policies that obtain a larger
surrogate return simply by concentrating occupancy on a subset of
high-reward state--action pairs.
}

\begin{algorithm}[t]
\caption{{EM-style Joint Learning of Expertise Levels and Scoring Model (Offline)}} 
    \label{alg:1}
\begin{algorithmic}[1]
\State \textbf{Input}: $D = \{(i, D_i)\}^m_{i=1}$, number of queries $n$
\Repeat
\State \textbf{Initialization}: Scoring model $f_\phi(s, a) $, expertise levels $\{ \alpha_i \}_{i=1}^m \sim \mathcal{U}(0.1,\,0.9),~\forall i\in [m]
 $
\For{$t = 1,2,..., T $}

    \State {\textbf{Scoring-model update (minimize surrogate risk):}}
    \State Update \(f_\phi\) by minimizing the empirical risk from Eq.~\ref{eq:ssc-risk-with-T} using $\{ \alpha_i \}_{i=1}^m$
    \State {\textbf{Expertise-level update (estimate $\{ \alpha_i \}_{i=1}^m$):}}
    \State {Update expertise levels $\alpha_i \leftarrow \frac{ \sum_{(s,a) \in D_i} \mathbf{1}\!\left\{f_\phi(s,a)>0.5\right\}}{|D_i|},~\forall i\in[m]$}
\EndFor

\Until {\textbf{Expertise Dispersion Criterion} is satisfied, i.e., $\mathrm{Var}(\{\alpha_i\}_{i=1}^m) \geq \epsilon$}
\State Query $n$ samples and correct predictions to satisfy \textbf{Optimality Alignment Criterion}


\State \textbf{Return} $\{ \alpha_i \}_{i=1}^m$, $f_\phi$
\end{algorithmic}
\end{algorithm}

\subsubsection{{EM-style alternating optimization of scoring model and expertise estimates}}
\label{sec:em_style_optimization}
However, neither optimality labels nor expertise are given in our setting.
{To jointly learn both, we propose an EM-style alternating self-training method (Algorithm~\ref{alg:1}).}\footnote{{To make the expertise level estimation problem solvable, we assume the existence of anchor points in the demonstration dataset, i.e., data points that belong to class $z=1$ and $z=0$ almost surely.}}
{Here, \emph{EM-style} refers to the alternating updates of the scoring model and demonstrator expertise estimates.}

The algorithm begins by randomly initializing the expertise levels $\{\alpha_i\}_{i=1}^m$ within the range $[0.1, 0.9]$. Given the initial $\{\alpha_i\}_{i=1}^m$, we update the scoring model $f_\phi$ using the surrogate objective Eq. \ref{eq:ssc-risk-with-T}. 
We treat the expected expertise level $\alpha'$ of the target environment as a hyperparameter and fix it to $0.5$. This choice corresponds to evaluating the scoring model using the \emph{balanced risk}, a criterion adopted in \citet{tangkaratt2020robust} and shown to be robust to label noise \cite{charoenphakdee2019symmetric, menon2015learning}. We further justify this choice through ablation studies presented in Supplementary Tables~\ref{tab:antv4_pi_test_results_acc_prec}--\ref{tab:alpha_ablation} and Supplementary Fig.~\ref{fig:abl_testPi}.
{Next, the current scoring model produces hard pseudo-labels $\hat z(s,a)=\mathbf{1}\!\left\{f_\phi(s,a)>0.5\right\}$. Interpreting $f_\phi(s,a)$ as the posterior probability $\mathbb{P}(z=1\mid s,a)$, this threshold corresponds to the Bayes-optimal classifier under symmetric $0$--$1$ loss. We then estimate each $\alpha_i$ as the empirical fraction of samples from demonstrator $i$ assigned $\hat z=1$.}
{This alternating self-training procedure is repeated for $T$ epochs.}

{To reduce sensitivity to the initialization of $\{\alpha_i\}^m_{i=1}$, we run the algorithm with multiple random initializations until the Expertise Dispersion Criterion is satisfied. We report the number of initializations required in Supplementary Table~\ref{tab:repeat_num}.}




\paragraph{Expertise Dispersion Criterion} 
Since our algorithm trains on self-labeled data, it is susceptible to error propagation and confirmation bias. In practice, some random initializations lead to degenerate solutions, where $f_\phi(s,a)$ becomes overly confident and the estimated expertise levels collapse to nearly identical values.
To assess the reliability of the learned expertise levels $\{\alpha_i\}_{i=1}^m$, we measure their \emph{dispersion} by computing the variance $\mathrm{Var}(\{\alpha_i\}_{i=1}^m)$.
If the variance falls below a small threshold $\epsilon > 0$, we consider the model to have failed in distinguishing between different expertise levels, yielding low-information, overfit predictions. Conversely, a higher variance suggests more informative and diverse estimates, suggesting the model has captured demonstrator heterogeneity. 
We therefore use this variance as a selection criterion to retain only solutions with sufficient dispersion in $\{\alpha_i\}_{i=1}^m$.
{If $\mathrm{Var}(\{\alpha_i\}_{i=1}^m)<\epsilon$, we restart Algorithm~\ref{alg:1} with a fresh random initialization; otherwise, we retain the learned model. We use $\epsilon=0.0005$ in all experiments (all hyperparameters are given in Appendix Table~\ref{tab:two_stage_params}).}

\paragraph{Optimality Alignment Criterion}
Due to the unsupervised nature of our setting, the learned scoring model may inadvertently invert the notion of optimality. In such cases, the estimated expertise levels $\alpha_i$ may effectively correspond to $1 - \alpha_i$. 

To detect and correct this issue, we query a small subset of state--action
pairs and ask an expert to verify their predicted labels.
We use five queries in the main experiments, while the ablation shows that
a single query performs comparably (Supplementary Tables~\ref{tab:acc_prec_stages_exp_ask_num}--\ref{tab:query_ablation} and Supplementary Fig.~\ref{fig:abl_expCheckNum}).

If the majority of results indicate a systematic inversion in predicted labels, we simply flip the predicted labels and correct the outputs as follows:
{\footnotesize
$\alpha_i \leftarrow 1 - \alpha_i, f_\phi(s,a)\leftarrow 1- f_\phi(s,a), \forall i \in [m], ~\forall (s,a)\in D.$
}

\begin{algorithm}[t]
\caption{Joint Learning of Scoring Model and Agent Policy with Demonstration Relabeling (Online)} 
\label{alg:joint_early_stop}
\begin{algorithmic}[1]
\State \textbf{Input}: Dataset $D = \{(i, D_i)\}_{i=1}^m$, learned scoring model $f_\phi(s,a)$, estimated expertise levels $\{\hat{\alpha}_i\}_{i=1}^m$, ratio of pseudo-optimal samples $k$
\State \textbf{Initialization}: $\text{early\_stop} \gets \text{False}$, agent policy $\pi_\theta$, $D_{\text{opt}} \leftarrow \{(s,a) \mid f_\phi(s,a) > 0.5\}$ 
\For{$t = 1,2,..., T$}
    \State Sample trajectories from agent: $D_{\pi_\theta} \gets \{ (s,a) \mid (s,a) \sim \pi_\theta \}$
    \State Update $f_\phi$ by minimizing the empirical risk (Eq.~\ref{eq:scoring_loss}) \label{alg2:scor_upd} 
    \If{not early\_stop}
        \State Relabel pseudo-optimal set: \
        {$D_{\mathrm{opt}} \leftarrow$ top-$\lfloor k|D|\rfloor$ $(s,a)$ samples in $D$ ranked by $f_\phi(s,a)$ (highest first)} 
        \label{alg2:relabel}
        \State Check \textbf{early stopping criterion} and update \text{early\_stop} if necessary \label{alg2:earlystop} 
    \EndIf

    \State {Update $\pi_\theta$ using RL with reward $r_\phi(s,a)=-\log(1-f_\phi(s,a))$} \label{alg2:agent_upd}
\EndFor

\State \textbf{Return} final policy $\pi_\theta$
\end{algorithmic}
\label{alg:2}
\end{algorithm}

\subsection{Imitation learning via scoring model}
After learning the scoring model $f_\phi$ and estimating the expertise levels $\{\alpha\}_{i=1}^m$, we perform IL.

\subsubsection{Progressive scoring model refinement via fine-grained sample selection}
The scoring model $f_\phi$ is initially trained only on demonstration data, limiting its generalization to learner trajectories due to covariate shift~\cite{chang2021mitigating}. To improve its robustness, we refine $f_\phi$ during agent learning using three complementary terms: \emph{demonstration discriminative loss} (Eq.~\ref{eq:ssc-risk-with-T}), reflecting the original training signal; \emph{agent-matching penalty}, discouraging the misclassification of learner-generated behaviors as optimal; and a \emph{mixup regularizer}~\cite{zhang2017mixup} which interpolates between agent and demonstrator behaviors to smooth the scoring boundary and stabilize training. The resulting objective is:

\begin{align}
\label{eq:scoring_loss}
{\min_{\phi}}\ 
& \widehat{\mathbb{E}}_{(s,a)\sim \rho}
{\big[\ell(T(f_\phi(s,a)),\bar{z})\big]} \nonumber + \widehat{\mathbb{E}}_{(s',a') \sim \rho^{\pi_\theta}}
{\big[-\log(1-f_\phi(s', a'))\big]} \nonumber \\
&+ {w_{\mathrm{mix}}} \widehat{\mathbb{E}}_{\substack{(s, a) \sim \hat{\rho}^{\pi^*} \\
(s',a') \sim \rho^{\pi_\theta}}}
\big[ \mathcal{L}_{\text{mixup}} (s,a,s',a') \big]
\end{align}
where $\widehat{\mathbb{E}}$ denotes empirical expectation; $\hat{\rho}^{\pi^*}$ represents pseudo-optimal data inferred from $D_{\text{opt}} \subseteq D$; and {$w_{\mathrm{mix}}>0$ denotes the mixup loss weight. We set $w_{\mathrm{mix}}=2$ for all of our experiments.} The mixup loss is
\begin{align*}
\mathcal{L}_{\text{mixup}}
= {-\lambda \log\!\big(f_\phi(\tilde{s},\tilde{a})\big)
- (1-\lambda)\log\!\big(1-f_\phi(\tilde{s},\tilde{a})\big)}, \\
\text{where} \ 
{(\tilde{s},\tilde{a})=\lambda(s,a)+(1-\lambda)(s',a')},
\quad \lambda\sim\mathcal{U}(0,1)
\end{align*}

Early in training, the novice agent tends to produce suboptimal behaviors. The agent-matching penalty discourages $f_\phi$ from assigning high scores to these early samples, thereby enhancing its ability to distinguish between agent and demonstrator behavior and reducing covariate shift.
In parallel, mixup acts as both data augmentation and regularization based on agent and expert data, yielding more stable and informative gradients, ultimately improving the effectiveness of policy learning in IL~\cite{orsini2021matters}.
The full training procedure is given in Algorithm~\ref{alg:joint_early_stop}.

To further refine $f_\phi$, we adopt a bootstrapping approach where $f_\phi$ is used to pseudo-label demonstration data $D$. At each iteration:
(1) we extract a pseudo-optimal subset $D_{\text{opt}} \subseteq D$ by applying a \emph{top-$k$ selection method}, {where we choose the $\lfloor k \cdot |D| \rfloor$ state--action pairs with the highest $f_\phi$ scores} ($\lfloor \cdot \rfloor$ denotes the floor function);
(2) we generate new agent trajectories $D_{\pi_\theta}$ to represent $\rho^{\pi_\theta} \ $; (3) we update $f_\phi$ using the loss in Eq.~\ref{eq:scoring_loss} (Algorithm \ref{alg:joint_early_stop} Line \ref{alg2:scor_upd}), computed over $D$, $D_{\pi_\theta}$, and $D_{\text{opt}}$; (4) we relabel $D_{\text{opt}}$ using the updated $f_\phi$ (Algorithm \ref{alg:joint_early_stop} Line \ref{alg2:relabel}).
This mutual reinforcement between scoring model refinement and pseudo-label improvement allows both components to jointly bootstrap toward higher fidelity.

\subsubsection{Policy learning with scoring model as a surrogate reward}
As training progresses, the agent begins to generate increasingly competent behaviors, which are nonetheless treated as suboptimal during scoring model refinement. 
{At this stage, the scoring model $f_\phi$ functions similarly to a discriminator in the GAIL framework \cite{ho2016generative}, struggling to distinguish between optimal and agent-generated data.
Despite this, our approach is fundamentally different from GAIL, which relies exclusively on optimal demonstrations: our method operates in an (almost) unsupervised setting, allowing $f_\phi$ to learn from offline suboptimal data from heterogeneous demonstrators. This makes our approach especially effective when demonstration quality is low.}

As the agent improves, $f_\phi$ becomes increasingly confused, and its predictions become unreliable. To prevent the accumulation of incorrect labels that could mislead policy learning, we introduce an \textbf{early stopping criterion} for the relabeling process:
let $\Delta_t$ denote the number of sample changes in $D_{\text{opt}}$ between iterations $t$ and $t-1$, counting both removed and newly added samples; if $\Delta_t < \delta$ for $p$ consecutive steps, we stop relabeling and freeze $D_{\text{opt}}$.
The updated $f_\phi$ is employed as a surrogate reward function for training the agent policy $\pi_\theta$ (Algorithm \ref{alg:joint_early_stop} Line~\ref{alg2:agent_upd}). {Specifically, letting $g_\phi(s,a)$ denote the scorer logit and $f_\phi(s,a)=\sigma(g_\phi(s,a))$, we use the surrogate reward $r_\phi(s,a)=-\log(1-f_\phi(s,a))=\operatorname{softplus}(g_\phi(s,a))$.} 
In practice, we update the policy using soft actor-critic \cite[SAC;][]{haarnoja2018soft}, alternating with scoring model refinement and pseudo-label updates.

\section{Experiments}

We test whether our algorithm can (1) train a scoring model on multiple sets of imperfect demonstration data collected from heterogeneous demonstrators in order to provide optimality labels and estimate demonstrator expertise levels, and (2) learn an optimal agent policy based on the scoring model and imperfect demonstrations.

To evaluate the robustness of our algorithm under widely varying and low demonstrator expertise, we design two experiments: (a) \textbf{General expertise test}, using demonstrations collected from varying expertise levels in a wide range $[0.1-0.9]$ (average optimal ratio 0.5), and (b) \textbf{Low expertise test}, with extremely low expertise levels in $[0.05-0.15]$ (average optimal ratio 0.1). 
Prior works mostly focused on cases where most demonstrations are optimal, with limited exploration of low-optimal-ratio settings, where their performance degrades significantly \cite{eysenbach2021replacing,wang2021learning,wang2023unlabeled}.
To address this gap, we specifically evaluate the low expertise setting.
In both tests, datasets are constructed from six demonstrators (each providing 5,000 samples) with uniformly distributed expertise levels (see Appendix Table~\ref{tab:exp_true}). We set the top-$k$ hyperparameter $k$ to match the average optimal ratio in each test. Code for our experiments is available at \url{https://github.com/ch-engpan/IL_imperfect}.

\begin{table*}[t]
\centering
\caption{Performance comparison with baselines, mean $\pm$ std of reward across last five checkpoints.}
\resizebox{0.99\textwidth}{!}{%
\begin{tabular}{l|cc|cc|cccc}
\toprule
Environment & Dataset & GAIL o.s. & BC & ILEED & GAIL & RIL & WGAIL & LEOS-IL (ours) \\
\midrule
\multicolumn{7}{l}{\textbf{General expertise test}} \\
\midrule
Ant-v4 & 4016.3 $\pm$ 2776.8 & 5660.4 $\pm$ 163.6 & 1341.8 $\pm$ 164.5 & 1428.2 $\pm$ 80.3 & 1387.2 $\pm$ 301.0 & 101.8 $\pm$ 118.0 & 1744.0 $\pm$ 919.1 & \textbf{6084.6 $\pm$ 146.9} \\
HalfCheetah-v4 & 9988.9 $\pm$ 5959.9 & 9278.2 $\pm$ 496.5 & 2318.1 $\pm$ 521.2 & 1526.2 $\pm$ 1020.4 & 3035.8 $\pm$ 1186.4 & 1631.6 $\pm$ 959.7 & 4336.2 $\pm$ 335.6 & \textbf{8751.2 $\pm$ 1797.9} \\
Swimmer-v4 & 190.7 $\pm$ 149.1 & 293.6 $\pm$ 28.2 & 21.3 $\pm$ 23.0 & 16.8 $\pm$ 20.5 & 76.6 $\pm$ 28.3 & 28.8 $\pm$ 16.3 & 115.6 $\pm$ 83.7 & \textbf{305.4 $\pm$ 15.7} \\
Walker2d-v4 & 3096.7 $\pm$ 2549.6 & 3703.0 $\pm$ 1011.9 & 308.4 $\pm$ 111.5 & 440.3 $\pm$ 313.0 & 948.6 $\pm$ 297.2 & 1187.2 $\pm$ 130.1 & 665.4 $\pm$ 330.2 & \textbf{3959.2 $\pm$ 671.4} \\
\midrule
\multicolumn{7}{l}{\textbf{Low expertise test}} \\
\midrule
Ant-v4 & 1816.4 $\pm$ 1726.3 & 5542.0 $\pm$ 168.1 & 1473.4 $\pm$ 97.7 & 1376.5 $\pm$ 62.1 & 1439.6 $\pm$ 202.0 & 2.2 $\pm$ 174.1 & 971.6 $\pm$ 591.0 & \textbf{5932.0 $\pm$ 335.2} \\
HalfCheetah-v4 & 5222.0 $\pm$ 3578.8 & 9194.2 $\pm$ 848.1 & 3753.9 $\pm$ 220.5 & 3767.6 $\pm$ 176.9 & 3162.8 $\pm$ 1417.2 & 964.2 $\pm$ 1100.1 & 3975.6 $\pm$ 369.9 & \textbf{7335.0 $\pm$ 1316.4} \\
Swimmer-v4 & 71.4 $\pm$ 89.5 & 244.4 $\pm$ 113.5 & 27.8 $\pm$ 11.6 & 34.0 $\pm$ 2.4 & 37.4 $\pm$ 10.5 & 44.6 $\pm$ 2.9 & 48.8 $\pm$ 53.7 & \textbf{195.4 $\pm$ 92.1} \\
Walker2d-v4 & 1062.8 $\pm$ 1546.5 & 3532.6 $\pm$ 782.2 & 436.6 $\pm$ 69.1 & 605.3 $\pm$ 94.5 & 737.8 $\pm$ 163.2 & 1140.0 $\pm$ 319.0 & 703.8 $\pm$ 291.2 & \textbf{2961.8 $\pm$ 339.8} \\
\bottomrule
\end{tabular}
}
\noindent\parbox{0.99\textwidth}{\footnotesize{\textit{Note:} Boldface denotes the highest mean among non-oracle methods. Paired $t$-tests ($\alpha=0.01$) confirm that LEOS-IL significantly outperforms all non-oracle baselines.}}
\label{tab:main_exp}
\end{table*}

\begin{table*}[t]
\centering
\caption{Estimation errors of Stage 1 scoring model for expertise levels (mean $\pm$ std).}
\label{tab:error_only}
\resizebox{0.9\columnwidth}{!}{%
\begin{tabular}{lccccc}
\hline
Experiment & Ant-v4 & HalfCheetah-v4 & Hopper-v4 & Swimmer-v4 & Walker2d-v4 \\
\hline
General expertise test & $(2.67 \pm 1.28)\times10^{-4}$ & $(2.00 \pm 2.00)\times10^{-5}$ & $(3.83 \pm 2.73)\times10^{-3}$ & $(2.00 \pm 2.00)\times10^{-5}$ & $(3.27 \pm 3.12)\times10^{-4}$ \\
Low expertise test & $(9.65 \pm 0.24)\times10^{-2}$ & $(3.67 \pm 0.15)\times10^{-2}$ & $(1.05 \pm 0.02)\times10^{-1}$ & $(2.81 \pm 0.11)\times10^{-2}$ & $(1.34 \pm 0.04)\times10^{-1}$ \\
\hline
\end{tabular}
}
\label{tab:expertise_est}
\end{table*}

\subsection{Environment and datasets}

We evaluate our algorithm against baseline methods on four MuJoCo environments \cite{todorov2012mujoco, towers2024gymnasium} and conduct ablation studies to assess the contributions of our proposed components.
Demonstrations are collected by training a policy with SAC with the true reward. An early-stage checkpoint  $\pi^\text{sub}$ provides suboptimal demonstrations, while the final checkpoint $\pi^*$ provides optimal demonstrations (see Appendix Table~\ref{tab:rl_policy} for the checkpoints' performance).
As in \citet{beliaev2022imitation, tangkaratt2020robust}, the imperfect dataset is mixture of state-action pairs from $\pi^*$ and $\pi^\text{sub}$. Unlike random behaviors, our suboptimal data comes from a real suboptimal policy, which reflects that experts may make mistakes, but their actions are still better than random actions. 
Using this approach, multiple imperfect demonstration sets $ D = \{(i, D_i)\}^m_{i=1} $ are collected.

\subsection{Comparison with Baselines}
\label{sec:mainexp}

Across the experiments, we benchmark our model against four IL baseline algorithms. 
(1) \textbf{GAIL o.s. (optimal subset)} \cite{ho2016generative}: GAIL trained on the purely optimal subset of demonstrations, serving as an oracle method;
(2) \textbf{GAIL}: GAIL trained on all (optimal and suboptimal) demonstrations;  
(3) \textbf{RIL} \cite{tangkaratt2020robust}: trains an optimality discriminator via pseudo-labeling with co-training;
(4) \textbf{WGAIL} \cite{wang2021learning}: estimates importance weights to emphasize near-optimal demonstrations;
(5) \textbf{ILEED}~\cite{beliaev2022imitation}: learns the policy in a offline training manner via maximum-likelihood;
(6) \textbf{BC}~\cite{pomerleau1988alvinn}.
For fair comparison, we adopt a consistent model architecture for discriminator/classifier/scoring model, and learner policy, modifying only the input and output layers to match the state and action spaces of each environment~\cite{arulkumaran2024pragmatic}. Moreover, all online baselines incorporate mixup regularization. All learner policies are updated using SAC with same hyperparameters. Training details are provided in Appendix Table~\ref{tab:training_paras}. We use 5 random seeds per condition and report the mean and standard deviation in all our results. For offline methods (ILEED and BC), for each random seed, we repeat the algorithm 10 times with random initializations and report the best result by selecting the run with the maximal likelihood.

The results are summarized in Table \ref{tab:main_exp} (see Appendix Fig.~\ref{fig:main_combined} for training curves, and Supplementary Table~\ref{tab:bc_ileed_scoring} for more detailed results of offline methods). In the general expertise test, our algorithm outperforms all baselines except GAIL o.s., achieving at least twice the rewards in all tasks. While GAIL o.s. attains comparable rewards, our algorithm still yields slightly higher rewards in Ant, Swimmer and Walker2d. This is attributed to the fact that our algorithm can achieve high classification performance on imperfect demonstrations (see Table \ref{tab:acc_prec_stages}), and, in contrast to GAIL o.s., is also capable of leveraging suboptimal data to perform better in the subsequent online IL stage.
In the low expertise test, the trends remain similar: our algorithm surpasses all baselines except GAIL o.s. However, decreased classification performance on imperfect demonstrations (see Table~\ref{tab:acc_prec_stages}) reduces overall rewards compared to the general expertise test. In this setting, GAIL o.s., which trains on purely optimal subset of demonstrations, achieves higher rewards in most tasks.

The low rewards achieved by other baselines in both experiments can be attributed to the several factors. For GAIL, which treats all demonstrations as optimal, its performance is limited when handling imperfect demonstrations. 
RIL assumes that most demonstrations are optimal, but in our experiments, at least half are suboptimal. Consequently, the discriminator’s classification risk  does not align with the risk of distinguishing between optimal and suboptimal demonstrations,
reducing the reliability of its predictions for policy learning. Moreover, even with co-training for pseudo-labeling, the overconfidence can only be partially mitigated; once error accumulation begins, it continues to degrade the discriminator's ability to predict optimality of demonstrations. 
WGAIL requires an optimal discriminator to compute importance weights, assuming the early-stage GAIL discriminator is near-optimal. Although early stopping is used for weight estimation, this assumption can fail due to the instability of adversarial learning.
{
ILEED is sensitive to state embedding quality and assumes uniform state coverage, making it prone to biased expertise estimates under sparse data. Being offline and MLE-based, it is vulnerable to distribution shift, and its tightly coupled joint optimization can lead to instability.
}

\begin{table*}[t]
\centering
\caption{Classification results of the scoring model on imperfect demonstrations for Stage 1 and Stage 2 (Accuracy / Precision, \%), with Stage 1 $\rightarrow$ Stage 2 improvement (\%) after relabeling.} 
\resizebox{0.99\textwidth}{!}{%
\begin{tabular}{lc|cc|cc|cc|cc}
\toprule
Test & Stage & Ant-v4 & Impr. & HalfCheetah-v4 & Impr. & Swimmer-v4 & Impr. & Walker2d-v4 & Impr. \\
\midrule
General & 1 & 99.91 / 99.89 & - & 100.00 / 100.00 & - & 100.00 / 100.00 & - & 99.92 / 99.90 & - \\
expertise & 2 & 99.89 / 99.89 & -0.02 / 0.00 & 100.00 / 100.00 & 0.00 / 0.00 & 99.94 / 99.94 & -0.06 / -0.06 & 99.94 / 99.94 & 0.02 / 0.04 \\
\midrule
Low & 1 & 90.33 / \textbf{51.78} & - & 96.33 / \textbf{73.68} & - & 97.19 / \textbf{78.84} & - & 86.39 / \textbf{43.01} & - \\
expertise & 2 & 99.65 / \textbf{98.23} & 10.32 / \textbf{89.71} & 99.98 / \textbf{99.91} & 3.79 / \textbf{35.60} & 97.58 / \textbf{87.91} & 0.40 / \textbf{11.50} & 98.48 / \textbf{92.40} & 13.99 / \textbf{114.83} \\
\bottomrule
\end{tabular}
}
\label{tab:acc_prec_stages}
\end{table*}

\begin{figure*}[t]
     \centering     \includegraphics[width=0.99\textwidth]{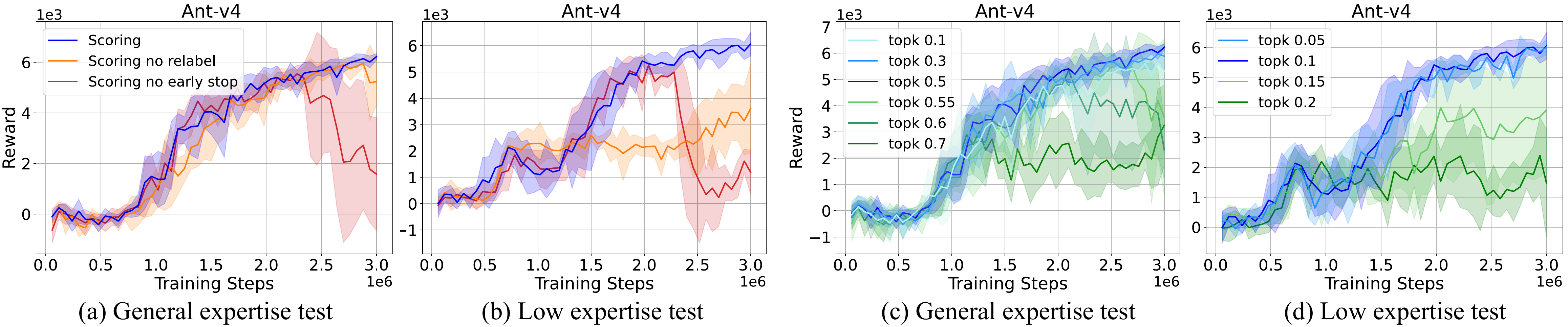}
     \caption{Ablation results on the impact of Stage 2 relabeling and early stopping, and  $k$ selection}
     \label{fig:abl_rel_early_topk}
\end{figure*}

\subsection{Estimation of expertise levels}
\label{sec:est_expertise}

Table~\ref{tab:expertise_est} reports the estimation errors of the scoring model trained in Stage 1 w.r.t. expertise levels relative to the ground truth, with detailed results provided in Appendix Tables~\ref{tab:expertise_est_details}-\ref{tab:acc_label_improvement}.
In the general expertise test, where true expertise levels are widely distributed within [0.1-0.9], errors are extremely low, remaining below 0.004 across all tasks. In contrast, in the low expertise test, with levels [0.05-0.15], the estimation errors increase to around 0.1 for most tasks. This indicates that the scoring model tends to be over-optimistic and produce more false positive (FP) labels on imperfect demonstrations (see Appendix Table \ref{tab:acc_label_improvement}). However, these incorrect labels are largely corrected during Stage 2.

\subsection{Optimality Prediction on Imperfection Demonstrations}
\label{sec:label_res}

We evaluate the scoring model $f_\phi$ on its ability to classify imperfect demonstrations by optimality. Labeling is performed in two stages: Stage 1 uses pretrained $f_\phi$ and Stage 2 uses fine-tuned $f_\phi$. 
Performance is measured using accuracy and precision, with a focus on precision, as it reflects the FP risk, which is more critical than false negative risk for policy learning \cite{irpan2019off}.

Results in Table~\ref{tab:acc_prec_stages} show that in the general expertise test, both accuracy and precision are nearly 1 in Stage 1, leaving almost no room for improvement in Stage 2. In contrast, in the low expertise test, where expertise levels are narrowly distributed, the accuracy in Stage 1 stays above 86\%, but the precision drops significantly, to 43-78\%. This indicates that surrogate demonstrator classification training handles datasets with widely varying expertise well, but struggles when expertise levels are similar. 
However, Stage 2 improves both accuracy and precision, with precision increasing by 12-115\% across tasks. 
This shows that for imperfect demonstrations with extremely low expertise, many false positives from a poorly pretrained $f_\phi$ can be corrected in Stage 2 after fine-tuning $f_\phi$. Detailed classification results are presented in Appendix Table \ref{tab:acc_label_improvement}.



\subsection{Ablation tests}
\label{sec:abl}

To assess the effect of each component of our algorithm, we conduct ablation experiments on the Ant-v4 environment.

\textbf{Relabeling and Early Stopping}:
We evaluate the effects of the relabeling step and early stopping in Stage 2 ({Algorithm~\ref{alg:2}}), shown in Fig.~\ref{fig:abl_rel_early_topk} (a)-(b) (see Appendix Table \ref{tab:ab_relabel_early} for numerical results). Without relabeling, performance is comparable to the original method in the general expertise test but drops by about half in the low expertise test due to FP errors generated in Stage 1 (see Sec. \ref{sec:label_res}), which cannot be corrected without Stage 2 relabeling. Removing early stopping degrades rewards in both general and low expertise tests as the agent approaches the optimal policy. Near-optimal agent samples confuse $f_\phi$, increasing classification errors and accumulating incorrect labels, which reinforces overconfidence of $f_\phi$ and corrupts the surrogate reward for the agent.


\textbf{Top-$k$ $k$ Value Selection:}
We study the effect of different $k$ values on IL performance .
Results are shown in Fig.~\ref{fig:abl_rel_early_topk} (c)-(d) and Appendix Table~\ref{tab:ab_topk}.
For both the general and low expertise tests, (average optimality ratios $k^*=0.5$ and $0.1$), rewards remain stable when $k \le k^*$ but decline for $k > k^*$. 
The reason is that {the pseudo-optimal set consists of the $\lfloor k \cdot |D|\rfloor$ state--action pairs with the highest $f_\phi$ scores}. Assuming $f_\phi$ is well trained, a $k < k^*$ leads to a conservative selection (subset of the true optimal set), while $k > k^*$ can introduce more FP samples to the pseudo optimal set. This indicates that FP risk is more critical, and a smaller $k$ is preferable.


\textbf{Number of Demonstrators:} We ablate the number of demonstrators and find that our method is robust to varying counts in terms of IL performance (see Supplementary Fig.~\ref{fig:abl_bagNum} and Supplementary Tables~\ref{tab:demo_num_expertise}--\ref{tab:demo_num_r}).
Supplementary Sec.~\ref{sec:more_analy} and supplementary Fig.~\ref{fig:analy_combined} show that our method consistently matches or outperforms variants (PN loss for discrimination, oracle settings with given expertise levels or optimality labels), highlighting its robustness and effectiveness.

\textbf{Number of human queries:}
We analyze sensitivity to the number of human queries. Results (Supplementary Tables~\ref{tab:acc_prec_stages_exp_ask_num}--\ref{tab:query_ablation} and Supplementary Fig.~\ref{fig:abl_expCheckNum}) show that even a single query performs comparably to five queries, due to the very high Stage 1 classification accuracy ($>$99\%) and similar reward performance across query settings.
Nevertheless, queries remain necessary to assign correct labels to clusters in the absence of ground-truth annotations.

\textbf{Expected Expertise $\alpha'$:} We analyze the effect of different values of $\alpha'$ (Supplementary Tables~\ref{tab:antv4_pi_test_results_acc_prec}--\ref{tab:alpha_ablation} and Supplementary Fig.~\ref{fig:abl_testPi}). Performance peaks near $\alpha' = 0.5$ and drops when $\alpha'$ is very small or large, indicating that $\alpha' = 0.5$ is optimal for both the general and low expertise tests.



\section{Conclusions}
We presented a general IL framework for handling unlabeled demonstrations from heterogeneous imperfect demonstrators with unknown expertise. Our approach combines {EM-style alternating self-training with an online refinement step} to learn a scoring model that guides policy learning. Experiments across challenging settings demonstrate strong performance, especially in low-expertise regimes, highlighting the value of our method's ability to leverage suboptimal data in an (almost) unsupervised manner.


\section*{Acknowledgements}
This work was supported by the European Union’s Horizon 2020 research, innovation programme under the Marie Skłodowska-Curie grant agreement No. 945363, and JST under Moonshot R\&D Grant Number JPMJMS2012.

\bmhead{Appendix}

{The appendices provide additional details on surrogate demonstrator classification training, the proof of Theorem~\ref{prop:optimal-policy}, detailed experimental results, and reproducibility details, including Algorithm~\ref{alg:ssc}, Fig.~\ref{fig:main_combined}, and Tables~\ref{tab:rl_policy}--\ref{tab:training_paras}.}

\bmhead{Supplementary Information}

Supplementary information comprises: (1) the supplement containing further discussion of related work, detailed experimental and ablation results, and additional analyses of method variants, including Supplementary Figs.~\ref{fig:abl_bagNum}--\ref{fig:abl_testPi} and Supplementary Tables~\ref{tab:demo_num_expertise}--\ref{tab:bc_ileed_scoring}; and (2) the accompanying code, available at \url{https://github.com/ch-engpan/IL_imperfect}.
          
\backmatter

\clearpage
\begin{appendices}
\counterwithin*{figure}{section}
\counterwithin*{table}{section}
\counterwithin*{algorithm}{section}
\counterwithin*{equation}{section}
\renewcommand{\thefigure}{\thesection\arabic{figure}}
\renewcommand{\thetable}{\thesection\arabic{table}}
\renewcommand{\thealgorithm}{\thesection\arabic{algorithm}}
\renewcommand{\theequation}{\thesection\arabic{equation}}

\section{Surrogate Demonstrator Classification training}
\label{apdx:ssc_learning}

We provide more details about the algorithm and theory of Surrogate Demonstrator Classification training.

\begin{algorithm}[!htp]
	\caption{Learning scoring model} \label{alg_uu}
\begin{algorithmic}[1]
\State \textbf{Input}: $\{ (i,  D_i)  \}^m_{i=1}$, $\{ \alpha_i  \}^m_{i=1}$ , and $\alpha'$
\State \textbf{Initialization}: scoring model $f_\phi(s, a)$\\
\State  Compute of $ T(\cdot) =  [ T_1(\cdot), \cdots ,  T_m(\cdot) ] $ using Eq. \ref{eq:ssc-transformation} based on $\{ \alpha_i  \}^m_{i=1}$ and $\alpha'$
\For{$t = 1,2, \cdots $}
\State Sample examples: $ \bar{z} , (s,a) \sim \{ (i, D_i)  \}^m_{i=1}  $
\State Update $f_\phi$ by minimizing the empirical risk from Eq.~\ref{eq:ssc-risk-with-T} 
\EndFor
\State \textbf{Return} $f_\phi$ 
\end{algorithmic}
\label{alg:ssc}
\end{algorithm}

We consider {demonstrator heterogeneity}, i.e., there $\exists~i,j\in[m], i\neq j$, such that $\alpha_i\neq \alpha_j$. According to \citet{lu2021binary}, the $f_{\phi^*_{\text{surr}}}$ that is recovered by removing transformation layer $T(\cdot)$ of the optimal surrogate classifier $\overline{f}_{\phi^*}$ (the minimizer of $\mathbb{E}_{(s,a)\sim \rho}[\ell(\overline{f}_{\phi}(s,a),\bar{z})]  =  \mathbb{E}_{(s,a)\sim \rho}\left[\ell(T(f_\phi(s,a)), \bar{z})\right]$) is identical to the optimal classifier ${f_{\phi^*}}$ that minimizes the risk $\bE_{(s, a) \sim \rho}\left[ \ell(f_\phi(s,a), z)\right]$, which is not possible if the heterogeneity setting is invalid. Thus, through the transformation function transformation $T(\cdot)$, the optimality scoring model $f_{\phi^*}$ can be learned by minimizing the surrogate classification risk in Eq. \ref{eq:ssc-risk-with-T}. 

{
\section{Proof of Theorem~\ref{prop:optimal-policy}}
\label{apdx:proof_optimal_policy}

\begin{proof}
Let $c=\mathbb{P}(z=1)\in(0,1)$ denote the optimality prior in the demonstration distribution. Since
$f_\phi$ is the exact optimality posterior, Bayes' rule gives, for every
$(s,a)\in\operatorname{supp}(\rho)$,
\begin{equation}
f_\phi(s,a)
=
c\frac{\rho^{\pi^\ast}(s,a)}{\rho(s,a)}.
\label{eq:posterior-density-ratio}
\end{equation}
Hence $r_\phi(s,a) = -\log\left(1-c\frac{\rho^{\pi^\ast}(s,a)}{\rho(s,a)}
\right)$.
In particular, if $(s,a)\notin\mathcal{X}_\ast$, then
$\rho^{\pi^\ast}(s,a)=0$, so $f_\phi(s,a)=0$ and therefore
$r_\phi(s,a)=0$. Thus, for any feasible policy $\pi$, $$J_r(\pi)=\sum_{(s,a)\in\mathcal{X}_\ast}
\rho^\pi(s,a)r_\phi(s,a).$$
Define vectors indexed by
$(s,a)\in\mathcal{X}_\ast$ as $x^\pi_{s,a} = \rho^\pi(s,a)\sqrt{\frac{r_\phi(s,a)}{\rho^{\pi^\ast}(s,a)}}$,
and $y_{s,a} = \sqrt{ \rho^{\pi^\ast}(s,a) r_\phi(s,a)}$.
Then $J_r(\pi) = \langle x^\pi,y\rangle$.
By the Cauchy--Schwarz inequality,
\begin{align}
J_r(\pi)
&\leq
\|x^\pi\|_2\|y\|_2.
\label{eq:reward-cs}
\end{align}
From
Eq.~\ref{eq:reward-weighted-occupancy-condition},
\[
\|x^\pi\|_2^2
=
\sum_{(s,a)\in\mathcal{X}_\ast}
\frac{\rho^\pi(s,a)^2r_\phi(s,a)}
     {\rho^{\pi^\ast}(s,a)}
\leq
\sum_{(s,a)\in\mathcal{X}_\ast}
\rho^{\pi^\ast}(s,a)r_\phi(s,a)
=
\|y\|_2^2.
\]
Therefore, $J_r(\pi)
\leq
\|x^\pi\|_2\|y\|_2
\leq
\|y\|_2^2$.
For $\pi=\pi^\ast$, $x^{\pi^\ast}_{s,a}
=
y_{s,a}$,
and hence $J_r(\pi^\ast)
=
\|y\|_2^2$.
It follows that $J_r(\pi)
\leq
J_r(\pi^\ast)$,
$\forall\pi$,
so $\pi^\ast$ is a global maximizer.
Now let $\hat{\pi}\in\arg\!\max_\pi J_r(\pi)$.
Since $J_r(\hat{\pi})
=
J_r(\pi^\ast)
=
\|y\|_2^2$,
both inequalities above must hold with equality. 
Equality in the
Cauchy--Schwarz inequality implies that there exists
$\beta\geq0$ such that $x^{\hat{\pi}}
=
\beta y$.
Therefore, $$\rho^{\hat{\pi}}(s,a)
=
\beta\rho^{\pi^\ast}(s,a), \quad \forall (s,a)\in\mathcal{X}_\ast.$$
Equality in
Eq.~\ref{eq:reward-weighted-occupancy-condition} further gives
$\|x^{\hat{\pi}}\|_2^2
=
\|y\|_2^2$.
Substituting
$x^{\hat{\pi}}=\beta y$ yields
$\beta^2\|y\|_2^2
=
\|y\|_2^2$,
and hence $\beta=1$. Thus,
$$\rho^{\hat{\pi}}(s,a)
=
\rho^{\pi^\ast}(s,a) ~~ \text{on} ~~ \mathcal{X}_\ast.$$
All feasible discounted occupancy measures have the same total mass,
\[
\sum_{s,a}\rho^\pi(s,a)
=
\sum_{t=1}^{T}\gamma^t
=:M.
\]
Since $\rho^{\hat{\pi}}$ already matches $\rho^{\pi^\ast}$ on
$\mathcal{X}_\ast$ whose total mass is $M$, $\hat{\pi}$ cannot place
additional occupancy outside $\mathcal{X}_\ast$. Therefore, $\rho^{\hat{\pi}}=\rho^{\pi^\ast}$.
Finally, for any state $s$ with
$\rho^{\pi^\ast}(s)>0$,
\[
\hat{\pi}(a\mid s)
=
\frac{\rho^{\hat{\pi}}(s,a)}
     {\rho^{\hat{\pi}}(s)}
=
\frac{\rho^{\pi^\ast}(s,a)}
     {\rho^{\pi^\ast}(s)}
=
\pi^\ast(a\mid s).
\]
Hence any global maximizer of the surrogate objective recovers the
optimal occupancy and matches the optimal policy on states visited by
$\pi^\ast$.
\end{proof}
}

\subsection{Detailed results}

\begin{figure}[h!]
    \centering

    \begin{subfigure}{\textwidth}
        \centering
        \begin{subfigure}{0.248\textwidth}
            \centering
            \includegraphics[width=\linewidth]{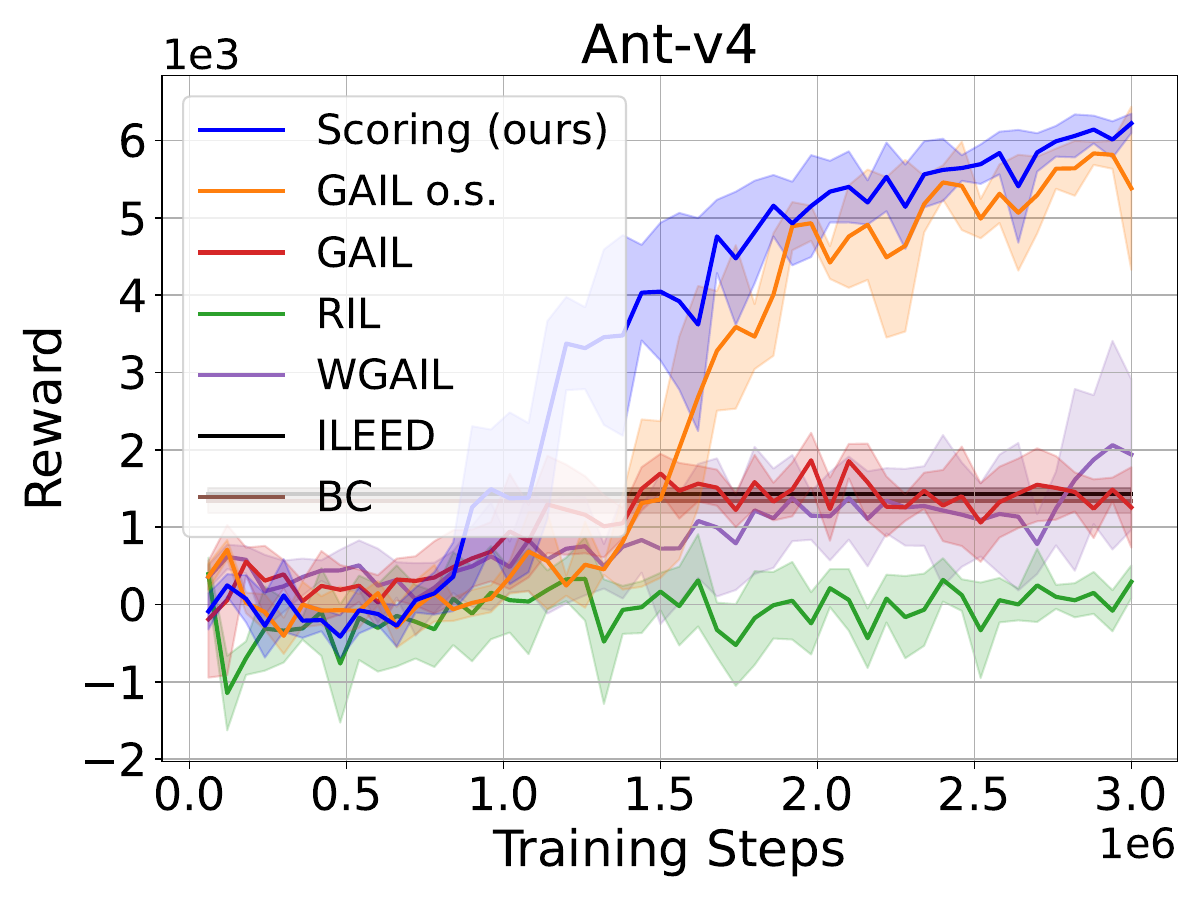}
        \end{subfigure}\hfill
        \begin{subfigure}{0.248\textwidth}
            \centering
            \includegraphics[width=\linewidth]{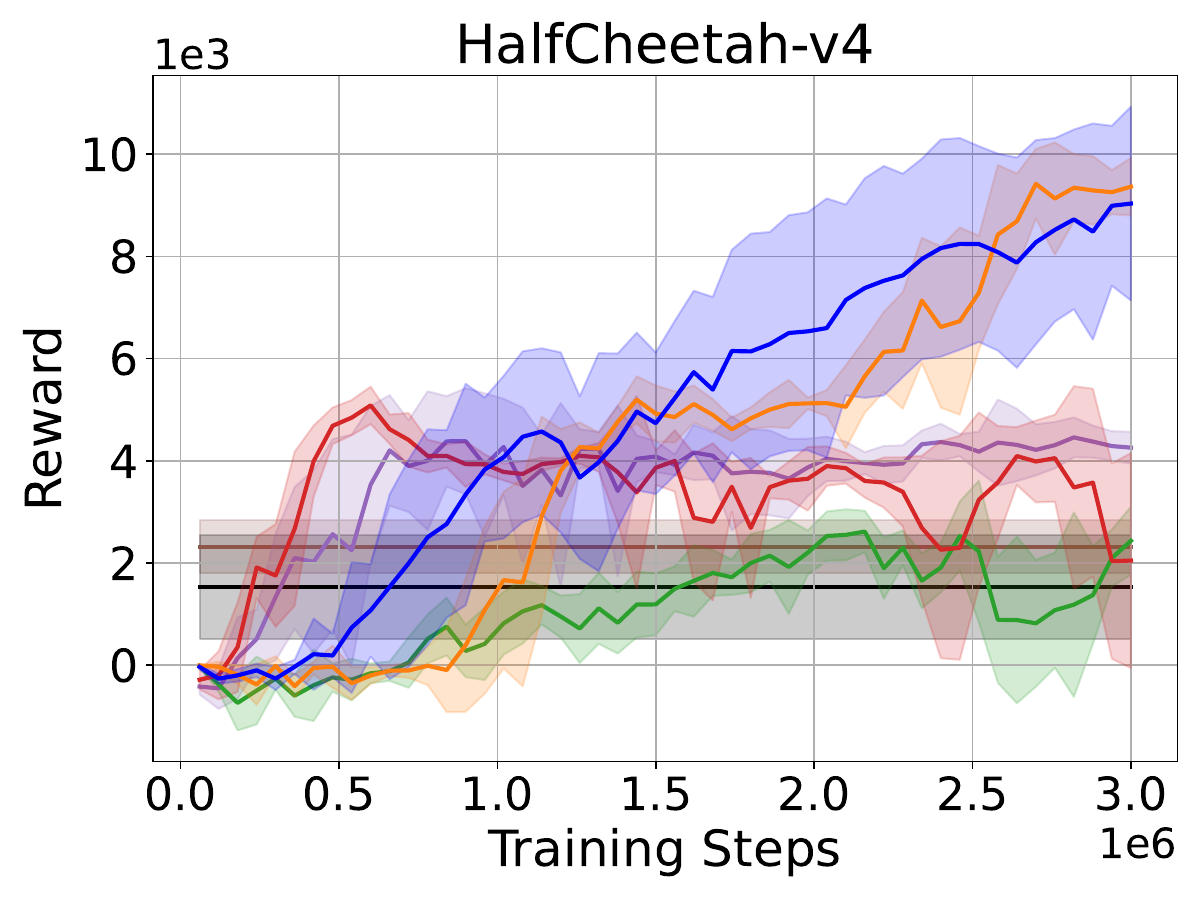}
        \end{subfigure}\hfill
        \begin{subfigure}{0.248\textwidth}
            \centering
            \includegraphics[width=\linewidth]{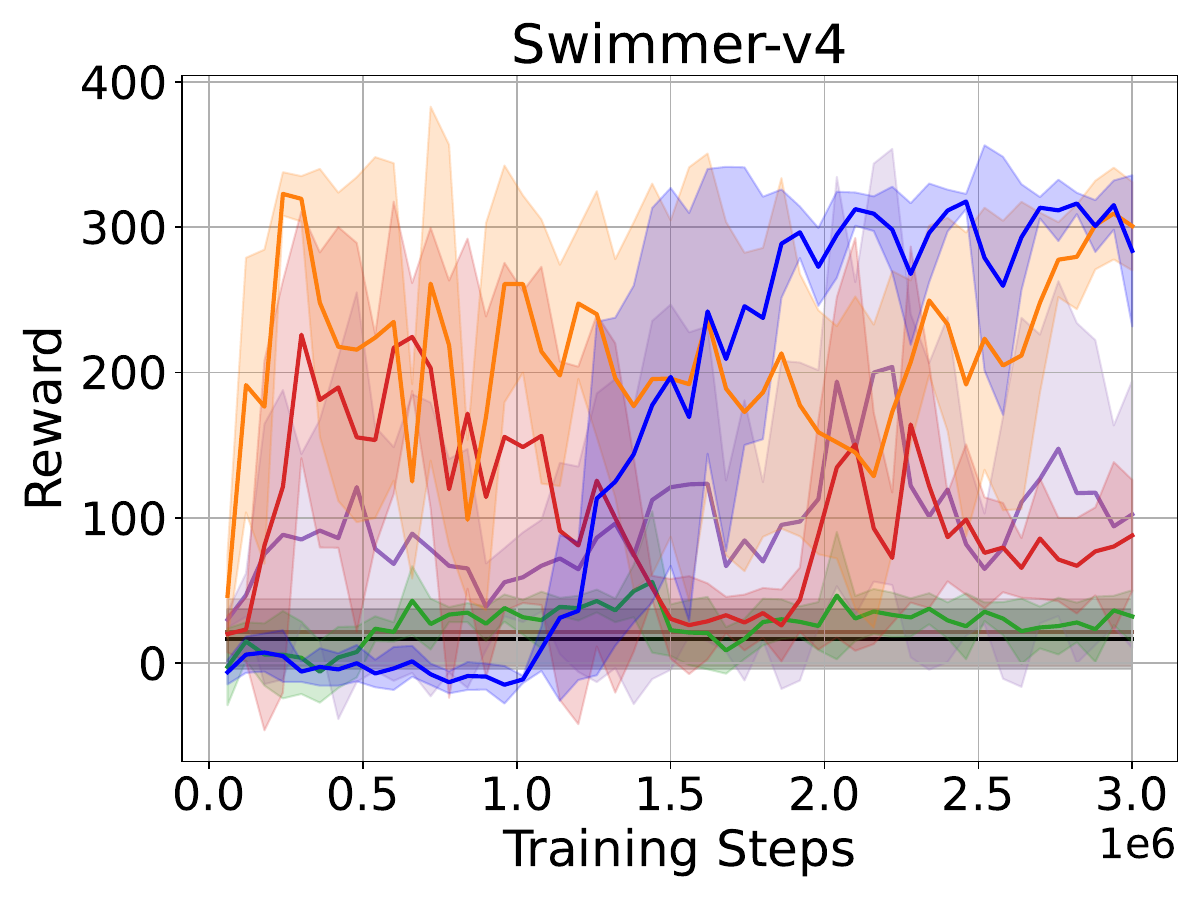}
        \end{subfigure}\hfill
        \begin{subfigure}{0.248\textwidth}
            \centering
            \includegraphics[width=\linewidth]{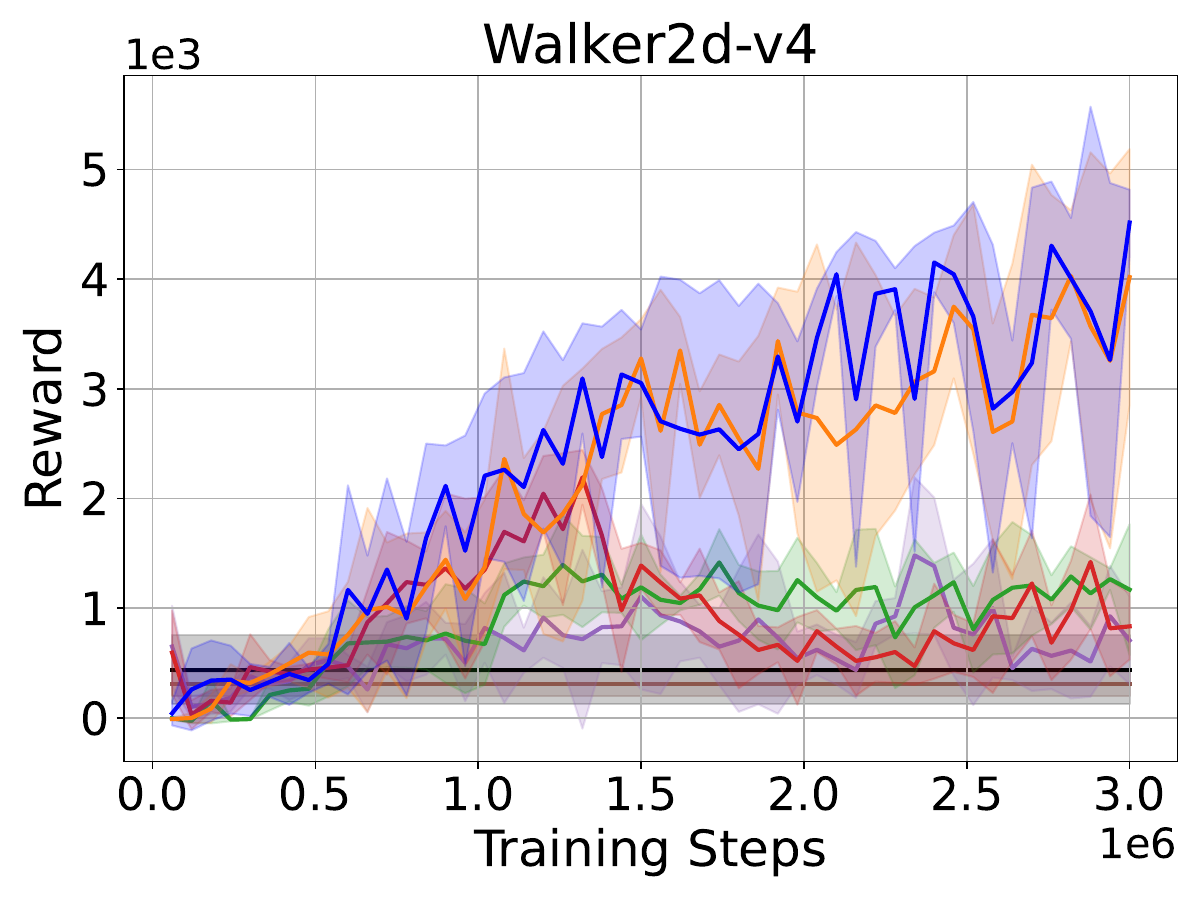}
        \end{subfigure}
        \caption{General expertise test}
        \label{fig:main1}
    \end{subfigure}

    \vspace{1em} 

    \begin{subfigure}{\textwidth}
        \centering
        \begin{subfigure}{0.248\textwidth}
            \centering
            \includegraphics[width=\linewidth]{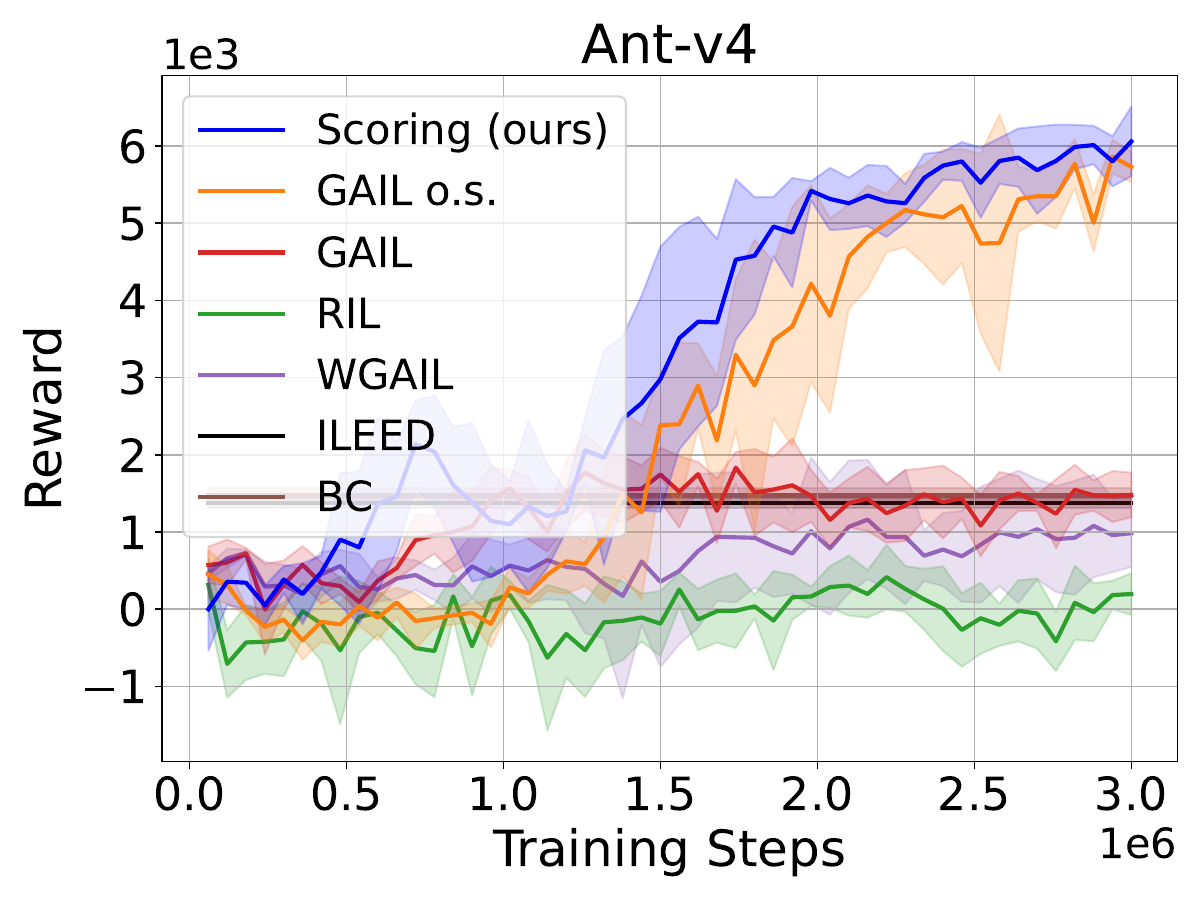}
        \end{subfigure}\hfill
        \begin{subfigure}{0.248\textwidth}
            \centering
            \includegraphics[width=\linewidth]{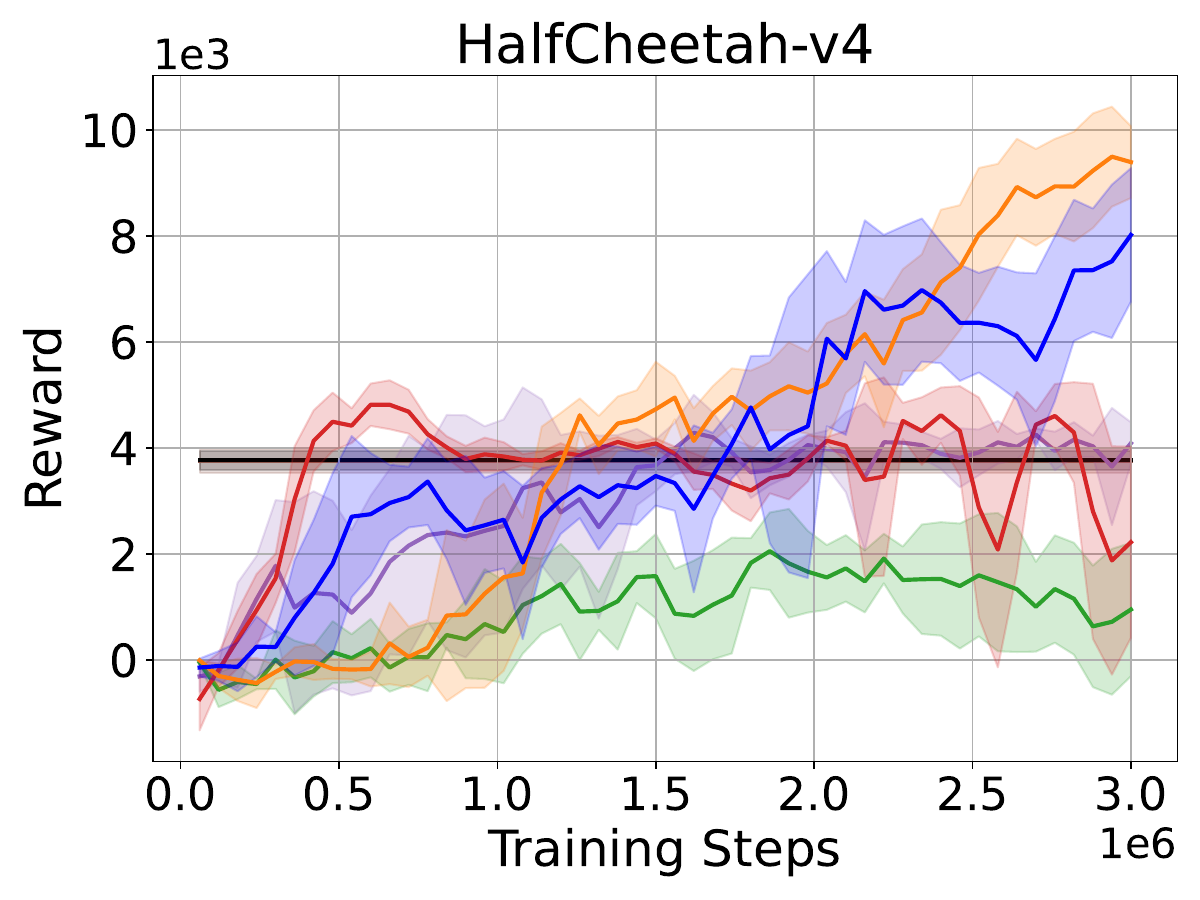}
        \end{subfigure}\hfill
        \begin{subfigure}{0.248\textwidth}
            \centering
            \includegraphics[width=\linewidth]{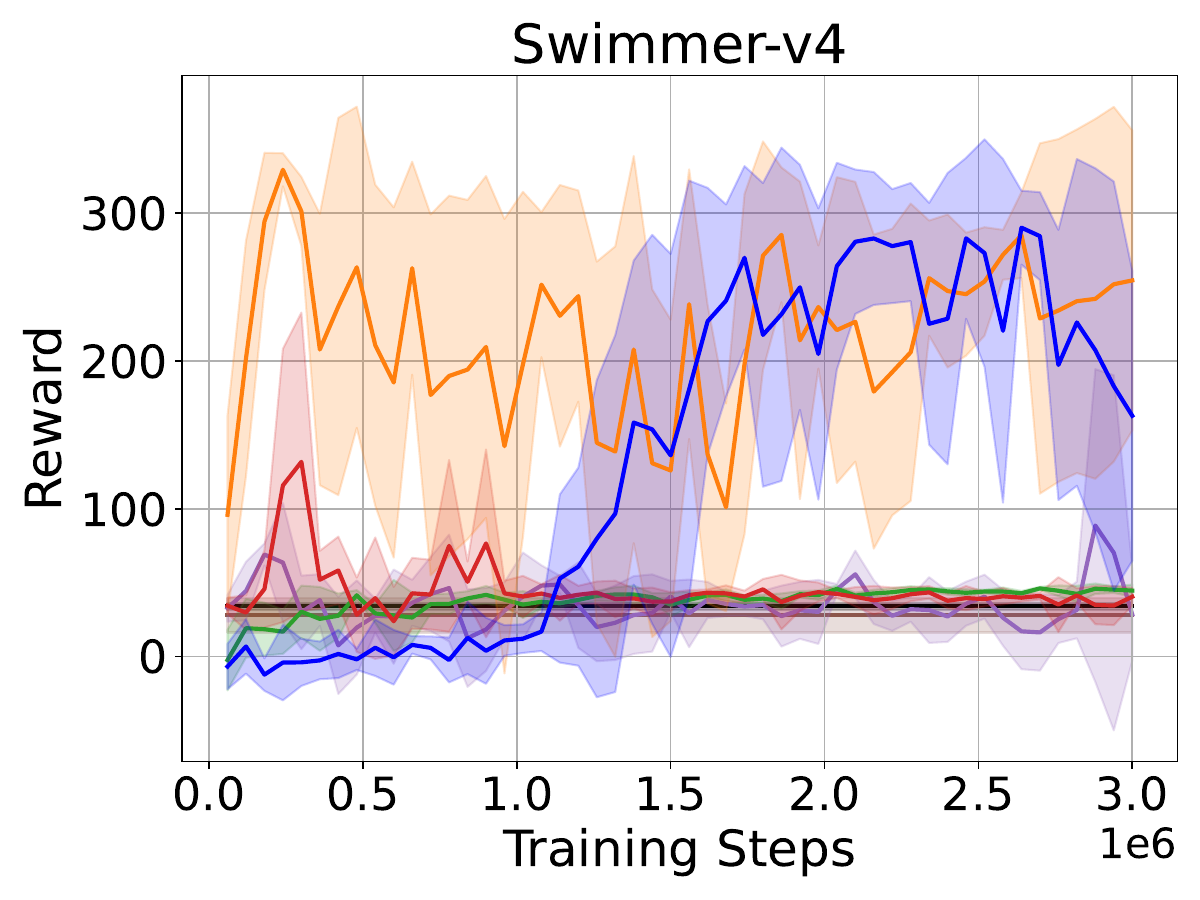}
        \end{subfigure}\hfill
        \begin{subfigure}{0.248\textwidth}
            \centering
            \includegraphics[width=\linewidth]{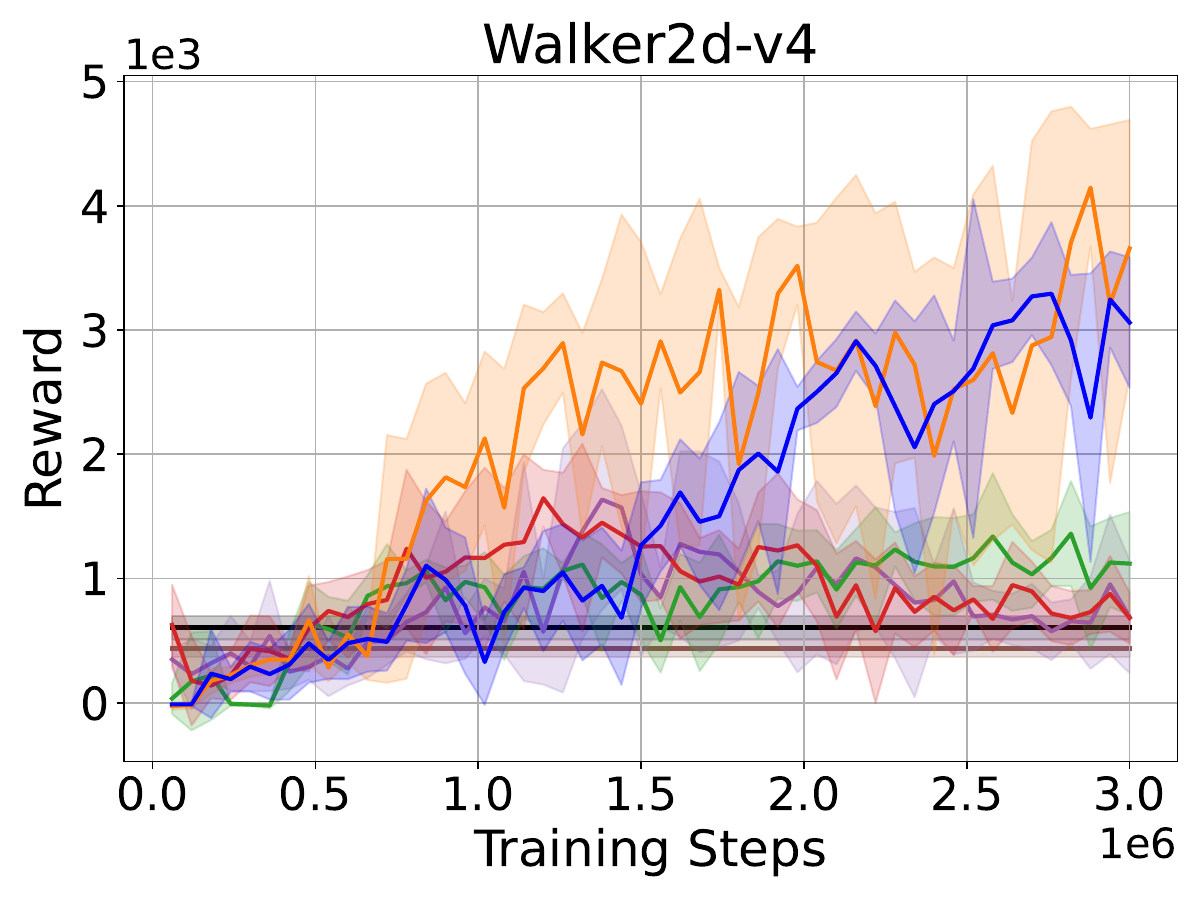}
        \end{subfigure}
        \caption{Low expertise test}
        \label{fig:main2}
    \end{subfigure}

    \caption{Performance comparison (mean $\pm$ std, shown as shaded region) with baselines. Additional results for Sec.~\ref{sec:mainexp} of the main article.}
    \label{fig:main_combined}
\end{figure}

\begin{table*}[h!]
\centering
\caption{Performance (reward) of checkpoints used as optimal and suboptimal policies across environments. Values are reported as mean $\pm$ standard deviation.}
\begin{tabular}{lcc}
\toprule
Environment & Optimal policy $\pi^*$ & Suboptimal policy $\pi^\text{sub}$ \\
\midrule
Ant-v4         & $6766.26 \pm 109.70$  & $1266.37 \pm 533.96$ \\
HalfCheetah-v4 & $15947.53 \pm 39.08$  & $4030.23 \pm 169.82$ \\
Swimmer-v4     & $339.80 \pm 1.15$     & $41.60 \pm 2.34$ \\
Walker-v4      & $5638.94 \pm 49.76$   & $554.38 \pm 267.99$ \\
\bottomrule
\end{tabular}
\label{tab:rl_policy}
\end{table*}

\begin{table*}[h!]
\centering
\caption{True expertise levels $\{\alpha_i\}_{i=1}^m$ for different numbers of demonstrators across two tests. Expertise levels are uniformly distributed within each range.}
\resizebox{0.99\columnwidth}{!}{%
\begin{tabular}{lcccccccc}
\hline
Demonstrator & 1 & 2 & 3 & 4 & 5 & 6 & 7 & 8 \\
\hline
\multicolumn{9}{l}{\textbf{General expertise test}} \\
\hline
2 demonstrators & 0.100 & 0.900 & --    & --    & --    & --    & --    & --    \\
4 demonstrators & 0.100 & 0.366 & 0.633 & 0.900 & --    & --    & --    & --    \\
6 demonstrators & 0.100 & 0.260 & 0.420 & 0.580 & 0.740 & 0.900 & --    & --    \\
8 demonstrators & 0.100 & 0.214 & 0.329 & 0.443 & 0.557 & 0.671 & 0.786 & 0.900 \\
\midrule
\multicolumn{9}{l}{\textbf{Low expertise test}} \\
\hline
2 demonstrators & 0.050 & 0.150 & --    & --    & --    & --    & --    & --    \\
4 demonstrators & 0.050 & 0.083 & 0.117 & 0.150 & --    & --    & --    & --    \\
6 demonstrators & 0.050 & 0.070 & 0.090 & 0.110 & 0.130 & 0.150 & --    & --    \\
8 demonstrators & 0.050 & 0.064 & 0.079 & 0.093 & 0.107 & 0.121 & 0.136 & 0.150 \\
\hline
\end{tabular}
}
\label{tab:exp_true}
\end{table*}

\begin{table*}[h!]
\centering
\caption{Estimated expertise levels $\{\hat{\alpha}_i\}_{i=1}^m$, showing the mean estimate and the mean $\pm$ standard deviation of error across five random seeds. True values are shown separately. Additional results for Sec.~\ref{sec:est_expertise} of the main article.}
\resizebox{0.99\columnwidth}{!}{%
\begin{tabular}{lccccccc}
\hline
Demonstrator & 1 & 2 & 3 & 4 & 5 & 6 & Mean error $\pm$ std \\
\hline
\multicolumn{8}{l}{\textbf{General expertise test}} \\
\hline
Ant-v4 & 0.100 & 0.260 & 0.420 & 0.580 & 0.740 & 0.900 & $(2.67 \pm 1.28)\times10^{-4}$ \\
HalfCheetah-v4 & 0.100 & 0.260 & 0.420 & 0.580 & 0.740 & 0.900 & $(2.00 \pm 2.00)\times10^{-5}$ \\
Hopper-v4 & 0.101 & 0.259 & 0.418 & 0.575 & 0.734 & 0.892 & $(3.83 \pm 2.73)\times10^{-3}$ \\
Swimmer-v4 & 0.100 & 0.260 & 0.420 & 0.580 & 0.740 & 0.900 & $(2.00 \pm 2.00)\times10^{-5}$ \\
Walker2d-v4 & 0.100 & 0.260 & 0.420 & 0.581 & 0.741 & 0.900 & $(3.27 \pm 3.12)\times10^{-4}$ \\
\hline
True expertise levels & 0.100 & 0.260 & 0.420 & 0.580 & 0.740 & 0.900 & \\
\hline
\multicolumn{8}{l}{\textbf{Low expertise test}} \\
\hline
Ant-v4 & 0.148 & 0.170 & 0.187 & 0.207 & 0.222 & 0.246 & $(9.65 \pm 0.24)\times10^{-2}$ \\
HalfCheetah-v4 & 0.086 & 0.105 & 0.127 & 0.146 & 0.167 & 0.190 & $(3.67 \pm 0.15)\times10^{-2}$ \\
Hopper-v4 & 0.156 & 0.176 & 0.195 & 0.216 & 0.231 & 0.257 & $(1.05 \pm 0.02)\times10^{-1}$ \\
Swimmer-v4 & 0.077 & 0.100 & 0.118 & 0.139 & 0.158 & 0.177 & $(2.81 \pm 0.11)\times10^{-2}$ \\
Walker2d-v4 & 0.179 & 0.206 & 0.219 & 0.248 & 0.263 & 0.287 & $(1.34 \pm 0.04)\times10^{-1}$ \\
\hline
True expertise levels & 0.050 & 0.070 & 0.090 & 0.110 & 0.130 & 0.150 & \\
\hline
\end{tabular}
}
\label{tab:expertise_est_details}
\end{table*}

\begin{table*}[h!]
\centering
\caption{Classification / labeling results (mean $\pm$ std). Stage 1: Labeling using the pretrained scoring model. Stage 2: Relabeling using the fine-tuned scoring model. TP (true positive), TN (true negative), FP (false positive), FN (false negative). Improvement (\%) shows the Stage 1 $\rightarrow$ Stage 2 percentage change of accuracy and precision after relabeling. Additional results for Sec.~\ref{sec:label_res} of the main article.}
\resizebox{0.99\columnwidth}{!}{%
\begin{tabular}{lccccccc}
\toprule
Environment & Stage & TP & TN & FP & FN & Accuracy / Precision & Improvement (\%) \\
\midrule
\multicolumn{8}{l}{\textbf{General expertise test}} \\
\hline
\multirow{2}{*}{Ant-v4} 
 & 1 & 14988.6 $\pm$ 3.9 & 14984.2 $\pm$ 10.1 & 15.8 $\pm$ 10.1 & 11.4 $\pm$ 3.9 & 0.9991 / 0.9989 & - \\
 & 2 & 14984.0 $\pm$ 5.5 & 14984.0 $\pm$ 5.5 & 16.0 $\pm$ 5.5 & 16.0 $\pm$ 5.5 & 0.9989 / 0.9989 & -0.02 / 0.00 \\
\midrule
\multirow{2}{*}{HalfCheetah-v4} 
 & 1 & 15000.0 $\pm$ 0.0 & 14999.4 $\pm$ 0.8 & 0.6 $\pm$ 0.8 & 0.0 $\pm$ 0.0 & 1.0000 / 1.0000 & - \\
 & 2 & 15000.0 $\pm$ 0.0 & 15000.0 $\pm$ 0.0 & 0.0 $\pm$ 0.0 & 0.0 $\pm$ 0.0 & 1.0000 / 1.0000 & 0.00 / 0.00 \\
\midrule
\multirow{2}{*}{Swimmer-v4} 
 & 1 & 15000.0 $\pm$ 0.0 & 14999.4 $\pm$ 1.2 & 0.6 $\pm$ 1.2 & 0.0 $\pm$ 0.0 & 1.0000 / 1.0000 & - \\
 & 2 & 14991.4 $\pm$ 16.2 & 14991.4 $\pm$ 16.2 & 8.6 $\pm$ 16.2 & 8.6 $\pm$ 16.2 & 0.9994 / 0.9994 & -0.06 / -0.06 \\
\midrule
\multirow{2}{*}{Walker2d-v4} 
 & 1 & 14992.0 $\pm$ 2.2 & 14985.0 $\pm$ 6.1 & 15.0 $\pm$ 6.1 & 8.0 $\pm$ 2.2 & 0.9992 / 0.9990 & - \\
 & 2 & 14990.4 $\pm$ 1.6 & 14990.4 $\pm$ 1.6 & 9.6 $\pm$ 1.6 & 9.6 $\pm$ 1.6 & 0.9994 / 0.9994 & 0.02 / 0.04 \\
\midrule

\multicolumn{8}{l}{\textbf{Low expertise test}} \\
\hline
\multirow{2}{*}{Ant-v4} 
 & 1 & 2997.6 $\pm$ 1.5 & 24101.4 $\pm$ 735.5 & 2898.6 $\pm$ 735.5 & 2.4 $\pm$ 1.5 & 0.9033 / 0.5178 & - \\
 & 2 & 2947.0 $\pm$ 32.9 & 26947.0 $\pm$ 32.9 & 53.0 $\pm$ 32.9 & 53.0 $\pm$ 32.9 & 0.9965 / 0.9823 & 10.32 / 89.71 \\
\midrule
\multirow{2}{*}{HalfCheetah-v4} 
 & 1 & 3000.0 $\pm$ 0.0 & 25898.2 $\pm$ 355.8 & 1101.8 $\pm$ 355.8 & 0.0 $\pm$ 0.0 & 0.9633 / 0.7368 & - \\
 & 2 & 2997.4 $\pm$ 5.2 & 26997.4 $\pm$ 5.2 & 2.6 $\pm$ 5.2 & 2.6 $\pm$ 5.2 & 0.9998 / 0.9991 & 3.79 / 35.60 \\
\midrule
\multirow{2}{*}{Swimmer-v4} 
 & 1 & 3000.0 $\pm$ 0.0 & 26156.2 $\pm$ 393.5 & 843.8 $\pm$ 393.5 & 0.0 $\pm$ 0.0 & 0.9719 / 0.7884 & - \\
 & 2 & 2637.2 $\pm$ 290.4 & 26637.2 $\pm$ 290.4 & 362.8 $\pm$ 290.4 & 362.8 $\pm$ 290.4 & 0.9758 / 0.8791 & 0.40 / 11.50 \\
\midrule
\multirow{2}{*}{Walker2d-v4} 
 & 1 & 2965.0 $\pm$ 42.7 & 22953.2 $\pm$ 890.1 & 4046.8 $\pm$ 890.1 & 35.0 $\pm$ 42.7 & 0.8639 / 0.4301 & - \\
 & 2 & 2772.0 $\pm$ 170.3 & 26772.0 $\pm$ 170.3 & 228.0 $\pm$ 170.3 & 228.0 $\pm$ 170.3 & 0.9848 / 0.9240 & 13.99 / 114.83 \\
\bottomrule
\end{tabular}
}
\label{tab:acc_label_improvement}
\end{table*}

\begin{table*}[h!]
\centering
\caption{Ablation results on the impact of relabeling and early stopping in Stage 2 on Ant-v4 task. Reported as mean $\pm$ std of reward across random seeds; each seed's value is the mean over its last five checkpoints. Additional results for Sec.~\ref{sec:abl} of the main article.}
\begin{tabular}{lcc}
\toprule
Method & Expertise [0.1 -- 0.9] & Expertise [0.05 -- 0.15] \\
\midrule
LEOS-IL (ours) & \textbf{6084.6 $\pm$ 146.9} & \textbf{5932.0 $\pm$ 335.2} \\
Scoring (no relabel) & 5593.0 $\pm$ 564.2 & 3316.2 $\pm$ 1207.8 \\
Scoring (no early stop) & 2129.8 $\pm$ 2205.8 & 1018.2 $\pm$ 250.0 \\
\bottomrule
\end{tabular}
\label{tab:ab_relabel_early}
\end{table*}

\begin{table*}[h!]
\centering
\caption{Ablation study on top-k hyperparameter on Ant-v4 task. Reported as mean $\pm$ std of reward across random seeds; each seed's value is the mean over its last five checkpoints. Additional results for Sec.~\ref{sec:abl} of the main article.}
\begin{tabular}{lcc}
\toprule
Top-k fraction & Expertise [0.1 -- 0.9] & Expertise [0.05 -- 0.15] \\
\midrule
0.05  & --                  & 5887.6 $\pm$ 263.8 \\
0.1   & 5403.8 $\pm$ 378.7  & \textbf{5932.0 $\pm$ 335.2} \\
0.15  & --                  & 3663.4 $\pm$ 1950.9 \\
0.2   & --                  & 1756.6 $\pm$ 902.6 \\
0.3   & 5809.0 $\pm$ 274.8  & -- \\
0.5   & \textbf{6084.6 $\pm$ 146.9} & -- \\
0.55  & 4109.2 $\pm$ 2204.9 & -- \\
0.6   & 3641.0 $\pm$ 1237.7 & -- \\
0.7   & 2383.0 $\pm$ 1059.7 & -- \\
\bottomrule
\end{tabular}
\label{tab:ab_topk}
\end{table*}

\FloatBarrier
\subsection{Reproducibility Statement}

{To ensure the reproducibility of this work, details of the training parameters are provided in Appendix Table~\ref{tab:training_paras}. All reported results are averaged over five random seeds (0–4) per condition/method, with mean and standard deviation reported. The corresponding code is publicly available at:
\url{https://github.com/ch-engpan/IL_imperfect}}

\begin{center}
\captionof{table}{{Training hyperparameters.}}
\label{tab:two_stage_params}
\begin{tabular}{l l c}
\hline
\textbf{Stage} & \textbf{Parameter} & \textbf{Value} \\
\hline
\multirow{6}{*}{Stage 1} 
& $f_\phi$ learning rate          & $1 \times 10^{-4}$ \\
& $f_\phi$ batch size             & 1000 \\
& $f_\phi$ optimizer              & Adam \\
& Variance threshold $\epsilon$ for expertise levels & 0.0005 \\
& Expertise levels estimation frequency        & Every 5 epochs \\
& $f_\phi$ training epochs                & 50 (Ant, HalfCheetah, Swimmer) \\ & &  100 (Walker2d) \\
\hline
\multirow{14}{*}{Stage 2}
& $\pi_\theta$ model                       & SAC (Stable-Baselines3) \\
& $\pi_\theta$ learning starts from                & 100 steps \\
& $\pi_\theta$ batch size                  & 1024 \\
& $\pi_\theta$ learning rate               & $2 \times 10^{-3}$ \\
& $\pi_\theta$ replay buffer size             & $1 \times 10^{6}$ \\
& $\pi_\theta$ training steps / IL iteration    & $1.5 \times 10^{4}$ \\
& $f_\phi$ learning rate          & $1 \times 10^{-4}$ \\
& $f_\phi$ batch size             & 1024 \\
& $f_\phi$ optimizer              & Adam \\
& {Mixup-loss weight $w_{\mathrm{mix}}$} & {2} \\
& $f_\phi$ updates epochs / IL iteration    & 500 \\
& $\Delta_t, p$ for relabeling early stopping         & 30 (for 5 consecutive steps) \\
& IL total iterations               & 200 \\
& checkpoints saving freqency              & 1 per IL iteration \\
\hline
\end{tabular}
\label{tab:training_paras}
\end{center}




\end{appendices}


\begingroup
\clearpage
\input{imperfect-imitation-merged.bbl}
\endgroup
\clearpage
\end{bibunit}

\pagenumbering{arabic}
\setcounter{footnote}{0}

\counterwithout*{figure}{section}
\counterwithout*{table}{section}
\counterwithout*{algorithm}{section}
\counterwithout*{equation}{section}

\begin{bibunit}[sn-mathphys-num]
\begin{center}
{\Large\bfseries Supplementary Information}\\[0.75em]
{\large Imitating the Imperfect: Offline-to-Online Robust Imitation Learning from Heterogeneous Demonstrators}\\[0.75em]
This document provides further discussion of related work, detailed experimental and ablation results, additional analysis of method variants, and reproducibility details for the accompanying article.
\end{center}

\begin{appendices}
\setcounter{section}{0}
\setcounter{figure}{0}
\setcounter{table}{0}
\setcounter{algorithm}{0}
\setcounter{equation}{0}
\renewcommand{\thesection}{S\arabic{section}}
\renewcommand{\sectionname}{Supplementary Sec.}
\renewcommand{\thefigure}{S\arabic{figure}}
\renewcommand{\thetable}{S\arabic{table}}
\renewcommand{\thealgorithm}{S\arabic{algorithm}}
\renewcommand{\theequation}{S\arabic{equation}}
\renewcommand{\figurename}{Supplementary Fig.}
\renewcommand{\tablename}{Supplementary Table}

\section{Further Discussion on Related Work}
\label{appendix:related_works}
\paragraph{Closer comparisons with ILEED~\cite{beliaev2022imitation}}
ILEED is the closest work to ours, as both address IL from multiple demonstrators with heterogeneous expertise. Despite the shared high-level motivation, the two approaches differ substantially in problem setting, modeling assumptions, methodologies and robustness properties, making them suitable for different practical regimes.
\begin{itemize}
    \item \textbf{Problem setting.} ILEED considers the problem setting where each demonstrator’s policy is a mixture of the optimal policy and a \emph{uniformly random} policy, with the mixing coefficient interpreted as an expertise level.\footnote{For continuous action spaces, ILEED replaces this uniform noise component with a high-entropy variant of the optimal policy. Specifically, the suboptimal component is modeled by scaling the variance of a Gaussian mixture policy, resulting in a near-uniform distribution rather than a structured suboptimal policy.}
    While convenient, this ``suboptimal = uniform noise’’ assumption is restrictive as real-world suboptimal behavior is typically structured, rather than near-uniform noise. For example, a demonstrator may consistently struggle with specific skills without acting randomly overall. 
    
    In contrast, we relax this assumption by modeling each demonstrator’s behavior as a mixture of the optimal policy and an \emph{unknown} suboptimal policy. Our formulation imposes no uniformity or high-entropy constraints on the suboptimal component, enabling the method to capture realistic and highly structured forms of suboptimality.

    \item \textbf{Expertise modeling.} ILEED models expertise as a state-dependent function defined as a sigmoid of the inner product between the learned state and demonstrator embeddings. Although this formulation can represent state-varying expertise, it relies heavily on the quality of the recovered embeddings for \emph{all} states visited by \emph{all} demonstrators. In practice, when demonstrators explore only partial or disjoint regions of the state space, the resulting embeddings, and thus the inferred expertise levels, can become severely biased, causing significant policy degradation.
    
    Our method instead models expertise through the class-prior probability $\mathbb{P}(z{=}1)$, estimated via a surrogate scoring model that directly distinguishes optimal from suboptimal samples across demonstrators. This approach does not require each demonstrator to provide full coverage of the state space. {Moreover, the trained scoring model outputs the sample-wise optimality probability $f_\phi(s,a)=\mathbb{P}(z{=}1\mid s,a)$, and we use the surrogate reward $r_\phi(s,a)=-\log(1-f_\phi(s,a))$ to train the RL agent (SAC).} 
    
    \item \textbf{Methodology.} ILEED performs fully \emph{offline} maximum-likelihood estimation of the policy, state embeddings, and demonstrator embeddings. As a result, it is sensitive to \emph{distribution shift} between the demonstration distribution and the agent’s on-policy distribution during training.
    
    {Our method instead employs a \emph{two-stage} procedure: we first learn a scoring model offline, use the surrogate reward $r_\phi(s,a)=-\log(1-f_\phi(s,a))$ to train an RL agent, and \emph{iteratively refine} the scorer using on-policy rollouts.} To explicitly mitigate covariate shift, we incorporate mixup augmentation, an agent-matching penalty, and a relabeling mechanism, making our approach more robust when the agent visits states that rarely or never appear in the demonstrations.
    


    \item \textbf{Optimization stability.}  
    ILEED jointly optimizes policy parameters, state embeddings, and demonstrator embeddings via a single maximum-likelihood objective. This fully coupled optimization can be unstable, as conflicting gradients among latent components may lead to degenerate solutions.

    {Our method instead employs an EM-style alternating self-training method that alternates between surrogate-risk minimization for the scoring model and expertise estimation from the current hard pseudo-labels. Policy learning then proceeds through a standard RL pipeline using the scoring model as a reward proxy, yielding an empirically effective optimization procedure.}
\end{itemize}

Our empirical results (see Table~\ref{tab:main_exp} and Appendix Fig.~\ref{fig:main_combined} in the main article, and Supplementary Table~\ref{tab:bc_ileed_scoring}) validate these distinctions. In particular, our method is advantageous when demonstrations contain structured suboptimality, exhibit low expertise levels, or when distribution shift between demonstrations and agent rollouts is significant.

\paragraph{Comparisons with T‑REX/D‑REX~\cite{brown2019extrapolating, brown2020better}}

Both \emph{trajectory-ranked reward extrapolation} (T-REX) and \emph{disturbance-based reward extrapolation} (D-REX) learn a reward function from ranked demonstrations, allowing an agent to potentially outperform the demonstrator via extrapolation. T-REX relies on experts to provide ranking comparisons, while D-REX can automatically generate rankings by injecting noise into a cloned policy, based on the assumption that added noise leads to a continuous degradation in performance. 

Compared to T-REX, our method does not require large amounts of supervision or expert ranking feedback during training. Moreover, while D-REX assumes homogeneous demonstrations and introduces heterogeneity artificially, our approach naturally handles heterogeneous datasets and is capable of inferring varying expertise levels within the data. Unlike D-REX, our method does not rely on the assumption that policy performance monotonically decreases with noise injection, which may not hold in more complex domains. In addition, both T-REX and D-REX first learn a surrogate reward function and then optimize a policy using RL on this fixed reward, which can lead to covariate shift when the agent encounters states not present in the demonstrations. Our method addresses this issue by continuously fine-tuning the scoring model with new data collected during RL, resulting in improved robustness and generalization. Overall, our approach provides a flexible and practical framework for leveraging demonstrations, particularly in settings with heterogeneous data and limited supervision.

\paragraph{Comparisons with VPIL~\cite{cai2023imitation}}
\emph{Vaguely pairwise imitation learning} (VPIL) investigates a weak-supervision setting in which the annotator provides vague pairwise feedback: a pair of demonstrations is labeled only when their quality differs substantially, resulting in two contaminated datasets that still contain mixtures of expert and non-expert trajectories.
VPIL leverages this pairwise structure to recover the expert occupancy measure through \emph{mixture-proportion estimation} \cite[MPE;][]{ramaswamy2016mixture} and risk rewriting, and its \emph{comparative imitation learning with risk rewriting with estimation} (COMPILER-E) algorithms integrate the recovered distribution into a GAIL-style adversarial IL framework.

Our problem setting is more general and arguably more challenging.
Unlike VPIL, which assumes access to vague pairwise comparison labels, we consider a scenario in which the agent receives only heterogeneous demonstrations of unknown and varying quality, with no preference, ranking, or optimality labels.
In the absence of such supervision, the learner must infer both the underlying demonstration quality and the reward surrogate directly from unlabeled trajectories.

The methodological differences therefore extend beyond supervision assumptions.
{Rather than reconstructing an expert occupancy measure from pairwise datasets, we learn a fine-grained state–action optimality scoring model through a surrogate demonstrator classification objective and an EM-style alternating self-training procedure for latent expertise estimation.}
The trained scoring model then serves as a surrogate reward within an iterative policy-learning loop, enabling robust imitation learning directly from heterogeneous, unlabeled demonstrators without relying on any preference-based supervision.

\paragraph{Comparisons with AIRL~\cite{fu2017learning}}
\emph{Adversarial inverse reinforcement learning} (AIRL) extends the GAIL framework by structuring the discriminator to decompose into a state-only reward term and a shaping term, enabling recovery of a reward function that is disentangled from environment dynamics. This structure allows AIRL to learn portable rewards that can be re-optimized in new or modified environments.

Our problem setting, however, is fundamentally different. Rather than assuming demonstrations come from a single (near-optimal) expert, we address imitation learning from heterogeneous, unlabeled, and potentially highly suboptimal demonstrators with unknown expertise levels. Although our method learns a scoring model that serves as a surrogate reward, the objective is not reward recovery or transfer, but robust policy learning when expert dominance does not hold.
AIRL, like GAIL, is not designed to handle heterogeneous demonstrators and presumes access to expert demonstrations.
Therefore, their assumptions and objectives cannot address the key challenges in our setting.

\section{Experiment Details}

\subsection{Detailed results \& Ablation studies}

\begin{table*}[h!]
\centering
\caption{Estimated expertise levels $\{\hat{\alpha}_i\}_{i=1}^m$ on Ant-v4 task, showing the mean estimate and the mean $\pm$ standard deviation of error across five random seeds. True values are shown separately. Results shown for different numbers of demonstrators (2, 4, 6, 8), each provides 5,000 samples. Supplementary results for Sec.~\ref{sec:abl} of the main article.}
\resizebox{0.98\columnwidth}{!}{%
\begin{tabular}{l l cccccccc c}
\hline
Demonstrator & Type & 1 & 2 & 3 & 4 & 5 & 6 & 7 & 8 & Mean error $\pm$ std \\
\hline
\multicolumn{11}{l}{\textbf{General expertise test}} \\
\hline
2 demonstrators  & Est.  & 0.101 & 0.899 & -     & -     & -     & -     & -     & -     & $(9.60 \pm 1.60)\times10^{-4}$ \\
                 & True  & 0.100 & 0.900 & -     & -     & -     & -     & -     & -     & - \\
\hline
4 demonstrators  & Est.  & 0.100 & 0.366 & 0.633 & 0.899 & -     & -     & -     & -     & $(4.50 \pm 2.96)\times10^{-4}$ \\
                 & True  & 0.100 & 0.367 & 0.633 & 0.900 & -     & -     & -     & -     & - \\
\hline
6 demonstrators  & Est.  & 0.100 & 0.260 & 0.420 & 0.580 & 0.740 & 0.900 & -     & -     & $(2.67 \pm 1.28)\times10^{-4}$ \\
                 & True  & 0.100 & 0.260 & 0.420 & 0.580 & 0.740 & 0.900 & -     & -     & - \\
\hline
8 demonstrators  & Est.  & 0.100 & 0.214 & 0.329 & 0.443 & 0.557 & 0.671 & 0.785 & 0.899 & $(2.50 \pm 1.97)\times10^{-4}$ \\
                 & True  & 0.100 & 0.214 & 0.329 & 0.443 & 0.557 & 0.671 & 0.786 & 0.900 & - \\
\hline
\multicolumn{11}{l}{\textbf{Low expertise test}} \\
\hline
2 demonstrators  & Est.  & 0.193 & 0.287 & -     & -     & -     & -     & -     & -     & $(1.40 \pm 0.03)\times10^{-1}$ \\
                 & True  & 0.050 & 0.150 & -     & -     & -     & -     & -     & -     & - \\
\hline
4 demonstrators  & Est.  & 0.106 & 0.143 & 0.173 & 0.208 & -     & -     & -     & -     & $(5.77 \pm 0.16)\times10^{-2}$ \\
                 & True  & 0.050 & 0.083 & 0.117 & 0.150 & -     & -     & -     & -     & - \\
\hline
6 demonstrators  & Est.  & 0.148 & 0.170 & 0.187 & 0.207 & 0.222 & 0.246 & -     & -     & $(9.65 \pm 0.24)\times10^{-2}$ \\
                 & True  & 0.050 & 0.070 & 0.090 & 0.110 & 0.130 & 0.150 & -     & -     & - \\
\hline
8 demonstrators  & Est.  & 0.073 & 0.087 & 0.101 & 0.115 & 0.130 & 0.144 & 0.160 & 0.175 & $(2.31 \pm 0.10)\times10^{-2}$ \\
                 & True  & 0.050 & 0.064 & 0.079 & 0.093 & 0.107 & 0.121 & 0.136 & 0.150 & - \\
\hline
\end{tabular}
}\label{tab:demo_num_expertise}
\end{table*}

\begin{table*}[h!]
\centering
\caption{Ablation on the different number of demonstrators (each providing 5,000 samples) for Ant-v4 task. Reported numbers are mean $\pm$ std across random seeds; each seed's value is the mean over its last five checkpoints. Supplementary results for Sec.~\ref{sec:abl} of the main article.}
\begin{tabular}{lcc}
\toprule
\ Number of Demonstrators  & Expertise [0.1 -- 0.9] & Expertise [0.05 -- 0.15] \\
\midrule
2 & 5634.0 $\pm$ 617.8 & 6050.6 $\pm$ 120.4 \\
4 & 5934.0 $\pm$ 264.3 & 5637.2 $\pm$ 772.5 \\
6 & \textbf{6084.6 $\pm$ 146.9} & 5932.0 $\pm$ 335.2 \\
8 & 5735.0 $\pm$ 225.6 & \textbf{6146.6 $\pm$ 107.2} \\
\bottomrule
\end{tabular}
\label{tab:demo_num_r}
\end{table*}

\begin{figure}[h]
    \centering
    \begin{subfigure}{0.4\textwidth}
        \centering
        \includegraphics[width=\linewidth]{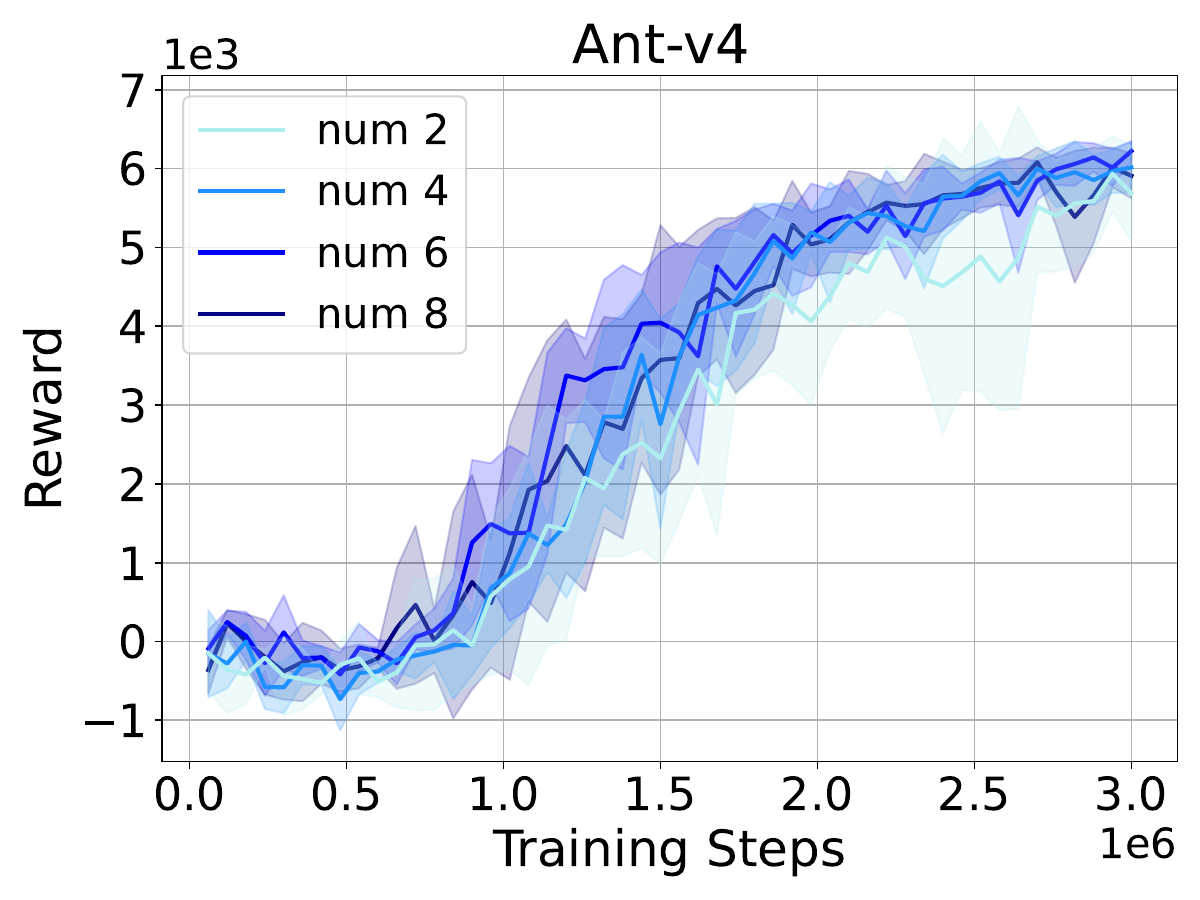}
        \caption{Expertise level range [0.1 - 0.9]}
    \end{subfigure}
    \hspace{0.5cm}
    \begin{subfigure}{0.4\textwidth}
        \centering
        \includegraphics[width=\linewidth]{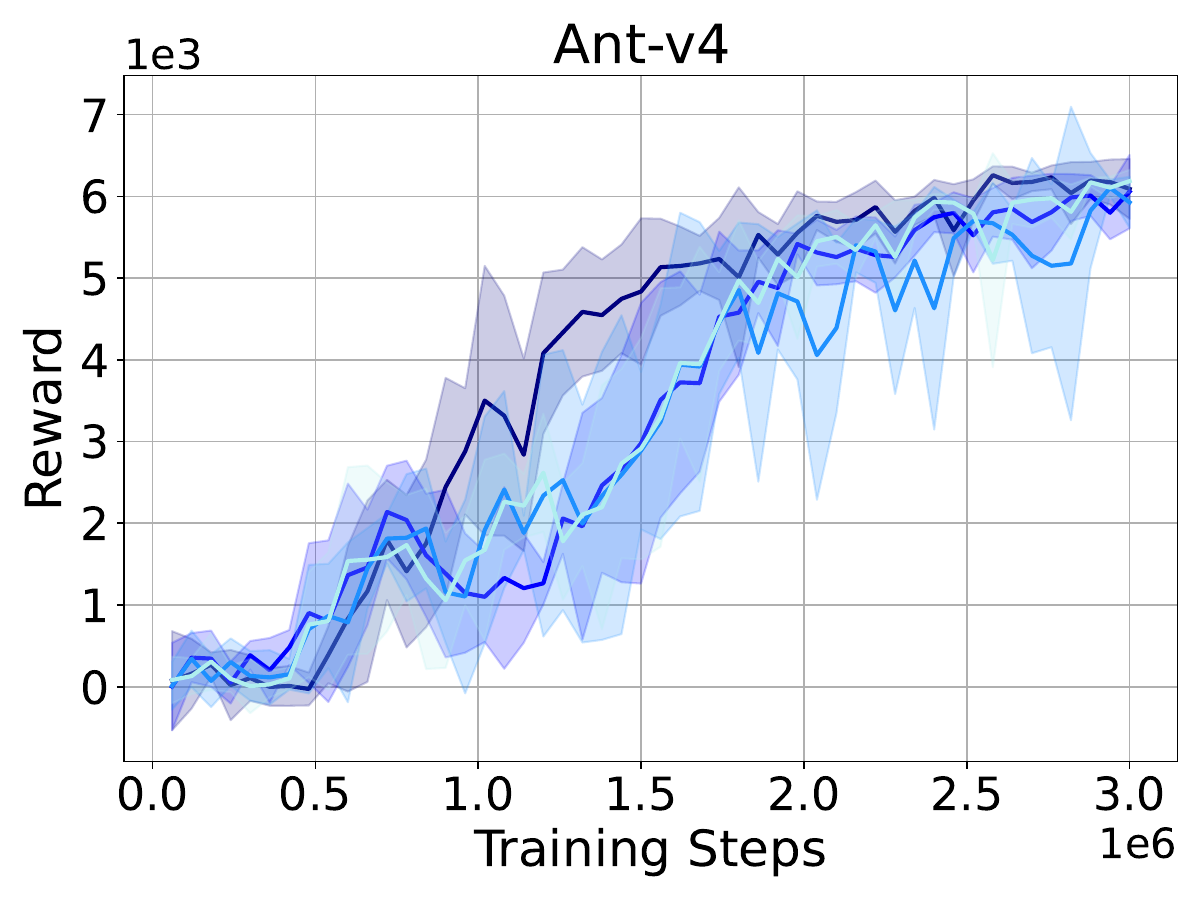}
        \caption{Expertise level range [0.05 - 0.15]}
    \end{subfigure}
    \caption{Ablation on different numbers of demonstrators for Ant-v4 task (mean $\pm$ std, shown as shaded region). Supplementary results for Sec.~\ref{sec:abl} of the main article.}
    \label{fig:abl_bagNum}
\end{figure}

\textbf{Number of Demonstrators:} The results in Supplementary Table~\ref{tab:demo_num_expertise} show that the estimation error of expertise levels in low expertise test increase when the number of demonstrators is reduced. However, the reward obtained by the IL agent in Stage 2 (see Supplementary Table~\ref{tab:demo_num_r}) does not decrease substantially. These results demonstrate the robustness of our method with respect to varying demonstrator counts in terms of IL performance.

\begin{figure}[h!]
    \centering

    \begin{subfigure}{\textwidth}
        \centering
        \begin{subfigure}{0.249\textwidth}
            \centering
            \includegraphics[width=\linewidth]{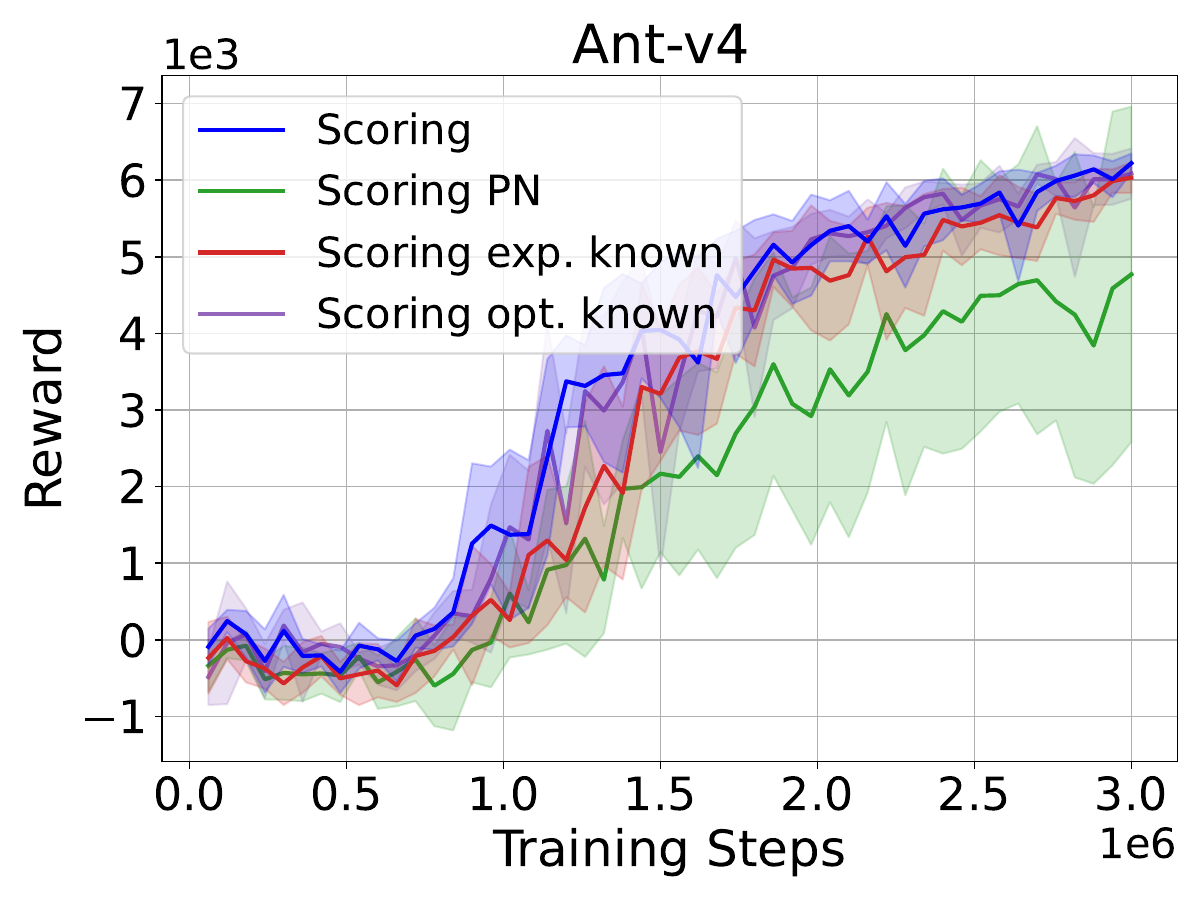}
        \end{subfigure}\hfill
        \begin{subfigure}{0.249\textwidth}
            \centering
            \includegraphics[width=\linewidth]{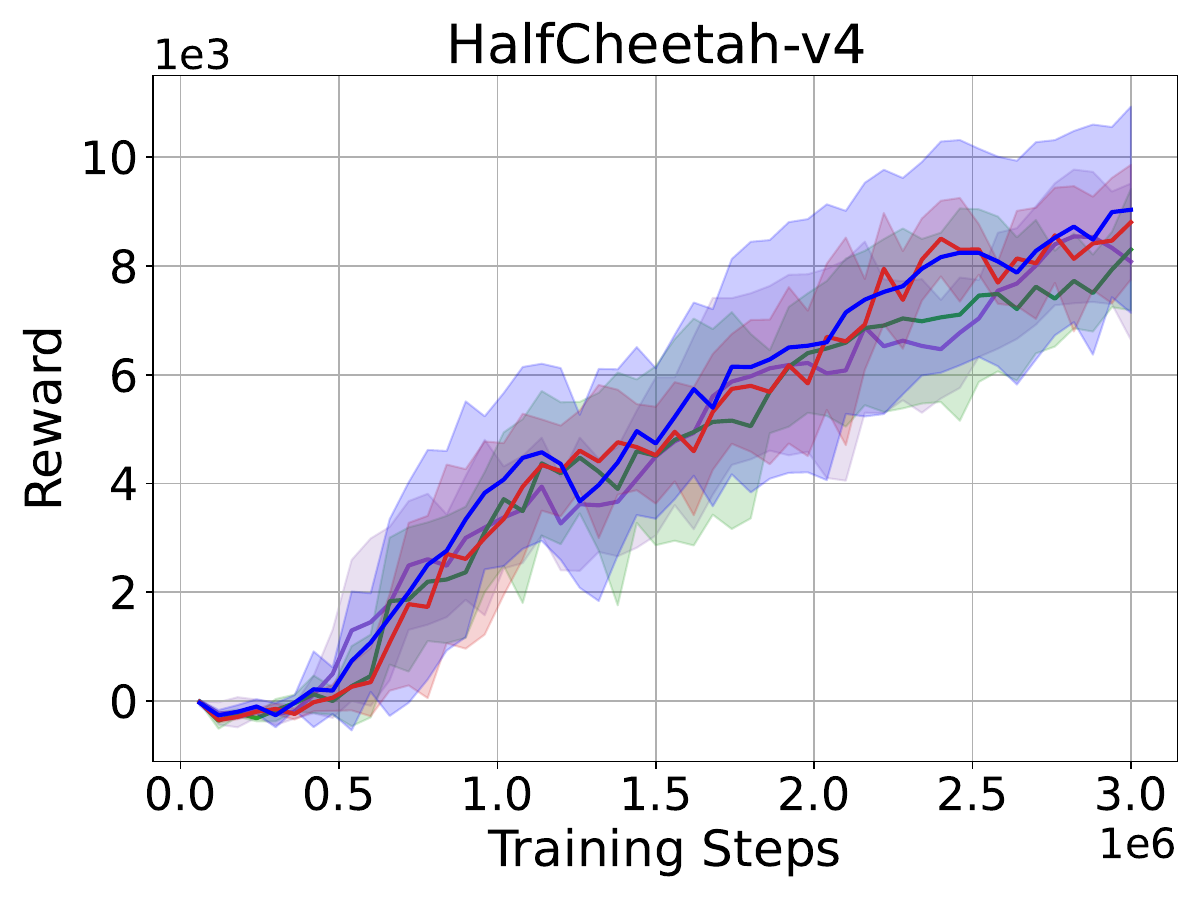}
        \end{subfigure}\hfill
        \begin{subfigure}{0.249\textwidth}
            \centering
            \includegraphics[width=\linewidth]{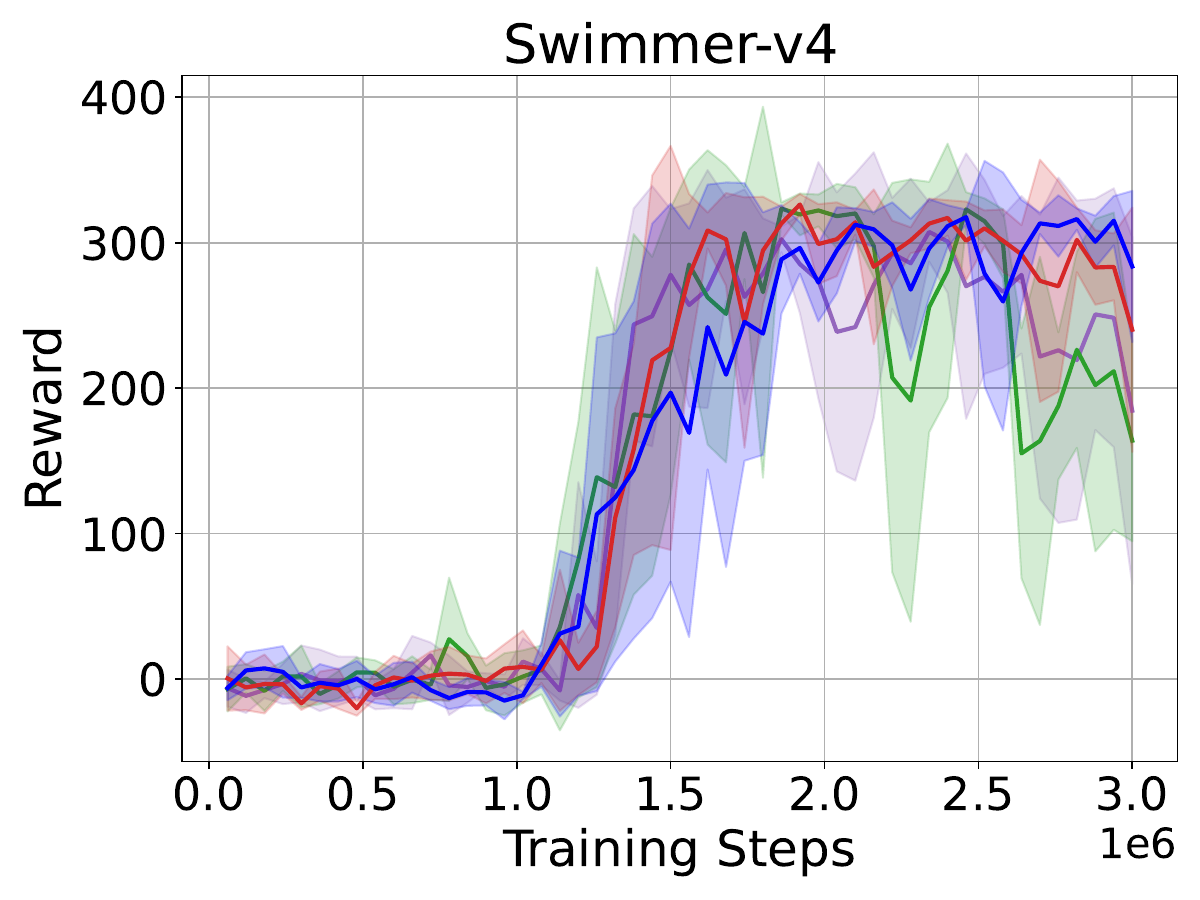}
        \end{subfigure}\hfill
        \begin{subfigure}{0.249\textwidth}
            \centering
            \includegraphics[width=\linewidth]{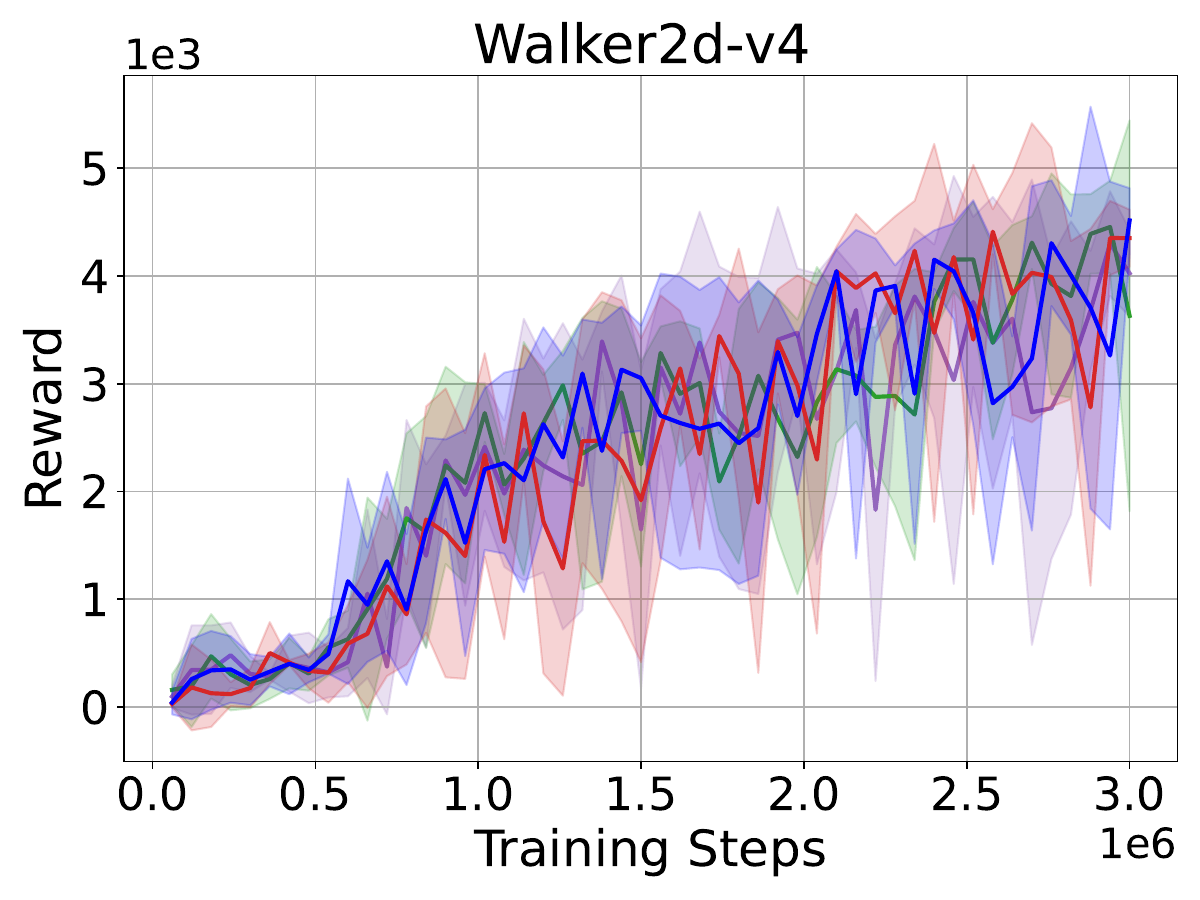}
        \end{subfigure}
        \caption{General expertise test}
        \label{fig:analy_1}
    \end{subfigure}

    \vspace{1em} 

    \begin{subfigure}{\textwidth}
        \centering
        \begin{subfigure}{0.249\textwidth}
            \centering
            \includegraphics[width=\linewidth]{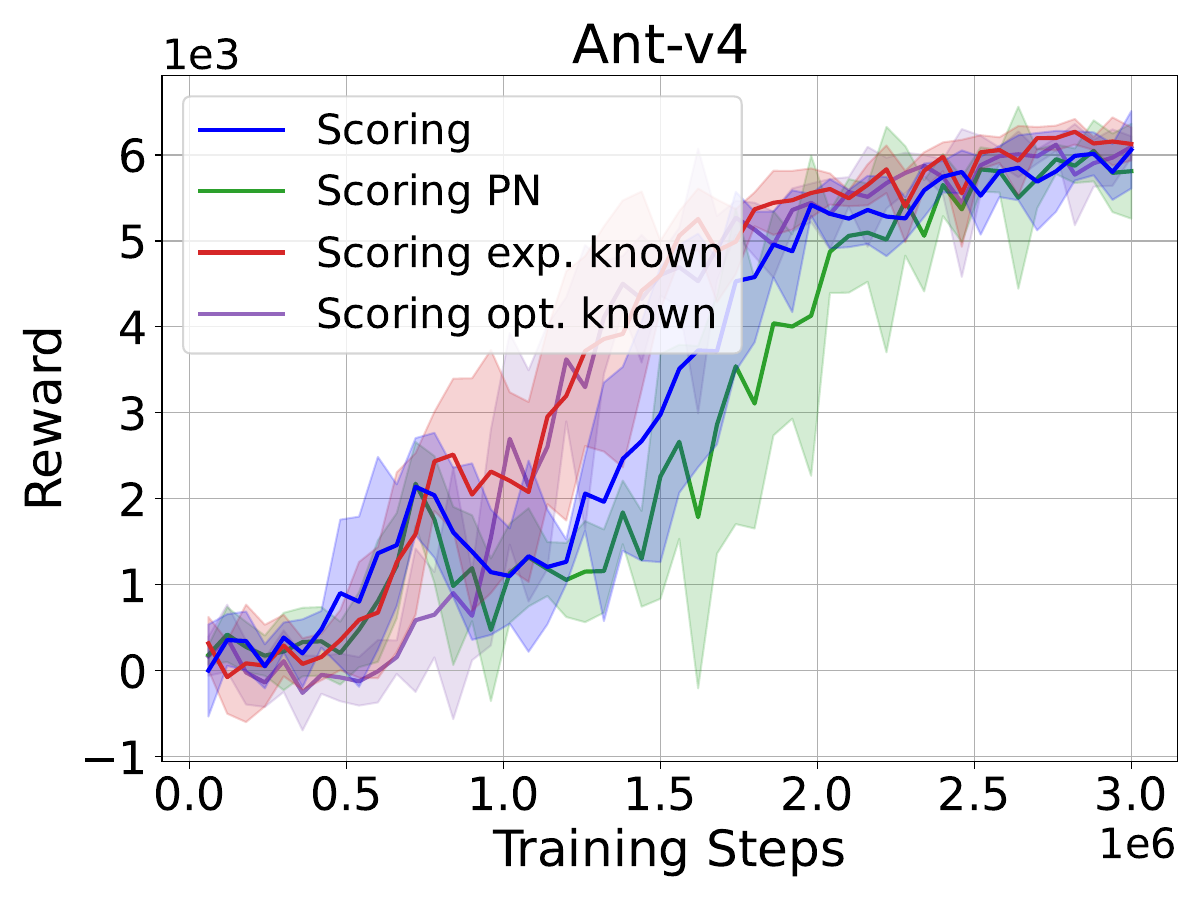}
        \end{subfigure}\hfill
        \begin{subfigure}{0.249\textwidth}
            \centering
            \includegraphics[width=\linewidth]{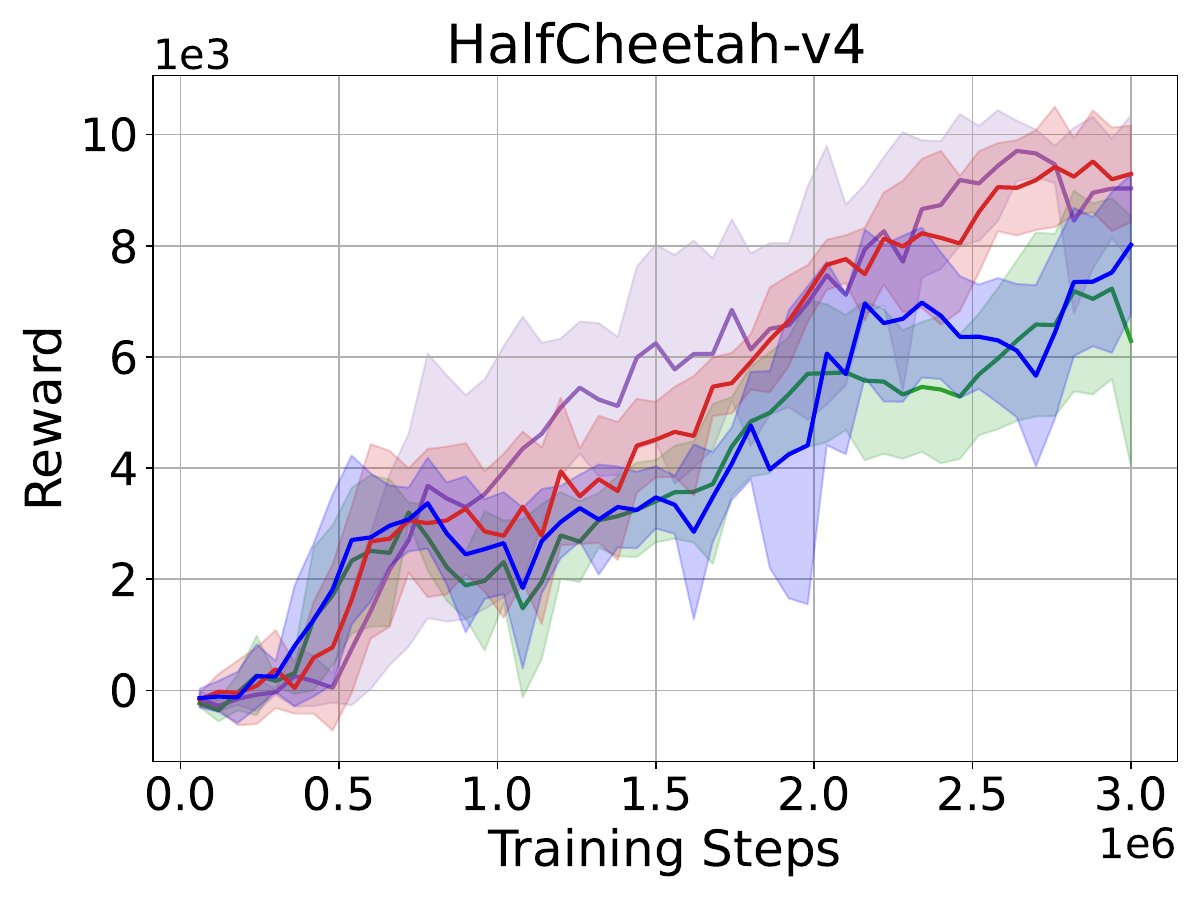}
        \end{subfigure}\hfill
        \begin{subfigure}{0.249\textwidth}
            \centering
            \includegraphics[width=\linewidth]{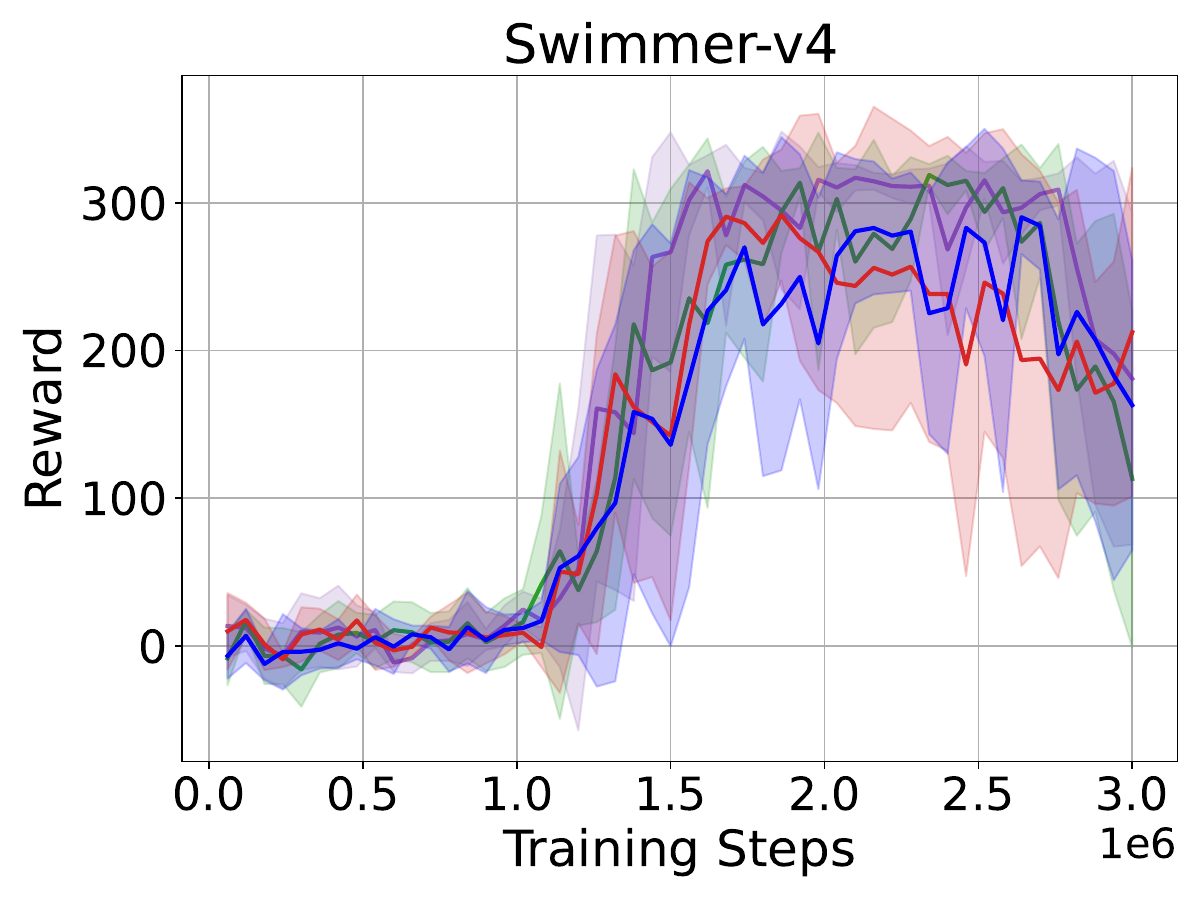}
        \end{subfigure}\hfill
        \begin{subfigure}{0.249\textwidth}
            \centering
            \includegraphics[width=\linewidth]{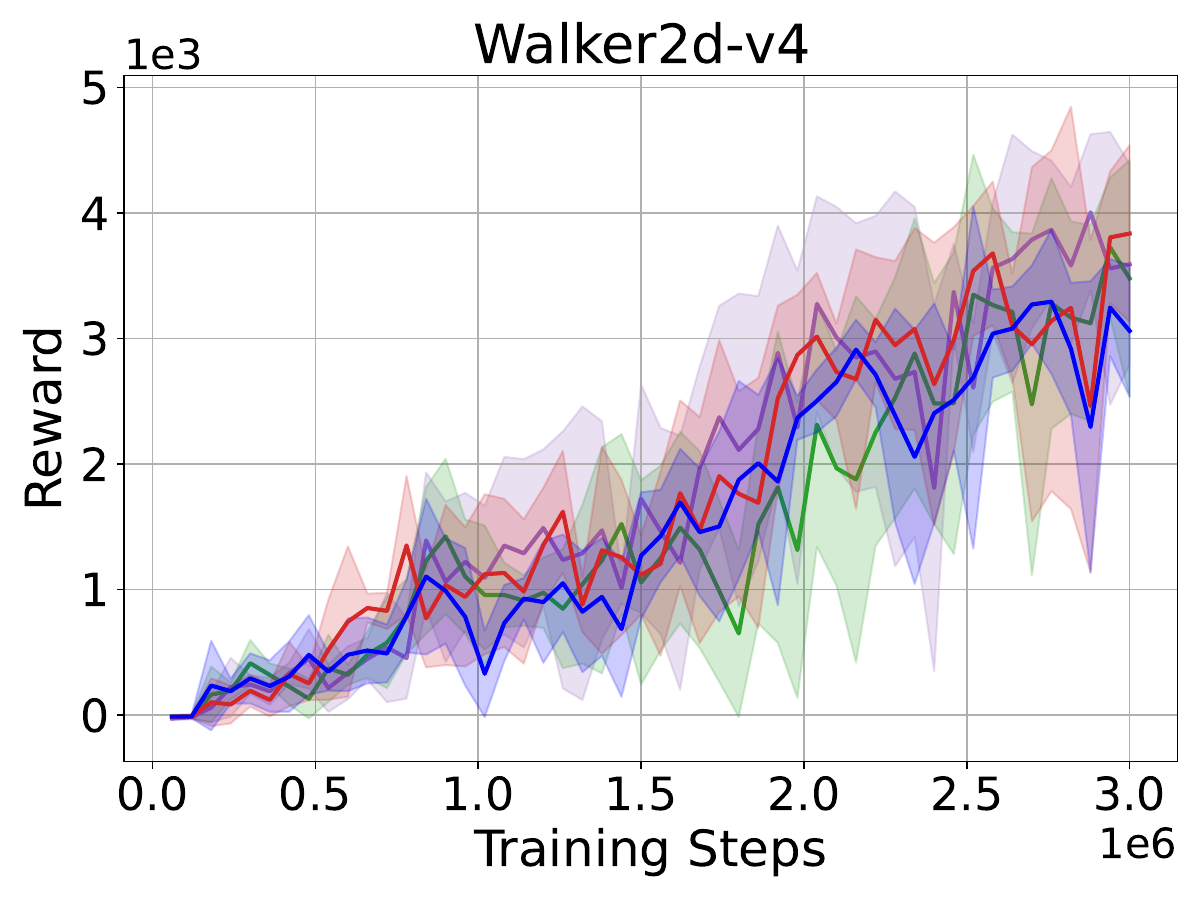}
        \end{subfigure}
        \caption{Low expertise test}
        \label{fig:analy_2}
    \end{subfigure}

    \caption{Comparative analysis of our method and its variants.}
    \label{fig:analy_combined}
\end{figure}

\begin{figure}[h]
    \centering
    \begin{subfigure}{0.4\textwidth}
        \centering
        \includegraphics[width=\linewidth]{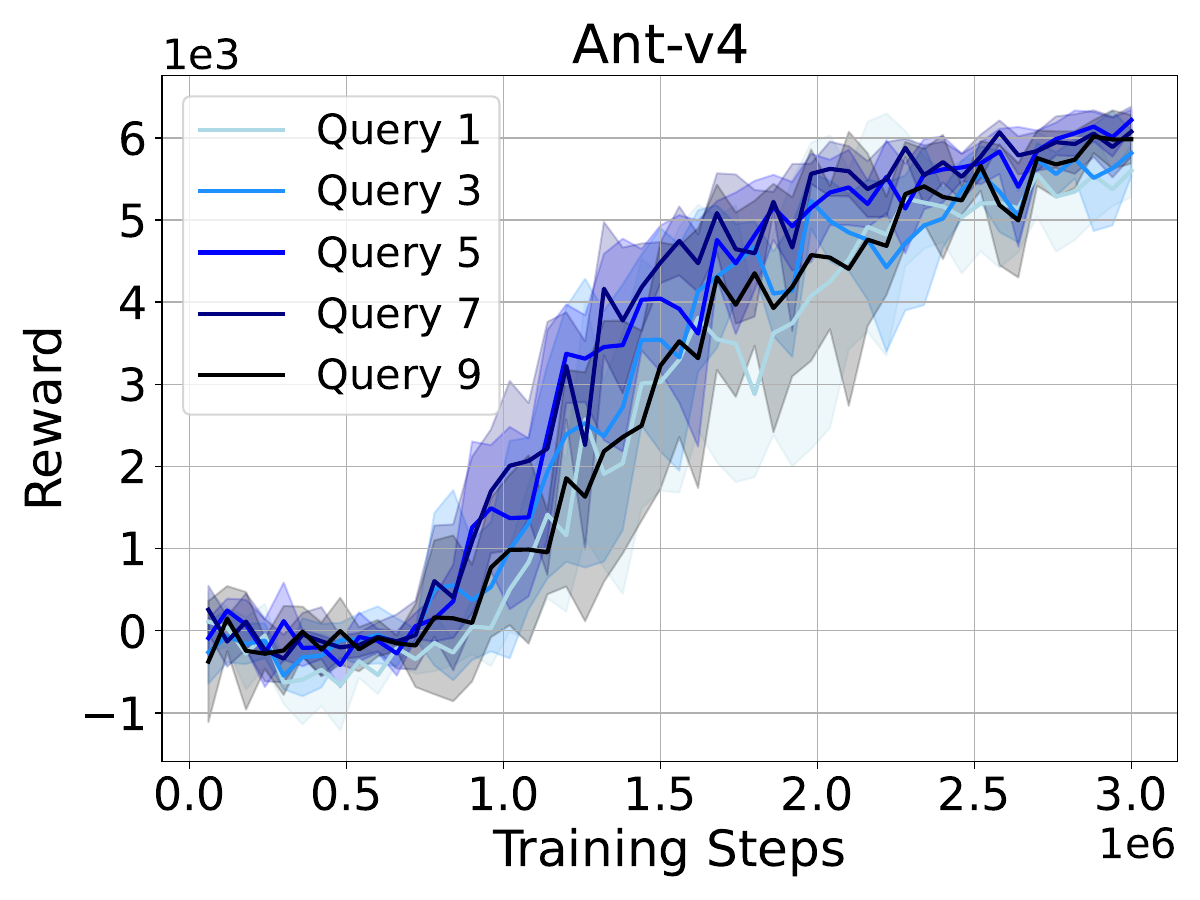}
        \caption{Expertise level range [0.1 - 0.9]}
    \end{subfigure}
    \hspace{0.5cm}
    \begin{subfigure}{0.4\textwidth}
        \centering
        \includegraphics[width=\linewidth]{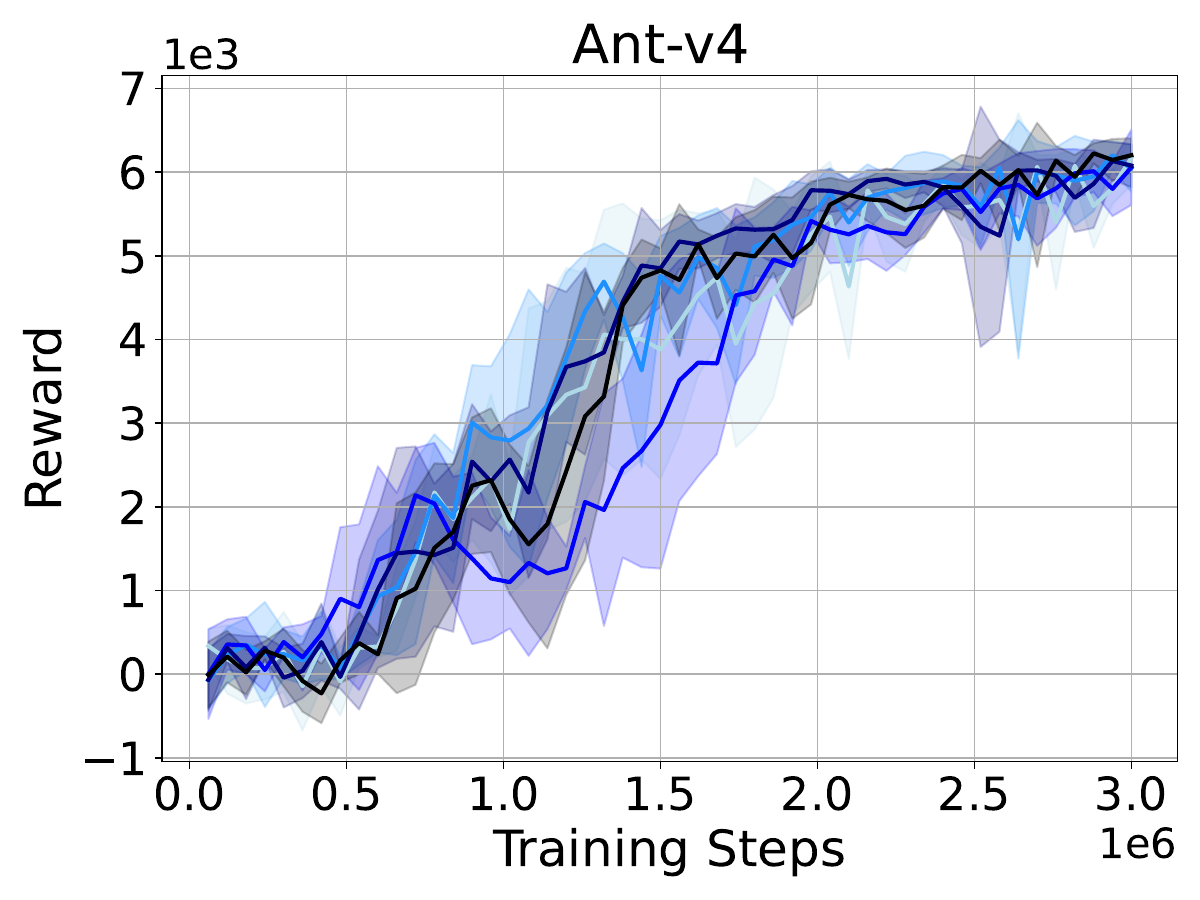}
        \caption{Expertise level range [0.05 - 0.15]}
    \end{subfigure}
    \caption{Ablation on the different expert query count for Ant-v4 task (mean $\pm$ std, shown as shaded region). See Supplementary Table~\ref{tab:acc_prec_stages_exp_ask_num}.}
    \label{fig:abl_expCheckNum}
\end{figure}

\textbf{Number of Human Queries}:
To study the sensitivity to the number of human queries, we conducted ablation experiments, varying the number of queries. The results are shown in Supplementary Tables~\ref{tab:acc_prec_stages_exp_ask_num}--\ref{tab:query_ablation} and Supplementary Fig.~\ref{fig:abl_expCheckNum}. We observe that even with only a single query, our method still achieves similar performance to that obtained with 5 queries. This is because the classification accuracy in Stage 1 is extremely high (above 99\%), making it very unlikely for the system to select an incorrectly labeled sample that would lead to completely opposite or flipped classification results. Also, the learned policies across different numbers of queries have similar performance in final reward achieved.

Then, we explain the necessity of this query. In our problem setting, we do not have labels for either sample optimality or expertise levels. Without any labels, our classification method effectively performs an operation similar to clustering on the unlabeled data. A limited number of human queries is then used to ensure that the resulting clusters are assigned the correct labels. We select 5 as the number of queries so that the procedure remains both effective and practical.

\begin{table*}[t]
\centering
\caption{Classification results of the scoring model on imperfect demonstrations (Ant-v4) for Stage 1 and Stage 2 (Accuracy / Precision, \%), with Stage 1 $\rightarrow$ Stage 2 relative improvement (\%). Ablation study about expert query count. See Supplementary Fig.~\ref{fig:abl_expCheckNum}.}
\resizebox{\textwidth}{!}{%
\begin{tabular}{lc|cc|cc|cc|cc|cc}
\toprule
\textbf{Test} & \textbf{Stage} 
& \textbf{Query 1} & \textbf{Impr.} 
& \textbf{Query 3} & \textbf{Impr.} 
& \textbf{Query 5} & \textbf{Impr.} 
& \textbf{Query 7} & \textbf{Impr.}  
& \textbf{Query 9} & \textbf{Impr.} \\
\midrule
General & 1 
& 99.93 / 99.97 & -
& 99.92 / 99.94 & -
& 99.91 / 99.89 & -
& 99.92 / 99.96 & -
& 99.94 / 99.95 & -  \\

Expertise & 2 
& 99.92 / 99.92 & -0.01 / -0.05 
& 99.90 / 99.90 & -0.02 / -0.04 
& 99.89 / 99.89 & -0.02 / 0.00
& 99.89 / 99.89 & -0.03 / -0.07
& 99.91 / 99.91 & -0.03 / -0.04  \\
\midrule

Low & 1 
& 96.26 / 75.31 & -
& 97.33 / 80.17 & -
& 90.33 / 51.78 & -
& 97.02 / 78.81 & -
& 97.14 / 79.26 & - \\

Expertise & 2 
& 99.64 / 98.22 & 3.51 / 30.42
& 99.52 / 97.60 & 2.25 / 21.74
& 99.65 / 98.23 & 10.32 / 89.71
& 99.48 / 97.40 & 2.54 / 23.59
& 99.73 / 98.65 & 2.67 / 24.46 \\
\bottomrule
\end{tabular}
}
\label{tab:acc_prec_stages_exp_ask_num}
\end{table*}

\begin{table*}[t]
\centering
\caption{Ablation study on number of queries for Ant-v4. Reported as mean $\pm$ std of reward across random seeds; each seed's value is the mean over its last five checkpoints.}
\resizebox{0.5\columnwidth}{!}{%
\begin{tabular}{lcc}
\toprule
Query times & Expertise [0.1 -- 0.9] & Expertise [0.05 -- 0.15] \\
\midrule
1  & 5433.6 $\pm$ 365.6  & 5810.2 $\pm$ 301.3 \\
3  & 5655.2 $\pm$ 307.3  & 6011.2 $\pm$ 219.9 \\
5  & {6084.6 $\pm$ 146.9} & {5932.0 $\pm$ 335.2} \\
7  & 5981.4 $\pm$ 264.6  & 5944.2 $\pm$ 279.8 \\
9  & 5881.6 $\pm$ 265.3  & 6130.8 $\pm$ 126.8 \\
\bottomrule
\end{tabular}
}
\label{tab:query_ablation}
\end{table*}

\begin{figure}[h]
    \centering
    \begin{subfigure}{0.4\textwidth}
        \centering
        \includegraphics[width=\linewidth]{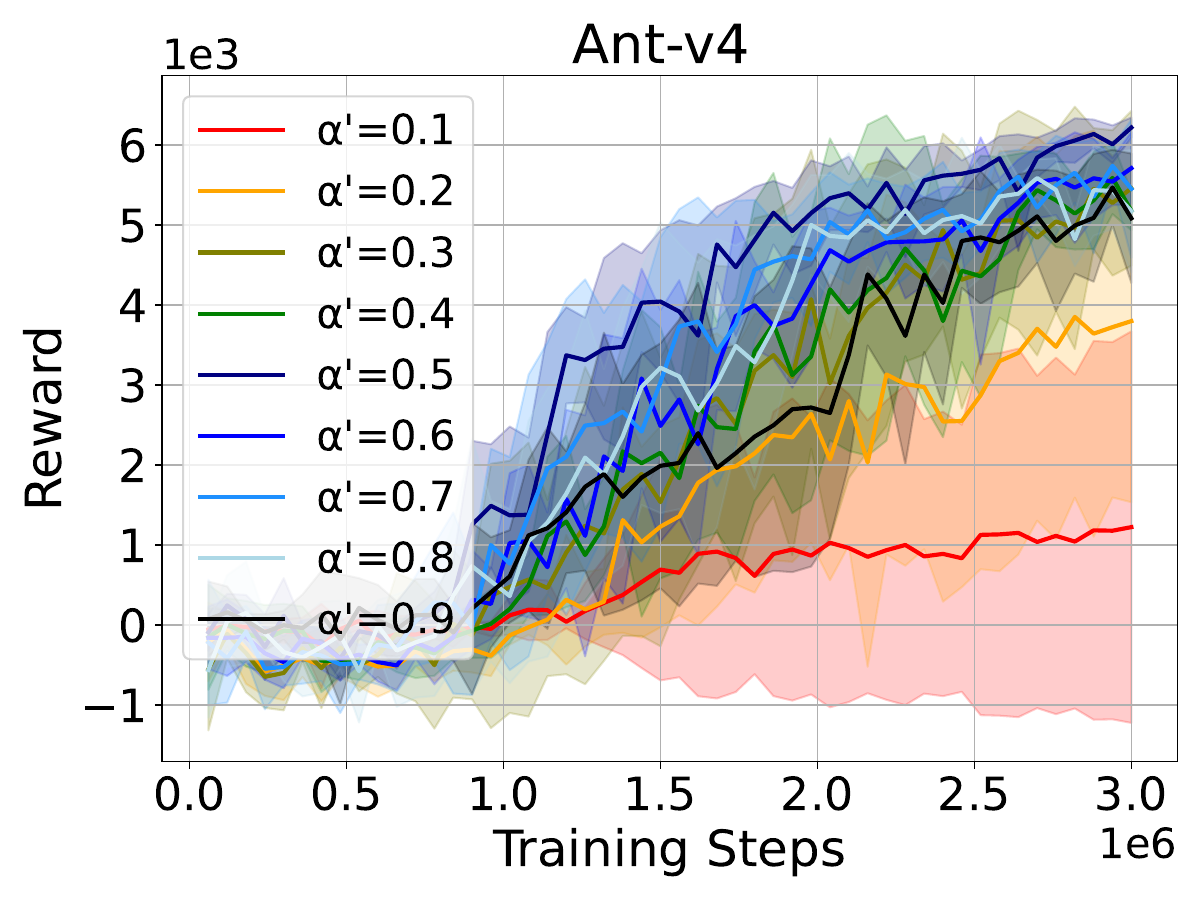}
        \caption{Expertise level range [0.1 - 0.9]}
    \end{subfigure}
    \hspace{0.5cm}
    \begin{subfigure}{0.4\textwidth}
        \centering
        \includegraphics[width=\linewidth]{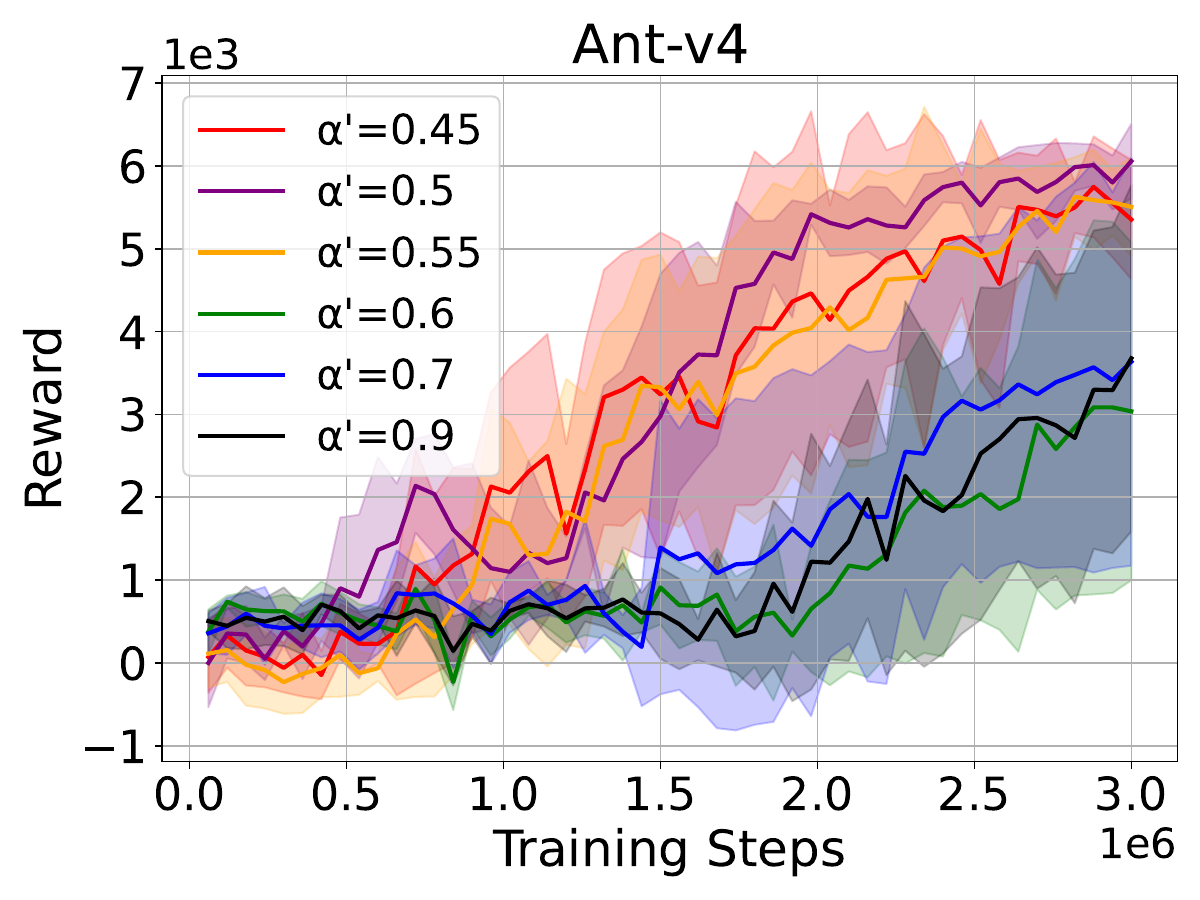}
        \caption{Expertise level range [0.05 - 0.15]}
    \end{subfigure}
    \caption{Ablation on the different  $\alpha'$ values (expected expertise in the target environment) for Ant-v4 task (mean $\pm$ std, shown as shaded region). See Supplementary Table~\ref{tab:antv4_pi_test_results_acc_prec}.}
    \label{fig:abl_testPi}
\end{figure}

\textbf{Expected Expertise $\alpha'$:}
To study the sensitivity to the value of $\alpha'$, we conduct ablation studies examining the effect of different $\alpha'$ values. Supplementary Table~\ref{tab:antv4_pi_test_results_acc_prec} presents the classification results of the scoring model, while Supplementary Table~\ref{tab:alpha_ablation} and Supplementary Fig.~\ref{fig:abl_testPi} report the rewards obtained by the learned policies. The performance of both the scoring model and the learned policy is maximized when $\alpha'$ is close to 0.5, and degrades when $\alpha'$ is extremely small or large.
These results indicate that selecting $\alpha' = 0.5$ yields the best performance for both the general expertise and low expertise test settings.

\begin{table*}[t]
\centering
\caption{Classification results of the scoring model on imperfect demonstrations for Stage 1 - Accuracy and Precision (\%, mean $\pm$ std) - for Ant-v4 across different  $\alpha'$ values (expected expertise in the target environment). See Supplementary Fig.~\ref{fig:abl_testPi}.}
\resizebox{0.6\columnwidth}{!}{%
\begin{tabular}{c|cc|cc}
\toprule
& \multicolumn{2}{c|}{\textbf{General expertise level}} 
& \multicolumn{2}{c}{\textbf{Low expertise level}} \\
\midrule
\textbf{$\alpha'$ }  
& \textbf{Accuracy} & \textbf{Precision} & \textbf{Accuracy} & \textbf{Precision} \\
\midrule
0.1 & 59.97 $\pm$ 19.94 & 20.00 $\pm$ 39.99 & 90.00 $\pm$ 0.00 & 0.00 $\pm$ 0.00 \\
0.2 & 89.94 $\pm$ 19.97 & 79.97 $\pm$ 39.99 & 90.00 $\pm$ 0.00 & 0.00 $\pm$ 0.00 \\
0.3 & 99.92 $\pm$ 0.02 & 99.96 $\pm$ 0.03 & - & - \\
0.4 & 99.94 $\pm$ 0.01 & 99.98 $\pm$ 0.02 & 90.00 $\pm$ 0.00 & 0.00 $\pm$ 0.00 \\
0.45 & - & - & 96.44 $\pm$ 2.43 & 88.46 $\pm$ 11.92 \\
0.5  & 99.91 $\pm$ 0.03 & 99.89 $\pm$ 0.07  &  90.33 $\pm$ 2.45 & 51.78 $\pm$ 7.56   \\
0.55 & - & - & 81.07 $\pm$ 35.54 & 77.18 $\pm$ 34.12 \\
0.6 & 99.92 $\pm$ 0.02 & 99.90 $\pm$ 0.05 & 20.03 $\pm$ 20.07 & 12.01 $\pm$ 4.02 \\
0.7 & 99.93 $\pm$ 0.02 & 99.93 $\pm$ 0.05 & 27.21 $\pm$ 34.42 & 22.35 $\pm$ 12.70 \\
0.8 & 99.92 $\pm$ 0.02 & 99.90 $\pm$ 0.04 & - & - \\
0.9 & 79.91 $\pm$ 24.43 & 79.87 $\pm$ 24.39 & 10.00 $\pm$ 0.00 & 10.00 $\pm$ 0.00 \\
\bottomrule
\end{tabular}
}
\label{tab:antv4_pi_test_results_acc_prec}
\end{table*}

\begin{table*}[h!]
\centering
\caption{Ablation study on $\alpha'$ for Ant-v4. Reported as mean $\pm$ std of reward across random seeds; each seed's value is the mean over its last five checkpoints. Note that there are no policy learning results appear for expertise [0.05 -- 0.15] when $\alpha' \leq 0.4$ because all imperfect demonstrations are labeled sub-optimal, leaving the optimal set empty.}
\begin{tabular}{lcc}
\toprule
$\alpha'$ value & Expertise [0.1 -- 0.9] & Expertise [0.05 -- 0.15] \\
\midrule
0.1  & 1149.4 $\pm$ 2298.8 & -- \\
0.2  & 3700.6 $\pm$ 2286.5 & -- \\
0.3  & 5241.6 $\pm$ 957.9  & -- \\
0.4  & 5308.0 $\pm$ 377.7  & -- \\
0.45 & -- & 5509.0 $\pm$ 493.4 \\
0.5  & {6084.6 $\pm$ 146.9} & 5932.0 $\pm$ 335.2 \\
0.55 & -- & 5496.6 $\pm$ 509.1 \\
0.6  & 5579.0 $\pm$ 416.2  & 2927.8 $\pm$ 2064.2 \\
0.7  & 5542.8 $\pm$ 261.6  & 3497.6 $\pm$ 2349.8 \\
0.8  & 5262.6 $\pm$ 330.8  & -- \\
0.9  & 5090.2 $\pm$ 573.1  & 3171.0 $\pm$ 1910.6 \\
\bottomrule
\end{tabular}
\label{tab:alpha_ablation}
\end{table*}

\textbf{Number of Initializations for Scoring Model Training}:
We report the number of initializations required by Algorithm~\ref{alg:1} in the main article in Supplementary Table~\ref{tab:repeat_num}. For the general expertise test, we use a single initialization until the Expertise Dispersion Criterion is satisfied. In more challenging low expertise cases, the number of initializations increases but remains fewer than 10. These results demonstrate that our algorithm is computationally efficient in practice.

\begin{table*}[t]
\centering
\begin{tabular}{c|cc}
\hline
\textbf{Env} & \textbf{General expertise (Mean ± Std)} & \textbf{Low expertise (Mean ± Std)} \\
\hline
Ant-v4 & $1.00 \pm 0.00$ & $8.80 \pm 3.31$ \\
HalfCheetah-v4 & $1.00 \pm 0.00$ & $9.80 \pm 1.94$ \\
Swimmer-v4 & $1.00 \pm 0.00$ & $3.60 \pm 4.72$ \\
Walker2d-v4 & $1.00 \pm 0.00$ & $2.80 \pm 3.12$ \\
\hline
\end{tabular}
\caption{Number of repeated computation loops required by Algorithm~\ref{alg:1} in the main article (mean $\pm$ std), using random initializations until the Expertise Dispersion Criterion is satisfied.}
\label{tab:repeat_num}
\end{table*}

\subsection{Additional Analysis of Method Variants}
\label{sec:more_analy}

We analyze our algorithm by comparing it against several variants under different settings. (1) \textbf{Scoring}: original method; (2) \textbf{Scoring PN}: in Stage 2, the first term {$\widehat{\mathbb{E}}_{(s,a)\sim \rho}\big[\ell(T(f_\phi(s,a)),\bar{z})\big]$} (demonstration discriminative loss) in Eq.~\ref{eq:scoring_loss} of the main article is replaced with a positive-negative (PN) classification loss (binary cross-entropy)
$\widehat{\mathbb{E}}_{\substack{(s,a)\sim\hat{\rho}^{\pi^*} \\ (s',a')\sim\hat{\rho}^{\pi^{\mathrm{sub}}}}}\big[\mathcal{L}_{\mathrm{PN}}(f_\phi(s,a),f_\phi(s',a'))\big]$,
where {$\mathcal{L}_{\mathrm{PN}}(f_\phi(s,a),f_\phi(s',a'))=-\log f_\phi(s,a)-\log(1-f_\phi(s',a'))$}.
Here, $\hat{\rho}^{\pi^*}$ and $\hat{\rho}^{\pi^{\mathrm{sub}}}$ represent the pseudo-optimal and pseudo-suboptimal data distributions inferred from $D_{\mathrm{opt}}$ and $D_{\mathrm{subopt}}=D\setminus D_{\mathrm{opt}}$, respectively. Thus, pseudo-optimal samples have target 1 and pseudo-suboptimal samples have target 0, consistent with the released binary-cross-entropy implementation. (3) \textbf{Scoring exp. known}: the true expertise levels are assumed to be given in both Stage 1 and 2. (4) \textbf{Scoring opt. known}: in Stage 2, $D_{\mathrm{opt}}$ is provided based on the true labels.

The results in Supplementary Fig.~\ref{fig:analy_combined} show that, overall, all methods achieve comparable performance.
\textbf{Scoring PN} yields slightly lower rewards in general expertise test on Ant and Swimmer tasks, which may be attributed to the overconfidence issue introduced by self-labeling when using the PN loss. This result highlights the effectiveness of our demonstration discriminative loss.
\textbf{Scoring exp. known} and \textbf{Scoring  opt. known} serve as oracle variants (with oracle knowledge of expertise levels and optimality labels, respectively) and achieve only marginally better performance than the original method in the low expertise test on HalfCheetah and Walker2d. This observation highlights the effectiveness of our approach in expertise-level estimation and optimality prediction when learning from imperfect demonstrations.

\begin{table*}[t]
\centering
\caption{Performance comparison of BC, ILEED, and LEOS-IL (ours), reported as mean $\pm$ std of reward. The “best policy” refers to the checkpoint with the highest true reward (oracle) over the entire training process, while “last 5 policies mean” denotes the average performance over the final five training checkpoints.}
\resizebox{\textwidth}{!}{%
\begin{tabular}{l|cc|ccc}
\toprule
Environment & BC (best policy) & ILEED (best policy) & BC (last 5 policies mean) & ILEED (last 5 policies mean)  & LEOS-IL (ours), (last 5 policies mean) \\
\midrule
\multicolumn{6}{l}{\textbf{General expertise test}} \\
\midrule
Ant-v4        & 2088.5 $\pm$ 42.0  & 2188.6 $\pm$ 70.3  & 1341.8 $\pm$ 164.5 & 1428.2 $\pm$ 80.3  & \textbf{6084.6 $\pm$ 146.9} \\
HalfCheetah-v4 & 3352.5 $\pm$ 214.2 & 3545.7 $\pm$ 173.0 & 2318.1 $\pm$ 521.2 & 1526.2 $\pm$ 1020.4  & \textbf{8751.2 $\pm$ 1797.9} \\
Swimmer-v4    & 329.4 $\pm$ 17.7  & 344.6 $\pm$ 1.4    & 21.3 $\pm$ 23.0    & 16.8 $\pm$ 20.5     & \textbf{305.4 $\pm$ 15.7} \\
Walker2d-v4   & 1472.7 $\pm$ 127.9 & 1439.2 $\pm$ 97.2  & 308.4 $\pm$ 111.5  & 440.3 $\pm$ 313.0   & \textbf{3959.2 $\pm$ 671.4} \\
\midrule
\multicolumn{6}{l}{\textbf{Low expertise test}} \\
\midrule
Ant-v4        & 2033.2 $\pm$ 55.7  & 2041.3 $\pm$ 32.6  & 1473.4 $\pm$ 97.7  & 1376.5 $\pm$ 62.1    & \textbf{5932.0 $\pm$ 335.2} \\
HalfCheetah-v4 & 4077.1 $\pm$ 24.9 & 4074.9 $\pm$ 33.3 & 3753.9 $\pm$ 220.5 & 3767.6 $\pm$ 176.9   & \textbf{7335.0 $\pm$ 1316.4} \\
Swimmer-v4    & 220.3 $\pm$ 71.3  & 255.2 $\pm$ 54.7   & 27.8 $\pm$ 11.6    & 34.0 $\pm$ 2.4       & \textbf{195.4 $\pm$ 92.1} \\
Walker2d-v4   & 1345.3 $\pm$ 110.0 & 1509.3 $\pm$ 213.9 & 436.6 $\pm$ 69.1   & 605.3 $\pm$ 94.5     & \textbf{2961.8 $\pm$ 339.8} \\
\bottomrule
\end{tabular}
}
\noindent\parbox{\textwidth}{\footnotesize{\textit{Note:} Boldface denotes the highest last-five-checkpoint mean. Paired $t$-tests ($\alpha=0.01$) confirm that LEOS-IL significantly outperforms baselines.}}
\label{tab:bc_ileed_scoring}
\end{table*}

\end{appendices}

\clearpage
\supplementbibliographytrue
\input{imperfect-imitation-merged.bbl}
\end{bibunit}
\end{document}

%% file: imperfect-imitation-merged.bbl
\ifsupplementbibliography

\else

\fi

%% file: imperfect-imitation-merged.bbl
\begin{thebibliography}{6}
\ifx \bisbn   \undefined \def \bisbn  #1{ISBN #1}\fi
\ifx \binits  \undefined \def \binits#1{#1}\fi
\ifx \bauthor  \undefined \def \bauthor#1{#1}\fi
\ifx \batitle  \undefined \def \batitle#1{#1}\fi
\ifx \bjtitle  \undefined \def \bjtitle#1{#1}\fi
\ifx \bvolume  \undefined \def \bvolume#1{\textbf{#1}}\fi
\ifx \byear  \undefined \def \byear#1{#1}\fi
\ifx \bissue  \undefined \def \bissue#1{#1}\fi
\ifx \bfpage  \undefined \def \bfpage#1{#1}\fi
\ifx \blpage  \undefined \def \blpage #1{#1}\fi
\ifx \burl  \undefined \def \burl#1{\textsf{#1}}\fi
\ifx \doiurl  \undefined \def \doiurl#1{\url{https://doi.org/#1}}\fi
\ifx \betal  \undefined \def \betal{\textit{et al.}}\fi
\ifx \binstitute  \undefined \def \binstitute#1{#1}\fi
\ifx \binstitutionaled  \undefined \def \binstitutionaled#1{#1}\fi
\ifx \bctitle  \undefined \def \bctitle#1{#1}\fi
\ifx \beditor  \undefined \def \beditor#1{#1}\fi
\ifx \bpublisher  \undefined \def \bpublisher#1{#1}\fi
\ifx \bbtitle  \undefined \def \bbtitle#1{#1}\fi
\ifx \bedition  \undefined \def \bedition#1{#1}\fi
\ifx \bseriesno  \undefined \def \bseriesno#1{#1}\fi
\ifx \blocation  \undefined \def \blocation#1{#1}\fi
\ifx \bsertitle  \undefined \def \bsertitle#1{#1}\fi
\ifx \bsnm \undefined \def \bsnm#1{#1}\fi
\ifx \bsuffix \undefined \def \bsuffix#1{#1}\fi
\ifx \bparticle \undefined \def \bparticle#1{#1}\fi
\ifx \barticle \undefined \def \barticle#1{#1}\fi
\bibcommenthead
\ifx \bconfdate \undefined \def \bconfdate #1{#1}\fi
\ifx \botherref \undefined \def \botherref #1{#1}\fi
\ifx \url \undefined \def \url#1{\textsf{#1}}\fi
\ifx \bchapter \undefined \def \bchapter#1{#1}\fi
\ifx \bbook \undefined \def \bbook#1{#1}\fi
\ifx \bcomment \undefined \def \bcomment#1{#1}\fi
\ifx \oauthor \undefined \def \oauthor#1{#1}\fi
\ifx \citeauthoryear \undefined \def \citeauthoryear#1{#1}\fi
\ifx \endbibitem  \undefined \def \endbibitem {}\fi
\ifx \bconflocation  \undefined \def \bconflocation#1{#1}\fi
\ifx \arxivurl  \undefined \def \arxivurl#1{\textsf{#1}}\fi
\csname PreBibitemsHook\endcsname

\bibitem[\protect\citeauthoryear{Beliaev et~al.}{2022}]{beliaev2022imitation}
\begin{bchapter}
\bauthor{\bsnm{Beliaev}, \binits{M.}},
\bauthor{\bsnm{Shih}, \binits{A.}},
\bauthor{\bsnm{Ermon}, \binits{S.}},
\bauthor{\bsnm{Sadigh}, \binits{D.}},
\bauthor{\bsnm{Pedarsani}, \binits{R.}}:
\bctitle{Imitation learning by estimating expertise of demonstrators}.
In: \bbtitle{International Conference on Machine Learning},
pp. \bfpage{1732}--\blpage{1748}
(\byear{2022}).
\bcomment{PMLR}
\end{bchapter}
\endbibitem

\bibitem[\protect\citeauthoryear{Brown et~al.}{2019}]{brown2019extrapolating}
\begin{bchapter}
\bauthor{\bsnm{Brown}, \binits{D.}},
\bauthor{\bsnm{Goo}, \binits{W.}},
\bauthor{\bsnm{Nagarajan}, \binits{P.}},
\bauthor{\bsnm{Niekum}, \binits{S.}}:
\bctitle{Extrapolating beyond suboptimal demonstrations via inverse
  reinforcement learning from observations}.
In: \bbtitle{International Conference on Machine Learning},
pp. \bfpage{783}--\blpage{792}
(\byear{2019}).
\bcomment{PMLR}
\end{bchapter}
\endbibitem

\bibitem[\protect\citeauthoryear{Brown et~al.}{2020}]{brown2020better}
\begin{bchapter}
\bauthor{\bsnm{Brown}, \binits{D.S.}},
\bauthor{\bsnm{Goo}, \binits{W.}},
\bauthor{\bsnm{Niekum}, \binits{S.}}:
\bctitle{Better-than-demonstrator imitation learning via automatically-ranked
  demonstrations}.
In: \bbtitle{Conference on Robot Learning},
pp. \bfpage{330}--\blpage{359}
(\byear{2020}).
\bcomment{PMLR}
\end{bchapter}
\endbibitem

\bibitem[\protect\citeauthoryear{Cai et~al.}{2023}]{cai2023imitation}
\begin{barticle}
\bauthor{\bsnm{Cai}, \binits{X.-Q.}},
\bauthor{\bsnm{Zhang}, \binits{Y.-J.}},
\bauthor{\bsnm{Chiang}, \binits{C.-K.}},
\bauthor{\bsnm{Sugiyama}, \binits{M.}}:
\batitle{Imitation learning from vague feedback}.
\bjtitle{Advances in Neural Information Processing Systems}
\bvolume{36},
\bfpage{48275}--\blpage{48292}
(\byear{2023})
\end{barticle}
\endbibitem

\bibitem[\protect\citeauthoryear{Ramaswamy et~al.}{2016}]{ramaswamy2016mixture}
\begin{bchapter}
\bauthor{\bsnm{Ramaswamy}, \binits{H.}},
\bauthor{\bsnm{Scott}, \binits{C.}},
\bauthor{\bsnm{Tewari}, \binits{A.}}:
\bctitle{Mixture proportion estimation via kernel embeddings of distributions}.
In: \bbtitle{International Conference on Machine Learning},
pp. \bfpage{2052}--\blpage{2060}
(\byear{2016}).
\bcomment{PMLR}
\end{bchapter}
\endbibitem

\bibitem[\protect\citeauthoryear{Fu et~al.}{2017}]{fu2017learning}
\begin{botherref}
\oauthor{\bsnm{Fu}, \binits{J.}},
\oauthor{\bsnm{Luo}, \binits{K.}},
\oauthor{\bsnm{Levine}, \binits{S.}}:
Learning robust rewards with adversarial inverse reinforcement learning.
arXiv preprint arXiv:1710.11248
(2017)
\end{botherref}
\endbibitem

\end{thebibliography}

\begin{thebibliography}{31}
\ifx \bisbn   \undefined \def \bisbn  #1{ISBN #1}\fi
\ifx \binits  \undefined \def \binits#1{#1}\fi
\ifx \bauthor  \undefined \def \bauthor#1{#1}\fi
\ifx \batitle  \undefined \def \batitle#1{#1}\fi
\ifx \bjtitle  \undefined \def \bjtitle#1{#1}\fi
\ifx \bvolume  \undefined \def \bvolume#1{\textbf{#1}}\fi
\ifx \byear  \undefined \def \byear#1{#1}\fi
\ifx \bissue  \undefined \def \bissue#1{#1}\fi
\ifx \bfpage  \undefined \def \bfpage#1{#1}\fi
\ifx \blpage  \undefined \def \blpage #1{#1}\fi
\ifx \burl  \undefined \def \burl#1{\textsf{#1}}\fi
\ifx \doiurl  \undefined \def \doiurl#1{\url{https://doi.org/#1}}\fi
\ifx \betal  \undefined \def \betal{\textit{et al.}}\fi
\ifx \binstitute  \undefined \def \binstitute#1{#1}\fi
\ifx \binstitutionaled  \undefined \def \binstitutionaled#1{#1}\fi
\ifx \bctitle  \undefined \def \bctitle#1{#1}\fi
\ifx \beditor  \undefined \def \beditor#1{#1}\fi
\ifx \bpublisher  \undefined \def \bpublisher#1{#1}\fi
\ifx \bbtitle  \undefined \def \bbtitle#1{#1}\fi
\ifx \bedition  \undefined \def \bedition#1{#1}\fi
\ifx \bseriesno  \undefined \def \bseriesno#1{#1}\fi
\ifx \blocation  \undefined \def \blocation#1{#1}\fi
\ifx \bsertitle  \undefined \def \bsertitle#1{#1}\fi
\ifx \bsnm \undefined \def \bsnm#1{#1}\fi
\ifx \bsuffix \undefined \def \bsuffix#1{#1}\fi
\ifx \bparticle \undefined \def \bparticle#1{#1}\fi
\ifx \barticle \undefined \def \barticle#1{#1}\fi
\bibcommenthead
\ifx \bconfdate \undefined \def \bconfdate #1{#1}\fi
\ifx \botherref \undefined \def \botherref #1{#1}\fi
\ifx \url \undefined \def \url#1{\textsf{#1}}\fi
\ifx \bchapter \undefined \def \bchapter#1{#1}\fi
\ifx \bbook \undefined \def \bbook#1{#1}\fi
\ifx \bcomment \undefined \def \bcomment#1{#1}\fi
\ifx \oauthor \undefined \def \oauthor#1{#1}\fi
\ifx \citeauthoryear \undefined \def \citeauthoryear#1{#1}\fi
\ifx \endbibitem  \undefined \def \endbibitem {}\fi
\ifx \bconflocation  \undefined \def \bconflocation#1{#1}\fi
\ifx \arxivurl  \undefined \def \arxivurl#1{\textsf{#1}}\fi
\csname PreBibitemsHook\endcsname

\bibitem[\protect\citeauthoryear{Zare et~al.}{2024}]{zare2024survey}
\begin{botherref}
\oauthor{\bsnm{Zare}, \binits{M.}},
\oauthor{\bsnm{Kebria}, \binits{P.M.}},
\oauthor{\bsnm{Khosravi}, \binits{A.}},
\oauthor{\bsnm{Nahavandi}, \binits{S.}}:
A survey of imitation learning: Algorithms, recent developments, and
  challenges.
IEEE Transactions on Cybernetics
(2024)
\end{botherref}
\endbibitem

\bibitem[\protect\citeauthoryear{Belkhale et~al.}{2023}]{belkhale2023data}
\begin{barticle}
\bauthor{\bsnm{Belkhale}, \binits{S.}},
\bauthor{\bsnm{Cui}, \binits{Y.}},
\bauthor{\bsnm{Sadigh}, \binits{D.}}:
\batitle{Data quality in imitation learning}.
\bjtitle{Advances in neural information processing systems}
\bvolume{36},
\bfpage{80375}--\blpage{80395}
(\byear{2023})
\end{barticle}
\endbibitem

\bibitem[\protect\citeauthoryear{Saxena et~al.}{2025}]{saxena2025matters}
\begin{botherref}
\oauthor{\bsnm{Saxena}, \binits{V.}},
\oauthor{\bsnm{Bronars}, \binits{M.}},
\oauthor{\bsnm{Arachchige}, \binits{N.R.}},
\oauthor{\bsnm{Wang}, \binits{K.}},
\oauthor{\bsnm{Shin}, \binits{W.C.}},
\oauthor{\bsnm{Nasiriany}, \binits{S.}},
\oauthor{\bsnm{Mandlekar}, \binits{A.}},
\oauthor{\bsnm{Xu}, \binits{D.}}:
What matters in learning from large-scale datasets for robot manipulation.
arXiv preprint arXiv:2506.13536
(2025)
\end{botherref}
\endbibitem

\bibitem[\protect\citeauthoryear{Lee et~al.}{2022}]{lee2022semi}
\begin{bchapter}
\bauthor{\bsnm{Lee}, \binits{G.}},
\bauthor{\bsnm{Oh}, \binits{W.}},
\bauthor{\bsnm{Oh}, \binits{J.}},
\bauthor{\bsnm{Shin}, \binits{S.}},
\bauthor{\bsnm{Kim}, \binits{D.}},
\bauthor{\bsnm{Jeong}, \binits{J.}},
\bauthor{\bsnm{Choi}, \binits{S.}},
\bauthor{\bsnm{Oh}, \binits{S.}}:
\bctitle{Semi-supervised imitation learning with mixed qualities of
  demonstrations for autonomous driving}.
In: \bbtitle{2022 22nd International Conference on Control, Automation and
  Systems (ICCAS)},
pp. \bfpage{20}--\blpage{25}
(\byear{2022}).
\bcomment{IEEE}
\end{bchapter}
\endbibitem

\bibitem[\protect\citeauthoryear{Zhang et~al.}{2024}]{zhang2024100}
\begin{bchapter}
\bauthor{\bsnm{Zhang}, \binits{C.}},
\bauthor{\bsnm{Wang}, \binits{W.}},
\bauthor{\bsnm{Chen}, \binits{Z.}},
\bauthor{\bsnm{Xi}, \binits{J.}}:
\bctitle{100 drivers, 2200 km: A natural dataset of driving style toward
  human-centered intelligent driving systems}.
In: \bbtitle{2024 IEEE Intelligent Vehicles Symposium (IV)},
pp. \bfpage{351}--\blpage{356}
(\byear{2024}).
\bcomment{IEEE}
\end{bchapter}
\endbibitem

\bibitem[\protect\citeauthoryear{Pomerleau}{1988}]{pomerleau1988alvinn}
\begin{botherref}
\oauthor{\bsnm{Pomerleau}, \binits{D.A.}}:
Alvinn: An autonomous land vehicle in a neural network.
Advances in neural information processing systems
\textbf{1}
(1988)
\end{botherref}
\endbibitem

\bibitem[\protect\citeauthoryear{Ho and Ermon}{2016}]{ho2016generative}
\begin{botherref}
\oauthor{\bsnm{Ho}, \binits{J.}},
\oauthor{\bsnm{Ermon}, \binits{S.}}:
Generative adversarial imitation learning.
Advances in neural information processing systems
\textbf{29}
(2016)
\end{botherref}
\endbibitem

\bibitem[\protect\citeauthoryear{Tangkaratt
  et~al.}{2020}]{tangkaratt2020robust}
\begin{botherref}
\oauthor{\bsnm{Tangkaratt}, \binits{V.}},
\oauthor{\bsnm{Charoenphakdee}, \binits{N.}},
\oauthor{\bsnm{Sugiyama}, \binits{M.}}:
Robust imitation learning from noisy demonstrations.
arXiv preprint arXiv:2010.10181
(2020)
\end{botherref}
\endbibitem

\bibitem[\protect\citeauthoryear{Wu et~al.}{2019}]{wu2019imitation}
\begin{bchapter}
\bauthor{\bsnm{Wu}, \binits{Y.-H.}},
\bauthor{\bsnm{Charoenphakdee}, \binits{N.}},
\bauthor{\bsnm{Bao}, \binits{H.}},
\bauthor{\bsnm{Tangkaratt}, \binits{V.}},
\bauthor{\bsnm{Sugiyama}, \binits{M.}}:
\bctitle{Imitation learning from imperfect demonstration}.
In: \bbtitle{International Conference on Machine Learning},
pp. \bfpage{6818}--\blpage{6827}
(\byear{2019}).
\bcomment{PMLR}
\end{bchapter}
\endbibitem

\bibitem[\protect\citeauthoryear{Wang et~al.}{2023}]{wang2023unlabeled}
\begin{botherref}
\oauthor{\bsnm{Wang}, \binits{Y.}},
\oauthor{\bsnm{Du}, \binits{B.}},
\oauthor{\bsnm{Xu}, \binits{C.}}:
Unlabeled imperfect demonstrations in adversarial imitation learning.
arXiv preprint arXiv:2302.06271
(2023)
\end{botherref}
\endbibitem

\bibitem[\protect\citeauthoryear{Eysenbach
  et~al.}{2021}]{eysenbach2021replacing}
\begin{botherref}
\oauthor{\bsnm{Eysenbach}, \binits{B.}},
\oauthor{\bsnm{Levine}, \binits{S.}},
\oauthor{\bsnm{Salakhutdinov}, \binits{R.}}:
Replacing rewards with examples: Example-based policy search via recursive
  classification.
arXiv preprint arXiv:2103.12656
(2021)
\end{botherref}
\endbibitem

\bibitem[\protect\citeauthoryear{Wang et~al.}{2021}]{wang2021learning}
\begin{bchapter}
\bauthor{\bsnm{Wang}, \binits{Y.}},
\bauthor{\bsnm{Xu}, \binits{C.}},
\bauthor{\bsnm{Du}, \binits{B.}},
\bauthor{\bsnm{Lee}, \binits{H.}}:
\bctitle{Learning to weight imperfect demonstrations}.
In: \bbtitle{International Conference on Machine Learning},
pp. \bfpage{10961}--\blpage{10970}
(\byear{2021}).
\bcomment{PMLR}
\end{bchapter}
\endbibitem

\bibitem[\protect\citeauthoryear{Beliaev et~al.}{2022}]{beliaev2022imitation}
\begin{bchapter}
\bauthor{\bsnm{Beliaev}, \binits{M.}},
\bauthor{\bsnm{Shih}, \binits{A.}},
\bauthor{\bsnm{Ermon}, \binits{S.}},
\bauthor{\bsnm{Sadigh}, \binits{D.}},
\bauthor{\bsnm{Pedarsani}, \binits{R.}}:
\bctitle{Imitation learning by estimating expertise of demonstrators}.
In: \bbtitle{International Conference on Machine Learning},
pp. \bfpage{1732}--\blpage{1748}
(\byear{2022}).
\bcomment{PMLR}
\end{bchapter}
\endbibitem

\bibitem[\protect\citeauthoryear{Tangkaratt
  et~al.}{2020}]{tangkaratt2020variational}
\begin{bchapter}
\bauthor{\bsnm{Tangkaratt}, \binits{V.}},
\bauthor{\bsnm{Han}, \binits{B.}},
\bauthor{\bsnm{Khan}, \binits{M.E.}},
\bauthor{\bsnm{Sugiyama}, \binits{M.}}:
\bctitle{Variational imitation learning with diverse-quality demonstrations}.
In: \bbtitle{International Conference on Machine Learning},
pp. \bfpage{9407}--\blpage{9417}
(\byear{2020}).
\bcomment{PMLR}
\end{bchapter}
\endbibitem

\bibitem[\protect\citeauthoryear{Kuhar et~al.}{2023}]{kuhar2023learning}
\begin{bchapter}
\bauthor{\bsnm{Kuhar}, \binits{S.}},
\bauthor{\bsnm{Cheng}, \binits{S.}},
\bauthor{\bsnm{Chopra}, \binits{S.}},
\bauthor{\bsnm{Bronars}, \binits{M.}},
\bauthor{\bsnm{Xu}, \binits{D.}}:
\bctitle{Learning to discern: Imitating heterogeneous human demonstrations with
  preference and representation learning}.
In: \bbtitle{Conference on Robot Learning},
pp. \bfpage{1437}--\blpage{1449}
(\byear{2023}).
\bcomment{PMLR}
\end{bchapter}
\endbibitem

\bibitem[\protect\citeauthoryear{Cao et~al.}{2024}]{cao2024limited}
\begin{bchapter}
\bauthor{\bsnm{Cao}, \binits{X.}},
\bauthor{\bsnm{Luo}, \binits{F.-M.}},
\bauthor{\bsnm{Ye}, \binits{J.}},
\bauthor{\bsnm{Xu}, \binits{T.}},
\bauthor{\bsnm{Zhang}, \binits{Z.}},
\bauthor{\bsnm{Yu}, \binits{Y.}}:
\bctitle{Limited preference aided imitation learning from imperfect
  demonstrations}.
In: \bbtitle{Forty-first International Conference on Machine Learning}
(\byear{2024})
\end{bchapter}
\endbibitem

\bibitem[\protect\citeauthoryear{Sutton et~al.}{1998}]{sutton1998reinforcement}
\begin{bbook}
\bauthor{\bsnm{Sutton}, \binits{R.S.}},
\bauthor{\bsnm{Barto}, \binits{A.G.}}, \betal:
\bbtitle{Reinforcement Learning: An Introduction}
vol. \bseriesno{1}.
\bpublisher{MIT Press},
\blocation{Cambridge, MA}
(\byear{1998})
\end{bbook}
\endbibitem

\bibitem[\protect\citeauthoryear{Lu et~al.}{2019}]{lu2018minimal}
\begin{bchapter}
\bauthor{\bsnm{Lu}, \binits{N.}},
\bauthor{\bsnm{Niu}, \binits{G.}},
\bauthor{\bsnm{Menon}, \binits{A.K.}},
\bauthor{\bsnm{Sugiyama}, \binits{M.}}:
\bctitle{On the minimal supervision for training any binary classifier from
  only unlabeled data}.
In: \bbtitle{ICLR}
(\byear{2019})
\end{bchapter}
\endbibitem

\bibitem[\protect\citeauthoryear{Lu et~al.}{2020}]{lu2020mitigating}
\begin{bchapter}
\bauthor{\bsnm{Lu}, \binits{N.}},
\bauthor{\bsnm{Zhang}, \binits{T.}},
\bauthor{\bsnm{Niu}, \binits{G.}},
\bauthor{\bsnm{Sugiyama}, \binits{M.}}:
\bctitle{Mitigating overfitting in supervised classification from two unlabeled
  datasets: A consistent risk correction approach}.
In: \bbtitle{AISTATS}
(\byear{2020})
\end{bchapter}
\endbibitem

\bibitem[\protect\citeauthoryear{Buja et~al.}{2005}]{buja2005loss}
\begin{botherref}
\oauthor{\bsnm{Buja}, \binits{A.}},
\oauthor{\bsnm{Stuetzle}, \binits{W.}},
\oauthor{\bsnm{Shen}, \binits{Y.}}:
Loss functions for binary class probability estimation and classification:
  Structure and applications.
Technical report
(2005)
\end{botherref}
\endbibitem

\bibitem[\protect\citeauthoryear{Lu et~al.}{2021}]{lu2021binary}
\begin{bchapter}
\bauthor{\bsnm{Lu}, \binits{N.}},
\bauthor{\bsnm{Lei}, \binits{S.}},
\bauthor{\bsnm{Niu}, \binits{G.}},
\bauthor{\bsnm{Sato}, \binits{I.}},
\bauthor{\bsnm{Sugiyama}, \binits{M.}}:
\bctitle{Binary classification from multiple unlabeled datasets via surrogate
  set classification}.
In: \bbtitle{International Conference on Machine Learning},
pp. \bfpage{7134}--\blpage{7144}
(\byear{2021}).
\bcomment{PMLR}
\end{bchapter}
\endbibitem

\bibitem[\protect\citeauthoryear{Charoenphakdee
  et~al.}{2019}]{charoenphakdee2019symmetric}
\begin{bchapter}
\bauthor{\bsnm{Charoenphakdee}, \binits{N.}},
\bauthor{\bsnm{Lee}, \binits{J.}},
\bauthor{\bsnm{Sugiyama}, \binits{M.}}:
\bctitle{On symmetric losses for learning from corrupted labels}.
In: \bbtitle{International Conference on Machine Learning},
pp. \bfpage{961}--\blpage{970}
(\byear{2019}).
\bcomment{PMLR}
\end{bchapter}
\endbibitem

\bibitem[\protect\citeauthoryear{Menon et~al.}{2015}]{menon2015learning}
\begin{bchapter}
\bauthor{\bsnm{Menon}, \binits{A.}},
\bauthor{\bsnm{Van~Rooyen}, \binits{B.}},
\bauthor{\bsnm{Ong}, \binits{C.S.}},
\bauthor{\bsnm{Williamson}, \binits{B.}}:
\bctitle{Learning from corrupted binary labels via class-probability
  estimation}.
In: \bbtitle{International Conference on Machine Learning},
pp. \bfpage{125}--\blpage{134}
(\byear{2015}).
\bcomment{PMLR}
\end{bchapter}
\endbibitem

\bibitem[\protect\citeauthoryear{Chang et~al.}{2021}]{chang2021mitigating}
\begin{barticle}
\bauthor{\bsnm{Chang}, \binits{J.}},
\bauthor{\bsnm{Uehara}, \binits{M.}},
\bauthor{\bsnm{Sreenivas}, \binits{D.}},
\bauthor{\bsnm{Kidambi}, \binits{R.}},
\bauthor{\bsnm{Sun}, \binits{W.}}:
\batitle{Mitigating covariate shift in imitation learning via offline data with
  partial coverage}.
\bjtitle{Advances in Neural Information Processing Systems}
\bvolume{34},
\bfpage{965}--\blpage{979}
(\byear{2021})
\end{barticle}
\endbibitem

\bibitem[\protect\citeauthoryear{Zhang et~al.}{2017}]{zhang2017mixup}
\begin{botherref}
\oauthor{\bsnm{Zhang}, \binits{H.}},
\oauthor{\bsnm{Cisse}, \binits{M.}},
\oauthor{\bsnm{Dauphin}, \binits{Y.N.}},
\oauthor{\bsnm{Lopez-Paz}, \binits{D.}}:
mixup: Beyond empirical risk minimization.
arXiv preprint arXiv:1710.09412
(2017)
\end{botherref}
\endbibitem

\bibitem[\protect\citeauthoryear{Orsini et~al.}{2021}]{orsini2021matters}
\begin{barticle}
\bauthor{\bsnm{Orsini}, \binits{M.}},
\bauthor{\bsnm{Raichuk}, \binits{A.}},
\bauthor{\bsnm{Hussenot}, \binits{L.}},
\bauthor{\bsnm{Vincent}, \binits{D.}},
\bauthor{\bsnm{Dadashi}, \binits{R.}},
\bauthor{\bsnm{Girgin}, \binits{S.}},
\bauthor{\bsnm{Geist}, \binits{M.}},
\bauthor{\bsnm{Bachem}, \binits{O.}},
\bauthor{\bsnm{Pietquin}, \binits{O.}},
\bauthor{\bsnm{Andrychowicz}, \binits{M.}}:
\batitle{What matters for adversarial imitation learning?}
\bjtitle{Advances in Neural Information Processing Systems}
\bvolume{34},
\bfpage{14656}--\blpage{14668}
(\byear{2021})
\end{barticle}
\endbibitem

\bibitem[\protect\citeauthoryear{Haarnoja et~al.}{2018}]{haarnoja2018soft}
\begin{bchapter}
\bauthor{\bsnm{Haarnoja}, \binits{T.}},
\bauthor{\bsnm{Zhou}, \binits{A.}},
\bauthor{\bsnm{Abbeel}, \binits{P.}},
\bauthor{\bsnm{Levine}, \binits{S.}}:
\bctitle{Soft actor-critic: Off-policy maximum entropy deep reinforcement
  learning with a stochastic actor}.
In: \bbtitle{International Conference on Machine Learning},
pp. \bfpage{1861}--\blpage{1870}
(\byear{2018}).
\bcomment{Pmlr}
\end{bchapter}
\endbibitem

\bibitem[\protect\citeauthoryear{Todorov et~al.}{2012}]{todorov2012mujoco}
\begin{bchapter}
\bauthor{\bsnm{Todorov}, \binits{E.}},
\bauthor{\bsnm{Erez}, \binits{T.}},
\bauthor{\bsnm{Tassa}, \binits{Y.}}:
\bctitle{Mujoco: A physics engine for model-based control}.
In: \bbtitle{2012 IEEE/RSJ International Conference on Intelligent Robots and
  Systems},
pp. \bfpage{5026}--\blpage{5033}
(\byear{2012}).
\bcomment{IEEE}
\end{bchapter}
\endbibitem

\bibitem[\protect\citeauthoryear{Towers et~al.}{2024}]{towers2024gymnasium}
\begin{botherref}
\oauthor{\bsnm{Towers}, \binits{M.}},
\oauthor{\bsnm{Kwiatkowski}, \binits{A.}},
\oauthor{\bsnm{Terry}, \binits{J.}},
\oauthor{\bsnm{Balis}, \binits{J.U.}},
\oauthor{\bsnm{De~Cola}, \binits{G.}},
\oauthor{\bsnm{Deleu}, \binits{T.}},
\oauthor{\bsnm{Goul{\~a}o}, \binits{M.}},
\oauthor{\bsnm{Kallinteris}, \binits{A.}},
\oauthor{\bsnm{Krimmel}, \binits{M.}},
\oauthor{\bsnm{KG}, \binits{A.}}, et al.:
Gymnasium: A standard interface for reinforcement learning environments.
arXiv preprint arXiv:2407.17032
(2024)
\end{botherref}
\endbibitem

\bibitem[\protect\citeauthoryear{Arulkumaran and
  Lillrank}{2024}]{arulkumaran2024pragmatic}
\begin{bchapter}
\bauthor{\bsnm{Arulkumaran}, \binits{K.}},
\bauthor{\bsnm{Lillrank}, \binits{D.O.}}:
\bctitle{A pragmatic look at deep imitation learning}.
In: \bbtitle{Asian Conference on Machine Learning},
pp. \bfpage{58}--\blpage{73}
(\byear{2024}).
\bcomment{PMLR}
\end{bchapter}
\endbibitem

\bibitem[\protect\citeauthoryear{Irpan et~al.}{2019}]{irpan2019off}
\begin{botherref}
\oauthor{\bsnm{Irpan}, \binits{A.}},
\oauthor{\bsnm{Rao}, \binits{K.}},
\oauthor{\bsnm{Bousmalis}, \binits{K.}},
\oauthor{\bsnm{Harris}, \binits{C.}},
\oauthor{\bsnm{Ibarz}, \binits{J.}},
\oauthor{\bsnm{Levine}, \binits{S.}}:
Off-policy evaluation via off-policy classification.
Advances in Neural Information Processing Systems
\textbf{32}
(2019)
\end{botherref}
\endbibitem

\end{thebibliography}
